  \documentclass{article}
\usepackage{amssymb,amsmath,mathrsfs,amsthm,eepic}
\usepackage{subfigure,graphicx,epsfig,color}

\newcommand{\RR}{\text{\sf I\!\!\:R}}

\title{An Elastic Image Registration Approach   \\
for Wireless Capsule Endoscope Localization}

 \author{
 {Isabel N. Figueiredo}\footnote{CMUC, Dep. of Mathematics, University of Coimbra, Portugal.}
\and {Carlos Leal\footnotemark[1]}
 \and {Lu\'is Pinto\footnotemark[1]}
 \and {Pedro N. Figueiredo}\footnote{Dep. of Gastroenterology,  CHUC and Faculty of Medicine, University of Coimbra, Portugal. }
\and {Richard Tsai}\footnote{ Dep. of Mathematics and the Institute for
 Computational Engineering and Sciences, The University of Texas at Austin,
 Austin, USA, and KTH Royal Institute of Technology, Sweden.}
 }

 \date{}

\begin{document}

\maketitle{}

\noindent
{ \textbf{Abstract}   -  Wireless Capsule Endoscope (WCE) is an innovative imaging device that permits physicians to examine all the areas of the Gastrointestinal (GI) tract. It is especially important for the small intestine, where traditional invasive endoscopies cannot reach. Although WCE represents an extremely important advance in medical imaging, a major drawback that remains unsolved  is the WCE precise location in the human body during its operating time. This is mainly due to the complex physiological environment and the inherent capsule effects during its movement. When an abnormality is detected, in the WCE images,  medical doctors do not know precisely where this abnormality is located relative to the intestine and therefore they can not proceed efficiently with the appropriate therapy.
The primary objective of the present paper is to give a contribution to WCE localization, using  image-based methods. The main focus of this work is on the  description of a multiscale elastic image registration approach,  its experimental application on WCE videos, and comparison with a multiscale affine registration. The proposed approach
 estimates the  motion of the walls of the  elastic small intestine,  in successive WCE frames. It includes registrations that capture  both rigid-like and non-rigid deformations, due respectively to the rigid-like WCE movement and the elastic deformation of the small intestine originated by the GI peristaltic movement. Under this approach  a qualitative information about the WCE speed  can be obtained,  as well as  the WCE location and orientation via projective geometry.
 The results of the experimental tests with real WCE video frames show  the good performance of the proposed  approach,  when elastic deformations of the small intestine are involved in successive frames, and its superiority with respect to a multiscale affine image registration, which accounts  for rigid-like deformations only and discards elastic deformations.

  \medskip

\noindent \textbf{Keywords - }{Elastic and Parametric Image Registration, Multiscale Representation, Wireless Capsule Endoscope.}

 \medskip



 \section{Introduction}\label{sec:intro}

Wireless capsule endoscopy is a medical technology, noninvasive,  devised for the {\it in vivo} and  painless inspection of the interior of
the GI tract. It is particularly important for the examination of the small intestine, since this organ
is not easily reached by  conventional endoscopic techniques.
The first capsule was developed by {\it Given Imaging} (Yoqneam, Israel)  in 2000 \cite{idan00} and after its approval in Europe  and the United States in 2001, it has been widely used by the medical community as a means of investigating small bowel diseases, namely GI bleeding and obscure GI bleeding (a bleeding of unknown origin that persists or recurs)
\cite{adler2003wireless,eliakim2010video, nakamura2008capsule}. This first capsule, for the small bowel examination,  is a very small device with the size and shape of a vitamin pill. It consists of a miniaturized
camera, a light source and a wireless circuit for the acquisition and transmission of signals \cite{moglia2008recent}. In a WCE
exam, a patient ingests  the capsule, and as it moves through the GI tract, propelled by peristalsis (a
contraction of the small intestine muscles that pushes the intestine content to move forward), images are
transmitted to a data recorder, worn on a belt outside the body. After about 8 hours, the WCE battery
lifetime, the stored images, approximately 50.000 images of the inside of the GI wall, are transferred to
a computer workstation for off-line viewing.
Despite the important medical benefits of wireless capsule endoscopy, one biggest drawback of this technology is the impossibility of knowing the WCE precise location when an abnormality is detected in the WCE video. For instance, for an abnormality in the small bowel, the principal medical goal is to know how far is the abnormality from a reference point  as for example, the pylorus (the opening from the stomach into the duodenum) or the ileocecal valve (the valve that separates the small from the large intestine), for planning
a surgical intervention  if necessary. Therefore, an accurate estimate of the WCE speed together with the location of one of these reference points (pylorus or ileocecal valve) would be medically extremely useful, since it would permit to measure  the distance from the reference point to the  capsule and consequently ({\it i.e.} equivalently)  the distance from the reference point to the region imaged by the capsule.

Recently, there have been many efforts to develop accurate localization methods for WCE and we refer to \cite{than2012review} for an extended review on this topic.  Generally, WCE localization techniques can be divided in three major categories: radio frequency (RF) signal based
\cite{RF_video_based, RF,RF3TOA,REF3DOATOA,pahlavan2012rf,RF2}, magnetic field based 
\cite{ciuti2011capsule,magnetic,magnetic2,liu2009capsule,liu2011hybrid,salerno2012discrete,magnetic3}, and image-based computer vision methods \cite{video_based_2,RF_video_based, bao2012modeling,cunha2008automated,video_based,liu2013wireless,liu2009capsule,
liu2011hybrid,spyrou2014video,szczypinski2009model,
zhou2014measurement,video_based_3}.  The first two typically require extra sensors installed outside the body.

The  monitoring of the RF waves emitted by the capsule antenna is a technique that has received considerable attention in the literature. Some of the strengths of this approach are that there is no need to redesign the capsule, since the RF antennas are already present in all
capsules, and also the potential high accuracy of the method.
For instance, in \cite{RF2}, using a three-dimensional human body model,
the authors suggest that it is possible to obtain an average localization error of $50$ mm in the digestive organs. An even lower error of $45.5$ mm is achieved in the small intestine. In particular, the technique presented is based on the measurement of the RF signal strength
using
receiving sensors placed on the surface of the human body model. In alternative, RF localization can also be based on the analysis of time-of-arrival (TOA) and direction-of-arrival (DOA) measurements \cite{RF,RF3TOA,REF3DOATOA}.
However, a number of difficulties remain to be resolved. First, the accuracy of these methods is highly
dependent on a relatively high number of external sensors. This external equipment can be very discomforting for the patient. Also, some of these techniques require the patient to be confined to a medical facility. These restrictions eliminate  some of the advantages that WCE has to offer. Moreover, the real human body is an an extremely complex medium having many non-homogeneous and non-isotropic parts that interfere with the RF signal. Therefore, in practice, the existing RF localization systems still suffer from high tracking errors.

The magnetic localization technique is similar in principle to RF signal techniques. The idea is to insert  a permanent magnet or a coil into the WCE and measure the resulting magnetic field with sensors placed outside the body. The permanent magnet method, unlike the coil based method, has the advantage that no external excitation current is needed. On the other hand, the latter, is less sensible to ambient electromagnetic noise. Magnetic based methods could benefit from the fact the human body has a very small influence on the magnetic field. Theoretically, the accuracy of these methods can be very high, e.g., average position errors of $3.3$ mm were reported in \cite{magnetic}. The main drawbacks associated with this technology are basically similar to those pointed out to RF methods: those are the need for a high number of external sensors and the restricted mobility of the patient. The modification of the capsule design may also be problematic. We also point out that magnetic localization systems are limited to 2D orientation estimation, since one rotation angle is missing.

One alternative technique that avoids any burden for the patient is based on computer-vision methods. Here only information extracted from
WCE images is used to estimate the displacement and orientation of the capsule. Generally, these methods involve as a first step image registration procedures between consecutive video frames. The registration process is carried out through the minimization of a global similarity measure, e.g. mutual information \cite{zhou2014measurement}, or the matching of local features, where algorithms like RANSAC and SIFT are the usual choices \cite{video_based,spyrou2014video}. The following step involves the estimation of the relative displacement and rotation of the wireless capsule.
Several different approaches have been proposed to achieve this goal. One such approach, and the one also followed
here, is to  relate the  scale and rotation parameters resulting from the registration scheme,  with
the capsule rotation and displacement,  using a projective transformation and the pinhole model \cite{spyrou2014video}. Another, more complex, approach is the model of deformable rings  \cite{szczypinski2009model}. Orientation estimation resorting to homography transformation \cite{video_based_3} or epipolar geometry \cite{liu2009capsule} has also been explored.

The main challenges in the computer based methods are the abrupt changes of the image content in consecutive frames and in the capsule motion,  caused by the peristaltic motion and the accompanying large deformation of the small intestine.
However a  common simplification used in image based WCE tracking, is to neglect the non-rigid deformations of the elastic intestine walls.  \begin{figure}[t!]
\centering
\includegraphics[width=4.25cm,height=4.25cm]{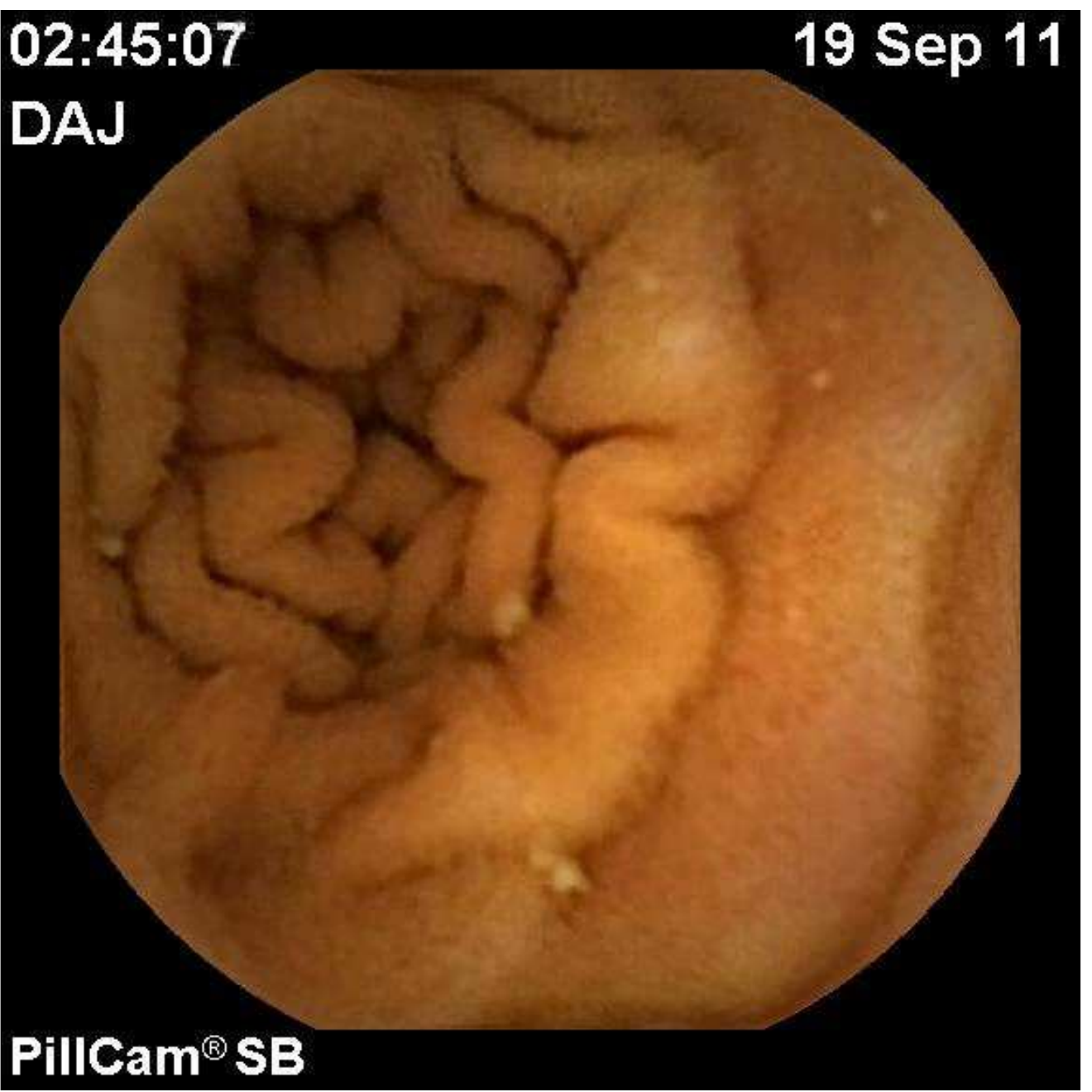} \hspace{2 mm}
\includegraphics[width=4.25cm,height=4.25cm]{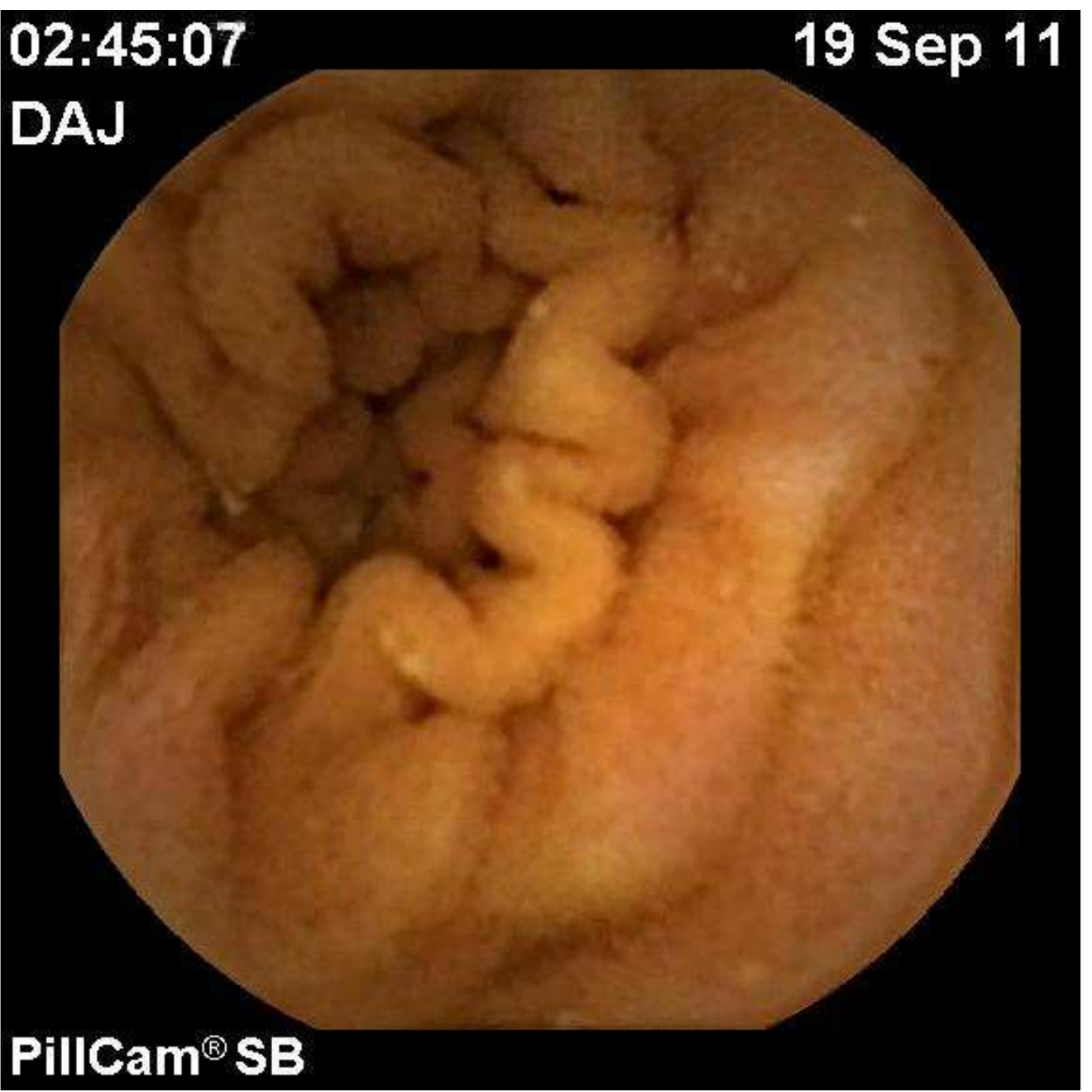}\hspace{2 mm}
\caption{Example of two consecutive frames in a WCE video.}\label{fig:exreal}
\end{figure}
In this paper we
develop an appropriate multiscale elastic  image registration strategy that tries to take into account this effect, and that overcomes the limitations of multiscale parametric  image registration (this latter  captures only rigid-like movements of the intestine walls in successive frames). By way of illustration Figure \ref{fig:exreal} shows two consecutive frames in a WCE video,  exhibiting elastic deformations, and demonstrating that an affine transformation composed of a planar rotation, scale   and translation transformations, is not enough to match (or equivalently to register)  the left  with the right frame.

 In fact, as observed in \cite{liu2013wireless}, and because WCE is propelled by peristalsis, the motion of the walls of the small intestine, in consecutive frames,  is a consequence of a combination of two types of movements: the WCE movement, which is rigid-like, and the nonrigid movement of the small intestine (because of the peristaltic movement, the small instestine, which  is  an elastic organ,  bends and deforms).  Therefore, in this paper we propose a
 multiscale elastic  image registration procedure, for measuring the motion of the walls of the small intestine  between consecutive frames,  that takes into account the combination of these two movements.  Firstly a parametric pre-registration is performed at a coarse scale, and gives the motion/deformation that corresponds to an affine alignment of the two images at a coarse scale, thus matching the most prominent and large features, and correcting the main distortions, originated by the WCE movement.
In the second step, and based on the result of the first step, a multiscale elastic registration is accomplished. This second step performs  the multiscale elastic motion/deformation, correcting the fine and local misalignments generated by the non-rigid movement of the gastrointestinal tract. The  motion obtained with this multiscale elastic  image registration,  in two consecutive video frames, is the final deformation resulting from these two aforementioned successive deformations. Moreover  we further enhance the quality of this approach, by iterating it twice.

  To the best of our knowledge this is the first time that a multiscale elastic  image registration (with an affine pre-registration) is proposed for WCE imaging motion. Moreover, under the proposed multiscale elastic image registration approach  we show that a qualitative information about the WCE speed can be obtained, as well as the WCE location and orientation by using projective geometry and following the aforementioned  arguments of  \cite{spyrou2014video} (that is, by  relating the  scale and rotation parameters resulting from the registration scheme,  with
the capsule orientation and displacement,  using  projective geometry analysis and the pinhole model). Furthermore,  the results of the tests and experiments  evidence a better performance of the multiscale elastic  image registration, when elastic deformations are involved (which is the realistic scenario because the capsule motion is driven by peristalsis), compared to the multiscale parametric image registration.

After this introduction, the rest of the paper is organized in three  sections. In Section \ref{sec:appr} we describe the  proposed multiscale  image registration approach (elastic with affine pre-registration) as well as  the fully parametric. In Section \ref{sec:results} we evaluate the proposed  procedure in  real  (and artificial)  WCE video frames and also compare it  with multiscale parametric image registration, in terms of the qualitative WCE speed information, the  dissimilarity measure for evaluating the registration,  and in terms of the WCE location and orientation by following  \cite{spyrou2014video}. We give an account of all the numerical tests done and the corresponding obtained results. Finally,  a  section with conclusions and future work closes the paper.

\section{Image Registration Approach}\label{sec:appr}

Let $(R,T)$ be a pair of images, one called  the reference $R$ (and that is kept unchanged) and the other called the template $T$,
 represented by the functions $R, \: T: \Omega \subset {\RR}^2 \longrightarrow \RR$, where $\Omega$ stands for the pixel domain, and  $x=(x_1, x_2) $ is the notation for an arbitrary pixel in $\Omega$.
The goal of image registration is to find a geometric transformation $\varphi$,   such  that the transformed template image, denoted by $T(\varphi)$, becomes similar to the reference image $R$, or equivalently, to solve  an optimization problem, where the objective is to find a  transformation $\varphi$ that minimizes the distance between $T(\varphi)$ and $R$, represented by a distance measure $\mathcal{D} \big (R, T(\varphi)\big )$.

In this paper we always consider the greyscale version of the WCE video frames to perform the registration and  the selected   distance measure $\mathcal{D} $, that quantifies the similarity (or alignment) of the reference and transformed  template images, under the transformation $\varphi$, is the    the sum of square differences  that directly compares the gray values of the reference and template images. This distance
  is defined by
\begin{equation}\label{eq:ssd}
  \displaystyle  \frac{1}{2} \big  \| T(\varphi) -R \big \|_{L^2(\Omega)}^2
  =   \displaystyle \frac{1}{2} \int_\Omega \Big (  T\big (\varphi(x)\big ) -R(x) \Big )^2\, dx
\end{equation}
where $L^2(\Omega)$ is  the space of square-integrable functions in $\Omega$.

In this section we describe the proposed image registration approach, which is a multiscale elastic image registration with an affine pre-registration, hereafter denoted by  MEIR. It relies on a multiscale representation of the image data (see Figure \ref{fig:msr}) that  originates   a sequence of image registration problems (that are optimization problems).   This multiscale representation is a strategy that attempts to diminish or eliminate several possible local minima  and  lead to convex optimization problems.

\subsection{Multiscale elastic image registration with affine pre-registration (MEIR)}\label{sec:meir}

 Let $\theta_i \in \RR$, with $i=0,1,\ldots, n$ and $n$ a positive integer, denote a decreasing sequence of  scale parameters, associated to a spline interpolation procedure \cite{modersitzki2009fair}.  By starting with the large initial $\theta_0$, that  is related to the coarse scale, we  denoted by  $R_{\theta_0}$ and $T_{\theta_0}$ the corresponding interpolated reference and template images. These will  retain only the most prominent features (small details in these images will disappear, as exemplified in Figure \ref{fig:msr}-c).  Then we perform a parametric pre-registration, that is, we search for a particular type of  affine transformation $\varphi$,
 a rigid-like one,
that is a composition of scaling, rotation and translations, defined by
\begin{equation}\label{eq:rig}
\varphi(x) :=  \omega_0
\left(
\begin{array}{ccc}
  \cos(\omega_1)   &  -\sin (\omega_1)  \\
  \sin (\omega_1)    &   \cos(\omega_1)
\end{array}
\right)
\left(
\begin{array}{c}
 x_1      \\
 x_2
\end{array}
\right) +
\left(
\begin{array}{c}
 \omega_2      \\
\omega_3
\end{array}
\right),
\end{equation}
and such that  $\varphi$ is the solution of the optimization problem
\begin{equation}\label{eq:opt}
\min_{\varphi}  \displaystyle  \frac{1}{2} \big  \| R_{\theta_0} - T_{\theta_0}(\varphi)\big  \|_{L^2(\Omega)}^2.
\end{equation}
In \eqref{eq:rig} $ [\omega_0 ; \, \omega_1 ; \,  \omega_2 ;  \, \omega_3] \in \RR^4$ is the vector with 4 parameters characterizing the rigid-like transformation  $\varphi$: $\omega_0$  represents the scale, $\omega_1$ is the rotation angle and  finally, $\omega_2$ and $\omega_3$ denote the translations on the $x-$ and $y-$ axis, respectively.

 We observe that a general affine transformation is  characterized not only by four parameters, as in \eqref{eq:rig}, but by six parameters. However we have restricted the search to transformations of the type \eqref{eq:rig}, because in this  initial pre-registration, at the coarse scale $\theta_0$, the objective  is to partially recover the rigid-like motion of the small intestine walls in  a pair of consecutive frames, due to the WCE movement which roughly induces  a two-dimensional rigid-like  apparent motion  of the form \eqref{eq:rig}  in the frames.

  Afterwards, the idea is to improve this rigid-like motion by complementing it with the non-rigid deformations of the small intestine walls.
  In fact, the WCE motion is caused by the intestine movement.

  \begin{figure}[t!]
\centering
\subfigure[]{\includegraphics[width=4cm,height=4cm]{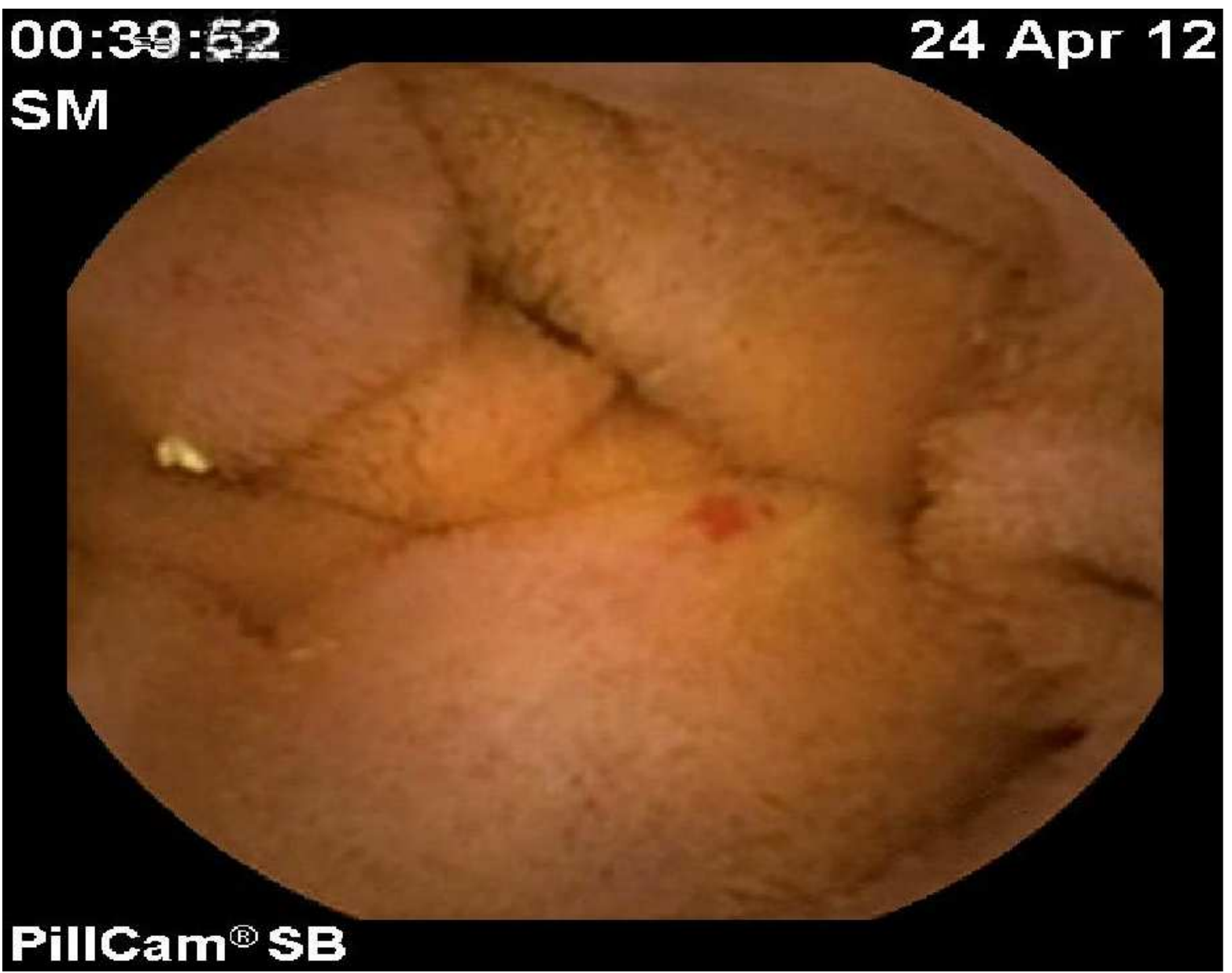}}\hspace{2 mm}
\subfigure[]{\includegraphics[width=4cm,height=4cm]{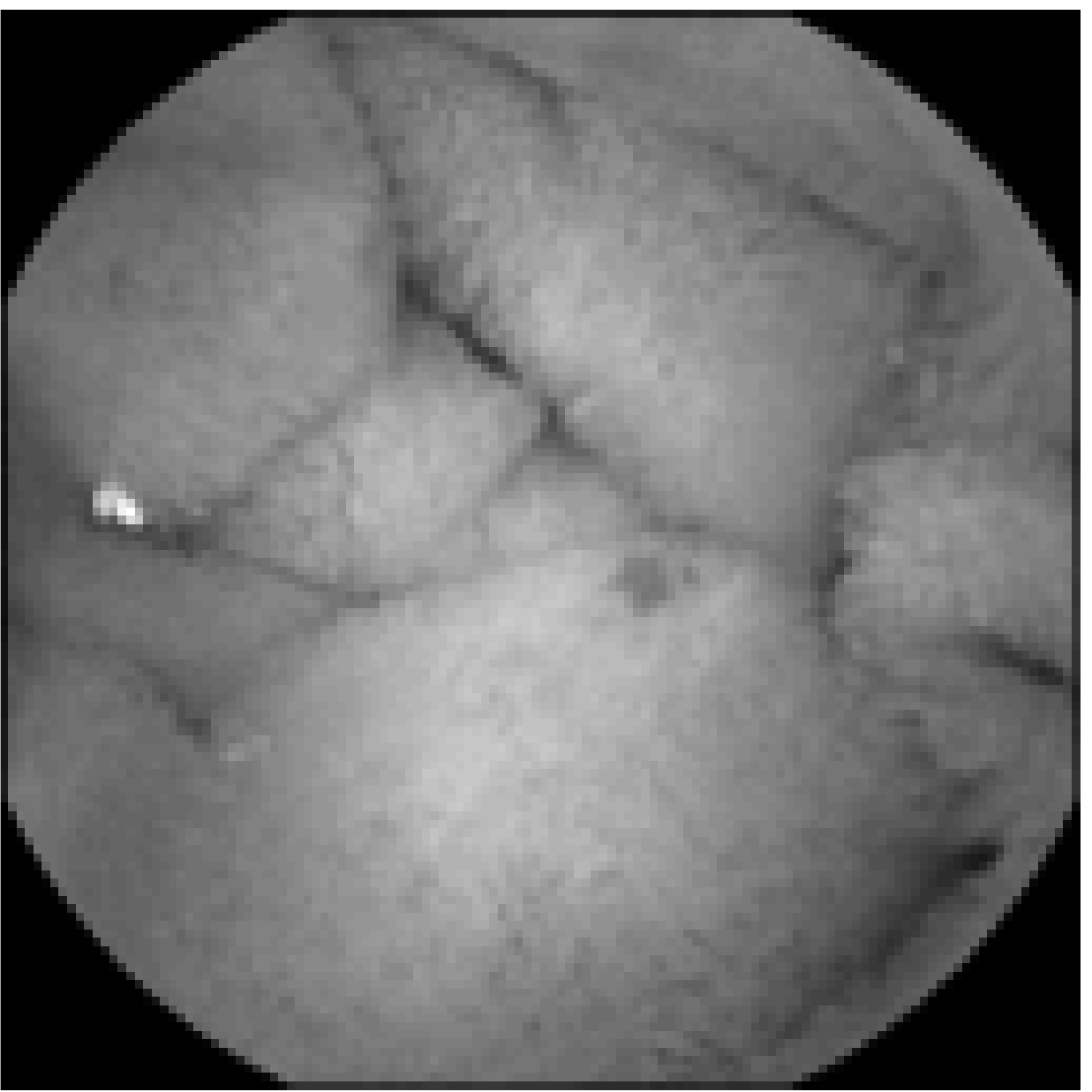}}\hspace{2 mm}\\
\subfigure[]{\includegraphics[width=4cm,height=4cm]{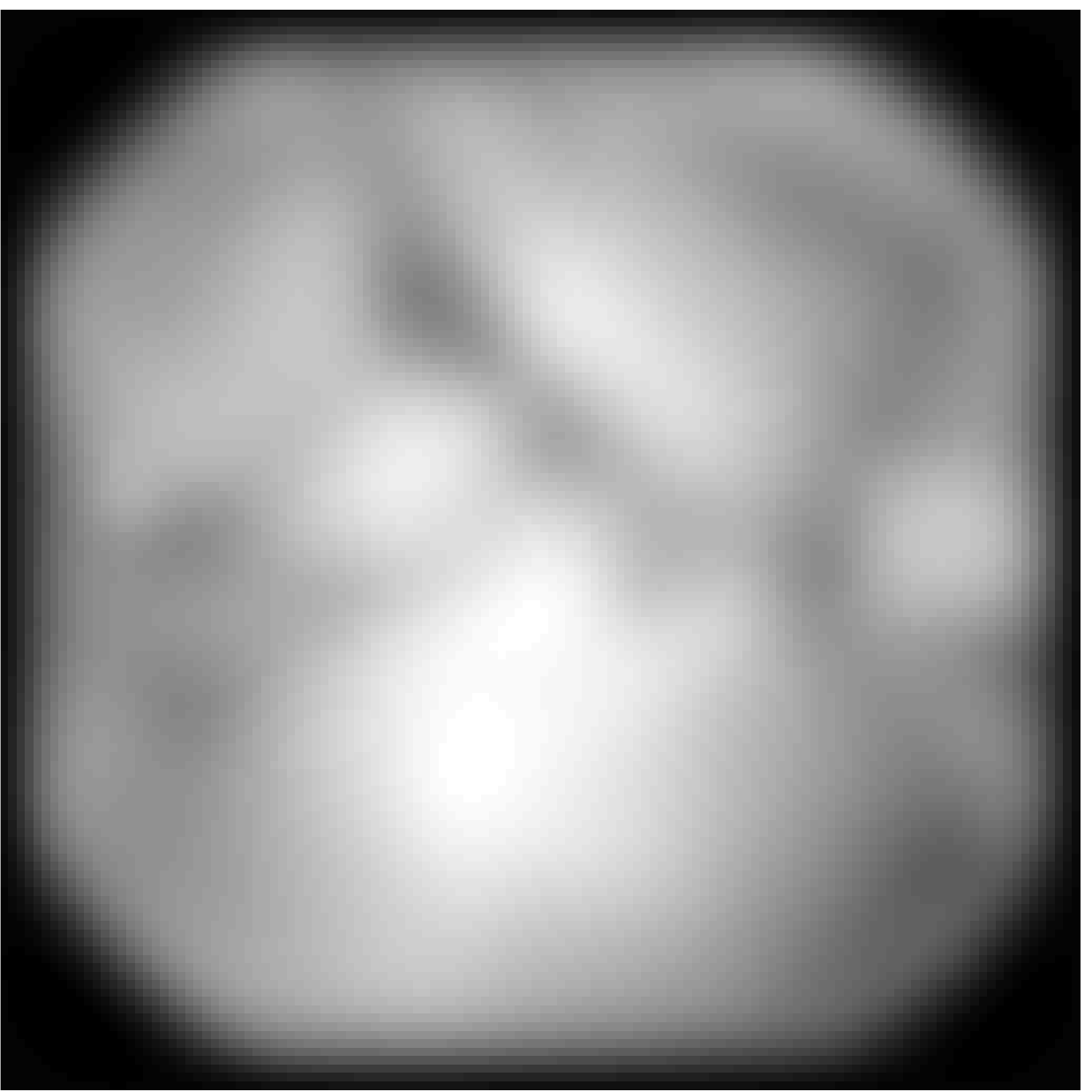}}\hspace{2 mm}
\subfigure[]{\includegraphics[width=4cm,height=4cm]{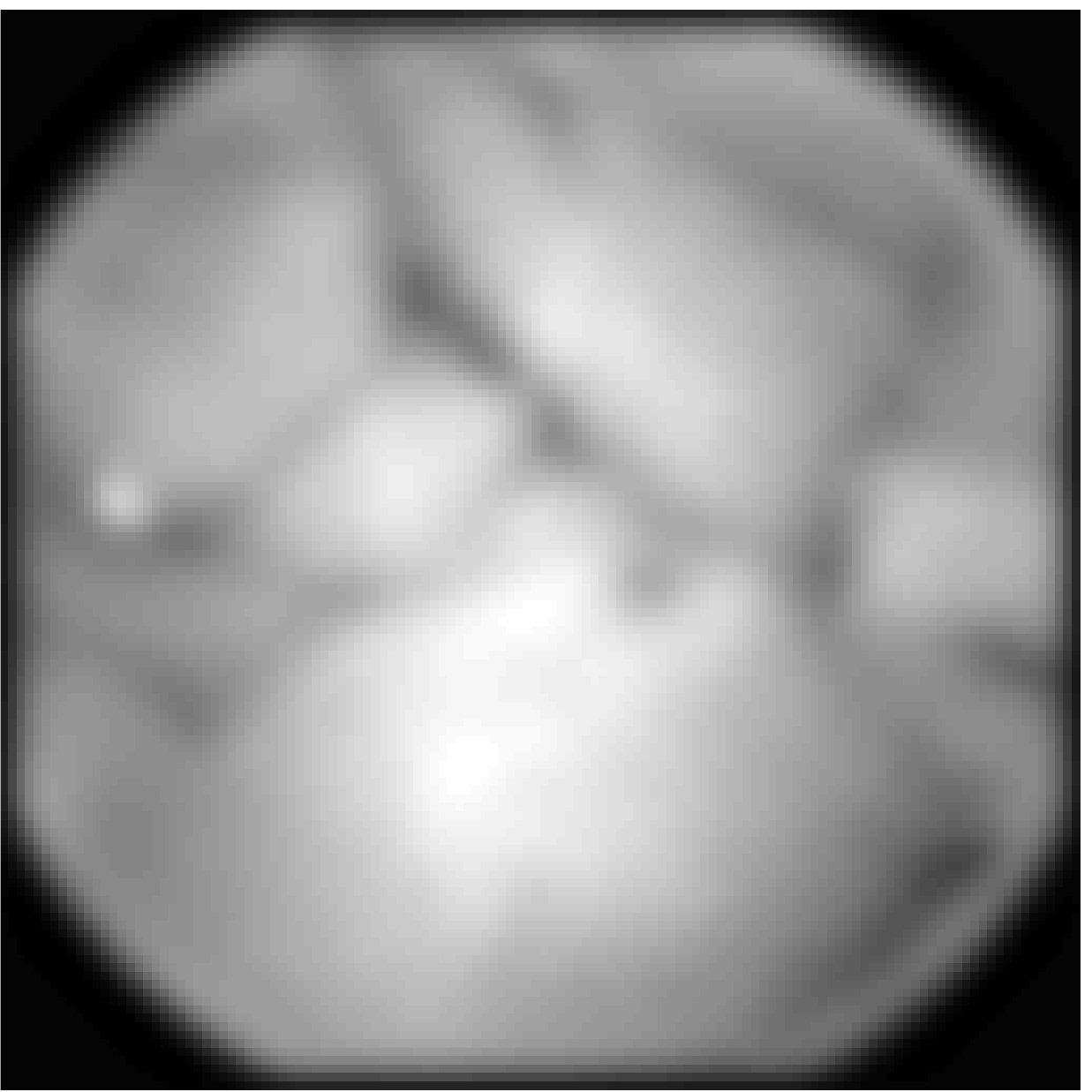}}\hspace{2 mm}
\subfigure[]{\includegraphics[width=4cm,height=4cm]{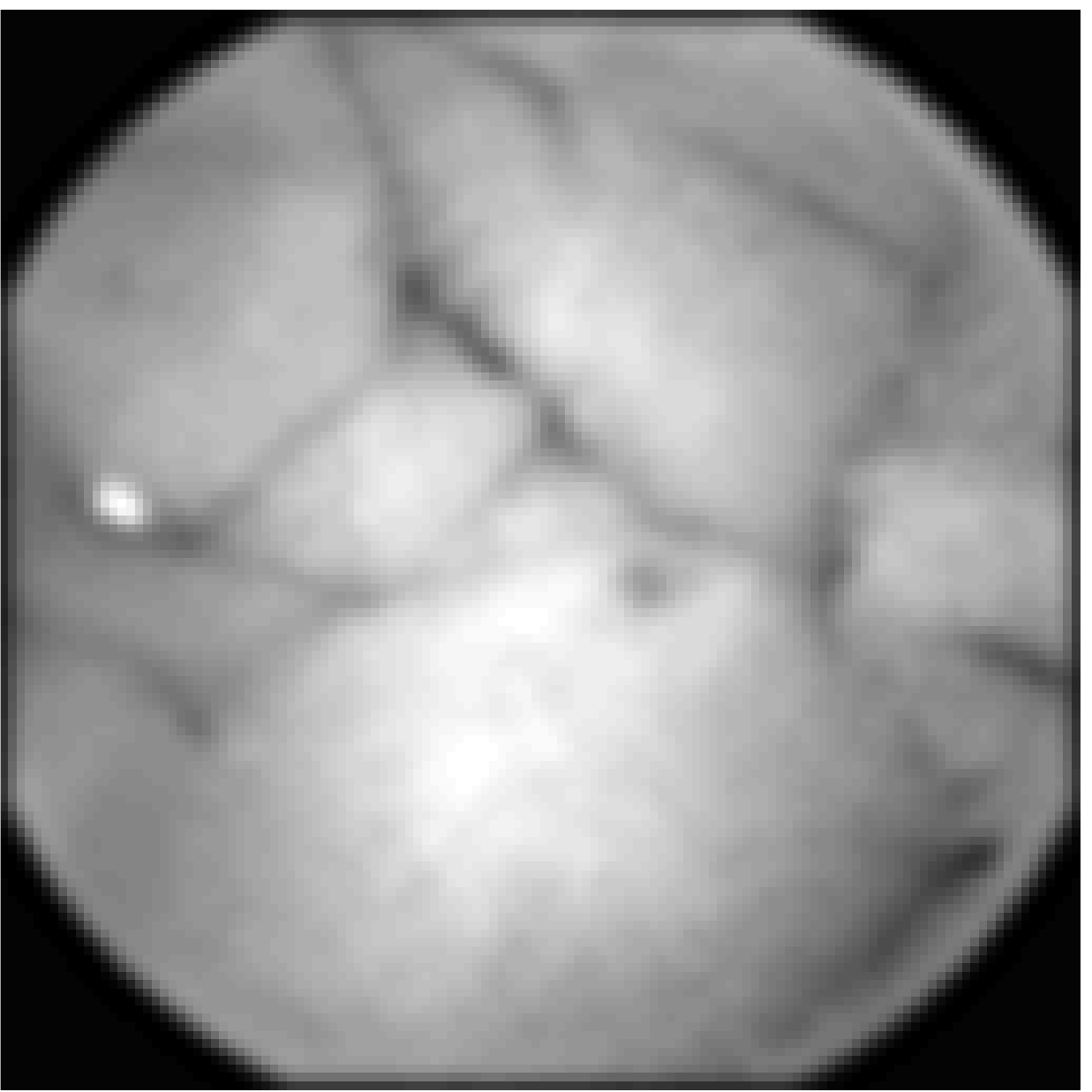}}
\caption{Multiscale representation of the grayscale version of a  WCE frame:  (a) Original frame displaying a bleeding region (the red spot). (b) Grayscale version coincident with the image representation at scale $\theta=0$.
(c), (d) and (e) Representations  at scales $\theta=100$,   $\theta=10 $ and  $\theta=1$, respectively.}\label{fig:msr}
\end{figure}

   Thus the goal is to do a loop over all the scales $\theta_i$, for carrying out the multiscale elastic registration,
 and using  the solution at scale $\theta_{i-1}$  as a starting point for the elastic image registration at the following finer scale $\theta_{i}$, aiming at speeding up the total optimization procedure  and avoiding possible local minima. To be precise, for each scale $\theta_i$, with $i=0,1,\ldots, n$  let  $R_{\theta_i}$ and $T_{\theta_i}$ be the corresponding interpolated reference and template images.  Figure \ref{fig:msr}  displays for a WCE video frame the  multiscale representation of its greyscale version, using 4 scales $\theta_0=100$, $\theta_1=10$, $\theta_2=1$, $\theta_3=0$.
 The objective is to find  a particular transformation $\varphi$ ({\it i.e.} an elastic deformation), that  for convenience is split into the trivial identity part and the deformation or displacement part $u$ (which means,
$\varphi(x):= (Id-u)(x) = x- u(x)$, with $ u:=(u_1, u_2)$), such that at scale  $\theta_{i}$  the transformed interpolated template image $T_{\theta_i}(\varphi)$ becomes similar to the interpolated reference image $R_{\theta_i}$. The  elastic registration problem to be solved at scale $\theta_i$  is the following optimization problem
 \begin{equation}\label{eq:multisca}
\min_{u} \Big [   \displaystyle  \frac{1}{2} \big  \| R_{\theta_i} , T_{\theta_i}\big (x-u(x)\big ) \big  \|_{L^2(\Omega)}^2 + \alpha S(u) \Big ],
\end{equation}
whose solution we denote by $u_{\theta_{i}}$. Here  $ S(u)$ is the elastic regularization term   (which should make the optimization problem well-posed  and  restrict the minimizer  $u$ to the group of  linear elastic transformations) defined by
\begin{equation}\label{eq:reg}
\mathcal {S}(u) :=  \displaystyle \int_\Omega \Big (  \frac{\lambda +\mu}{2} \|div \,  u \|^2 +  \frac{\mu}{2} \sum_{i=1}^{2} \|\nabla u_i \|^2  \Big ) \, dx,
\end{equation}
with  $\nabla$ and $div$ denoting, respectively, the gradient and divergence operators
\begin{equation}
 \nabla u_i    :=       \big ( \partial_1 u_i , \partial_2 u_i  \big ) , \qquad
div \, u   :=   \partial_1 u_1 + \partial_2 u_2, \quad \mbox{(for $i=1,2$),}
 \end{equation}
$\|. \|$ is the notation for  the Euclidean norm, and the parameters $\lambda$ and $\mu$ are the Lam\'e constants characterizing the elastic material.The constant $\alpha  > 0$  is a regularization parameter that balances the influence of the similarity and regularity terms in the cost functional of the optimization problem  \eqref{eq:multisca}.

\begin{figure}[t!]
\centering
\includegraphics[width=4.0cm,height=4.0cm]{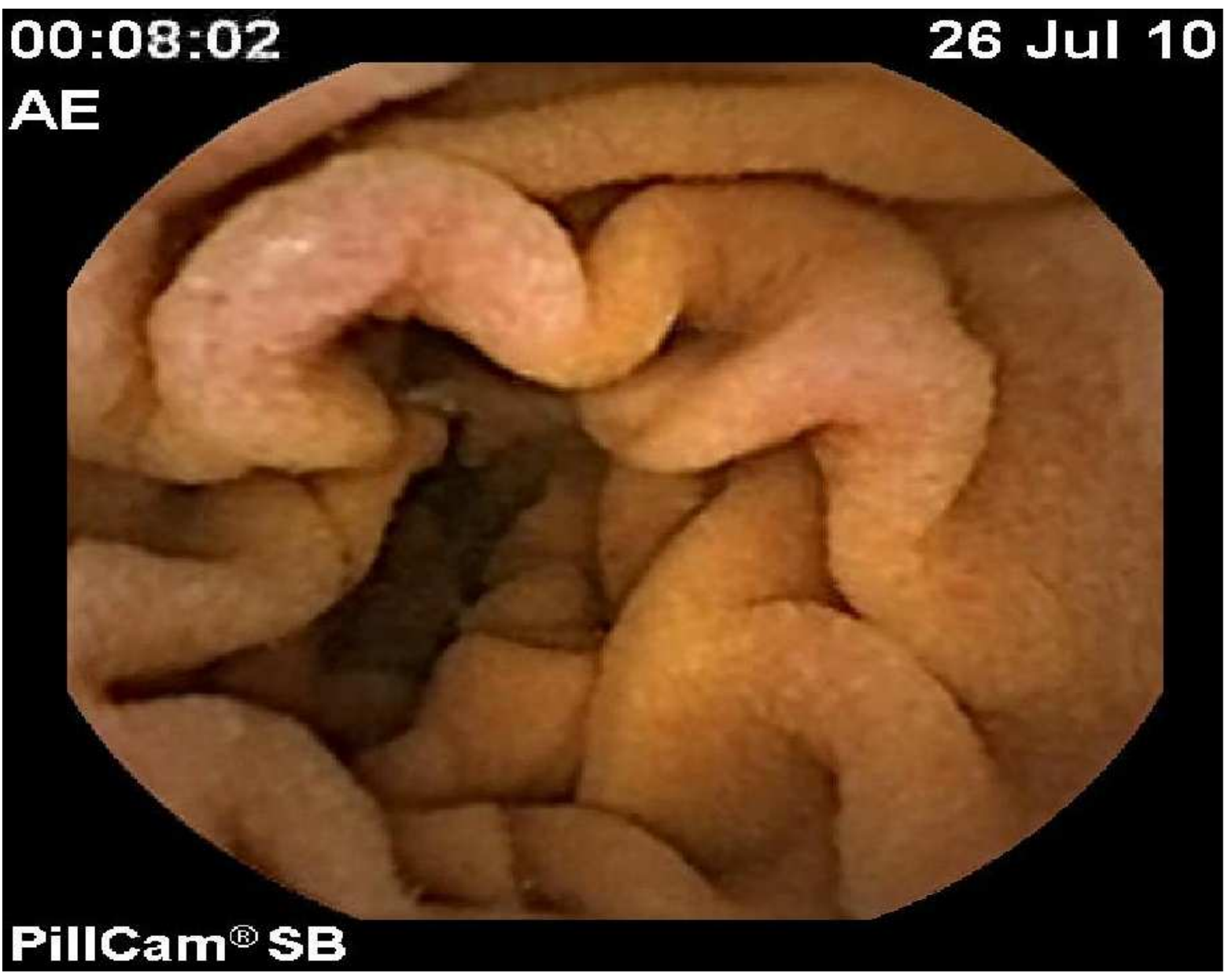}\hspace{2 mm}
\includegraphics[width=4.0cm,height=4.0cm]{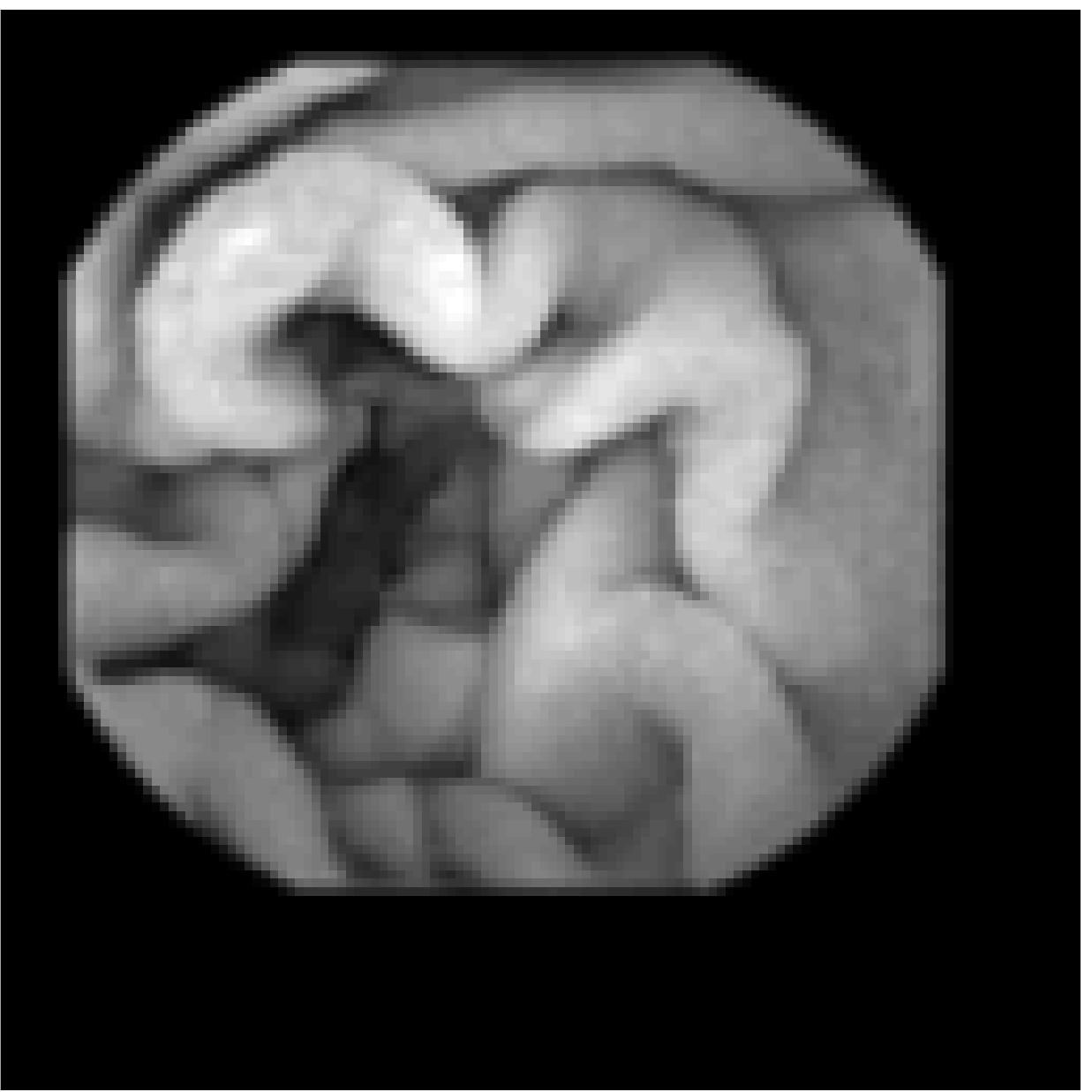}\hspace{2 mm}
\includegraphics[width=4.0cm,height=4.0cm]{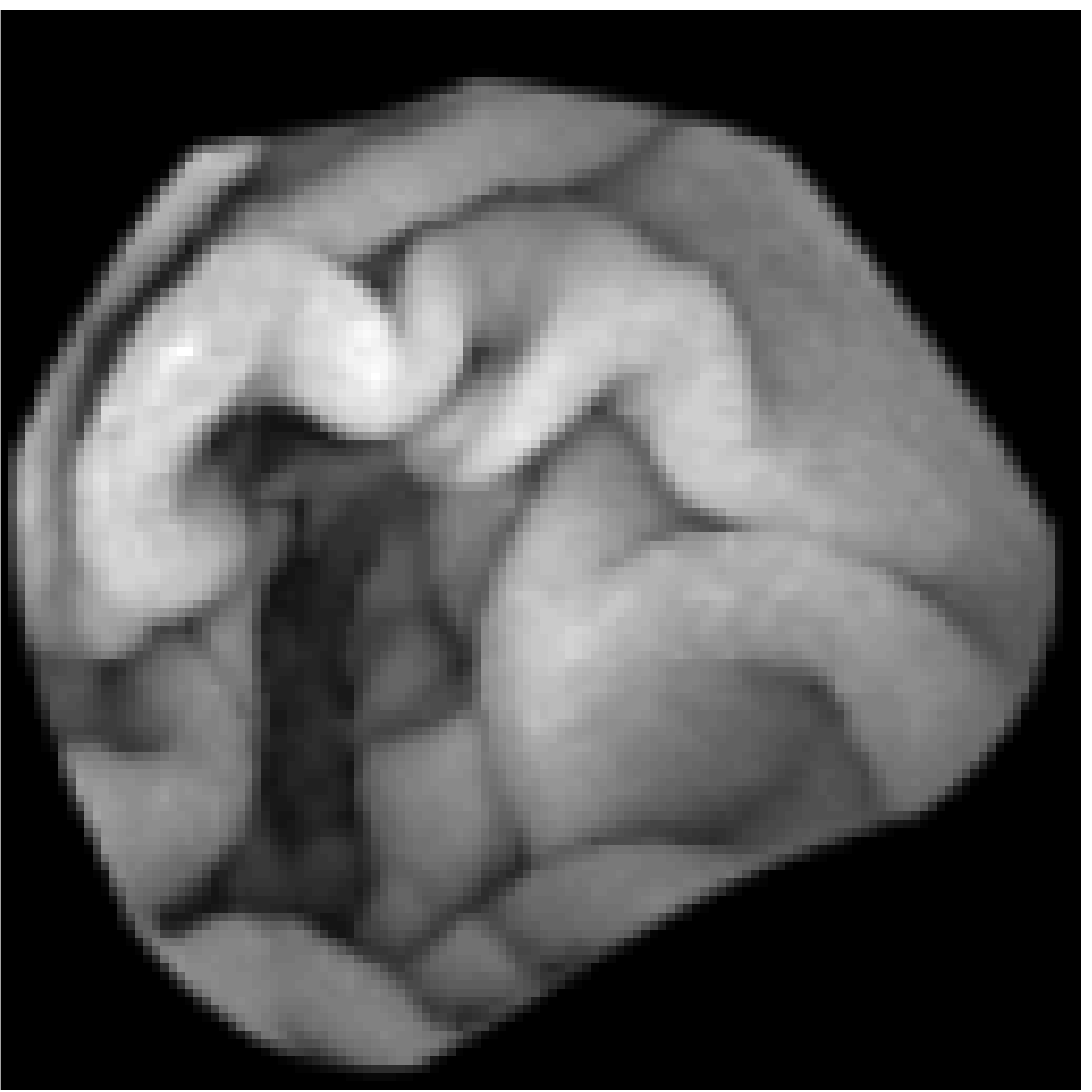}\\
\vspace{2 mm}
\includegraphics[width=4.0cm,height=4.0cm]{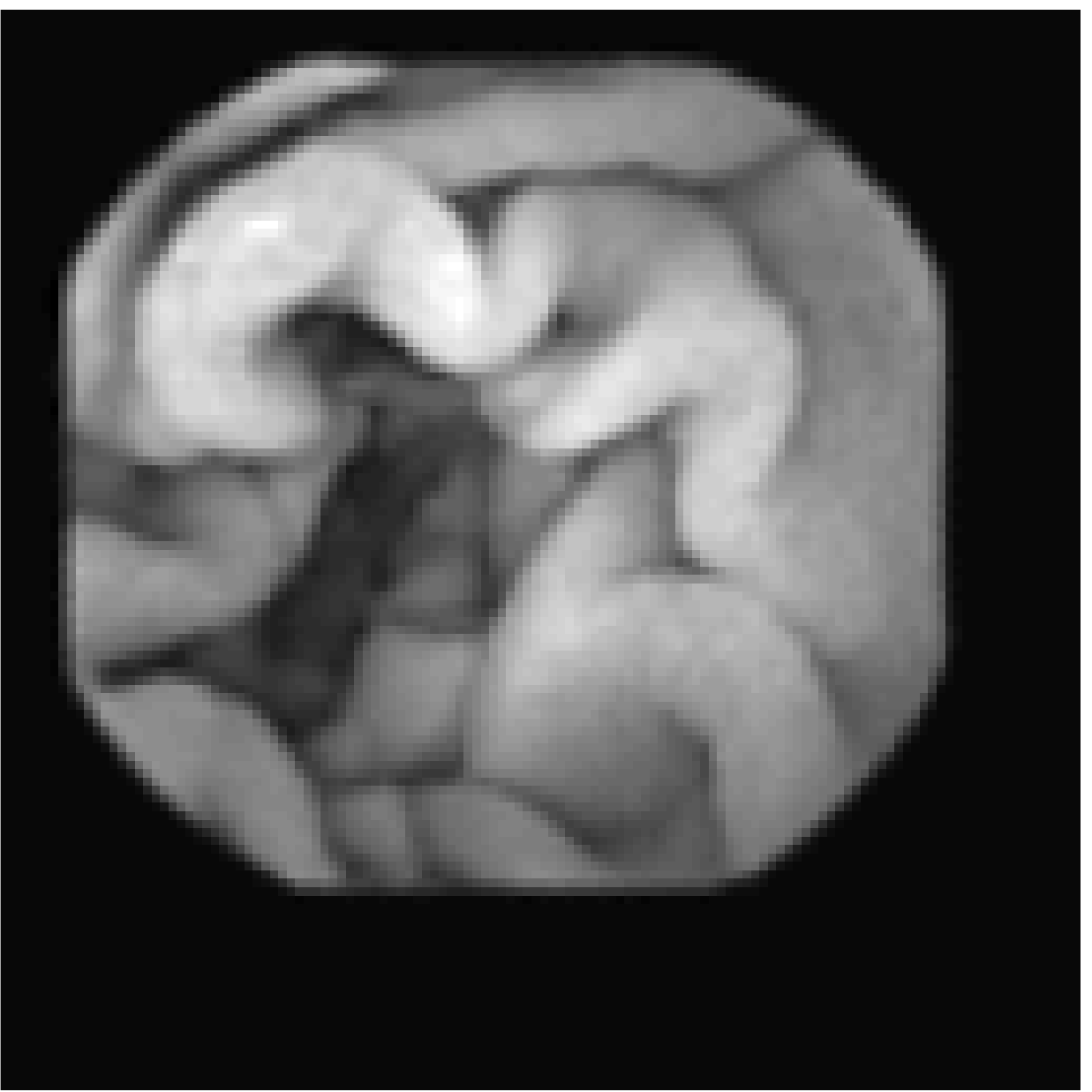}\hspace{2 mm}
\includegraphics[width=4.0cm,height=4.0cm]{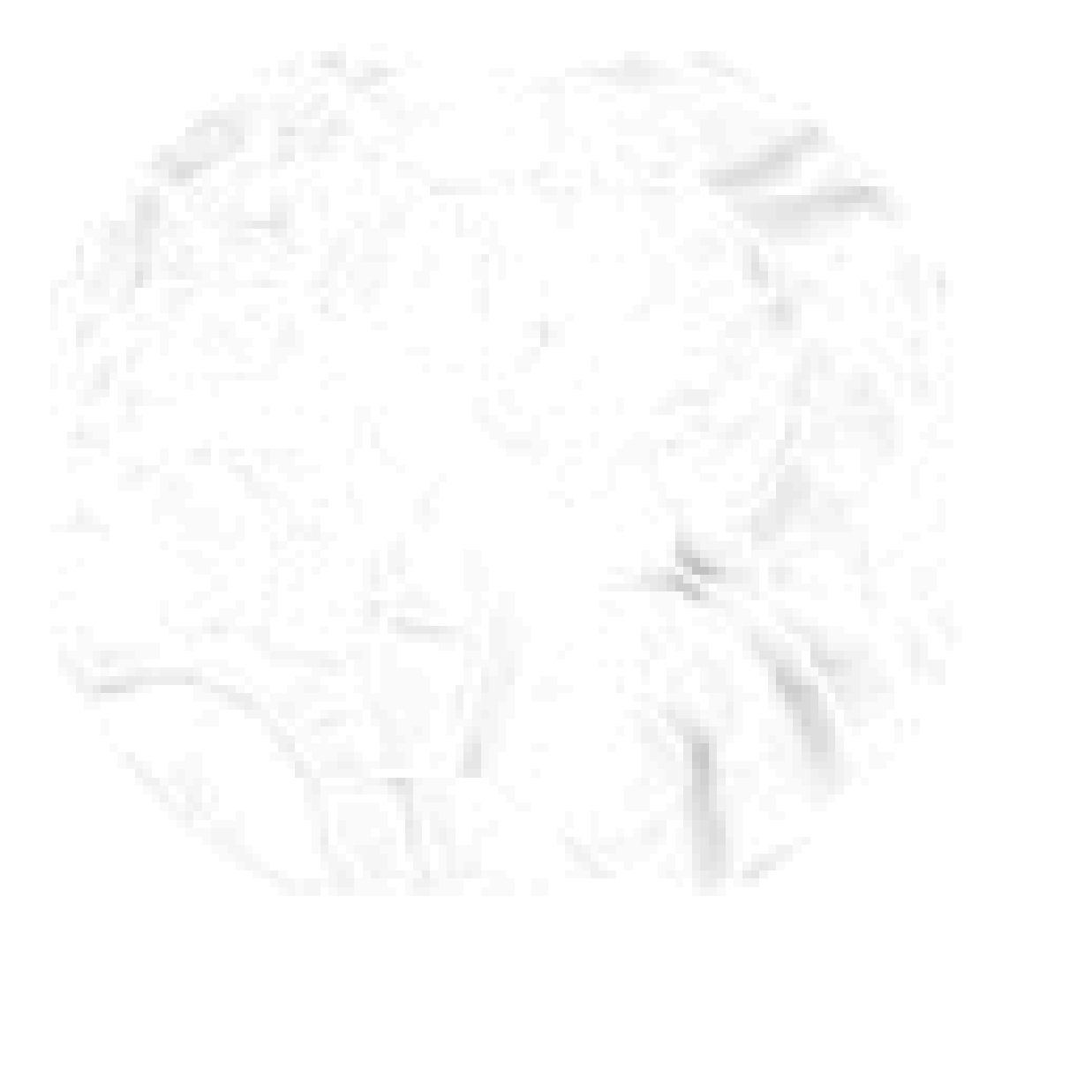}\hspace{2 mm}
\includegraphics[width=4.0cm,height=4.0cm]{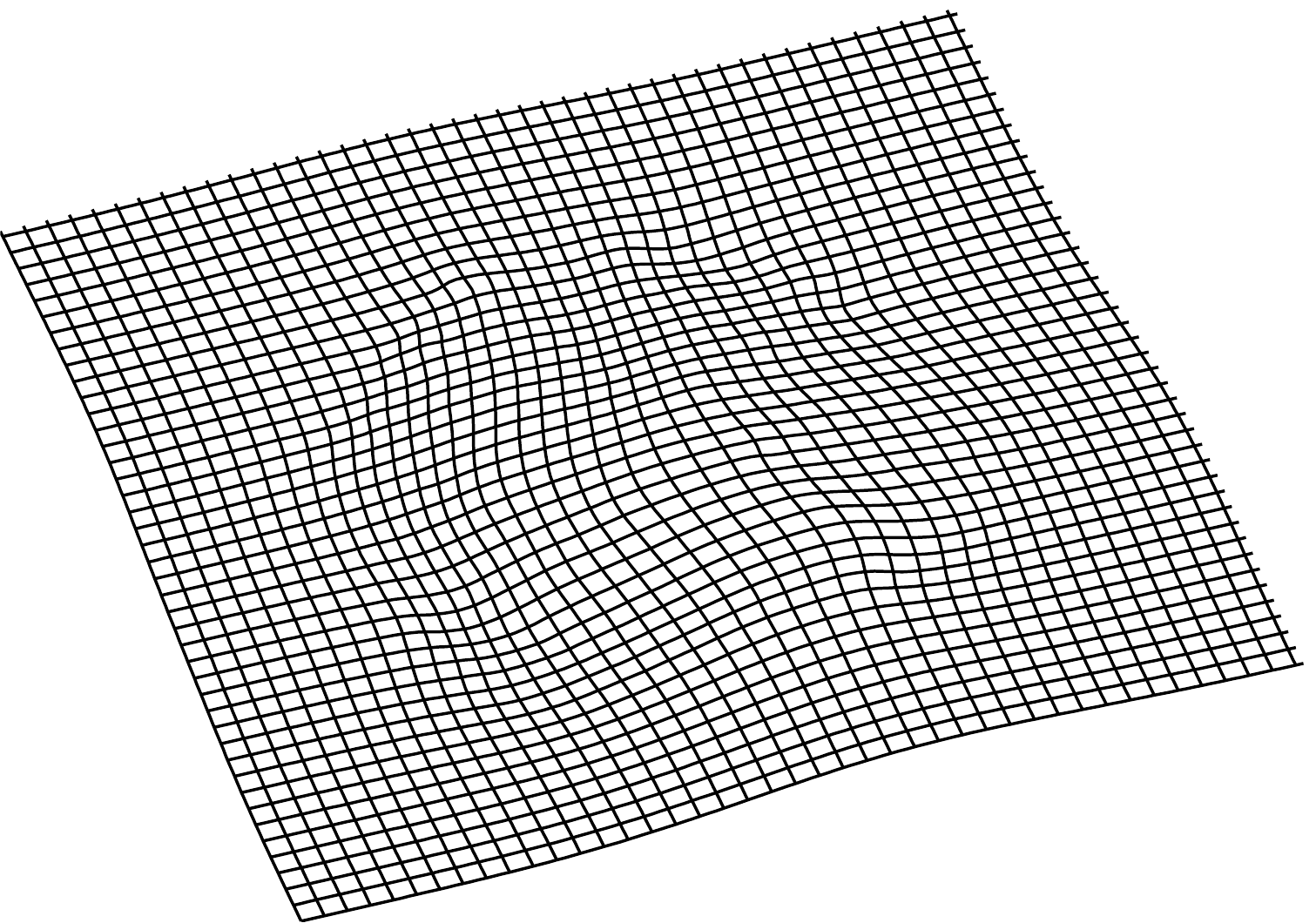}\\
\vspace{2 mm}
\includegraphics[width=4.0cm,height=4.0cm]{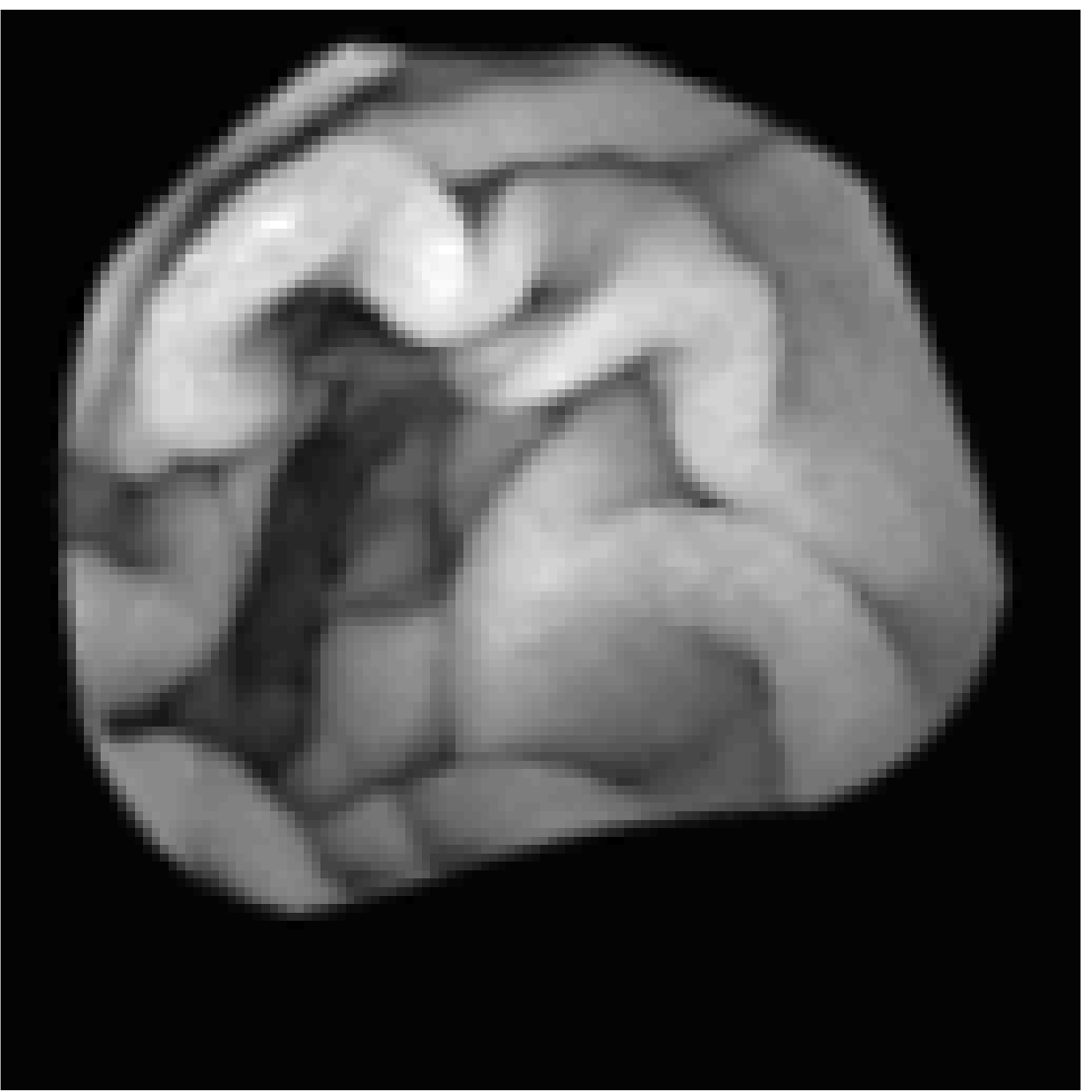}\hspace{2 mm}
\includegraphics[width=4.0cm,height=4.0cm]{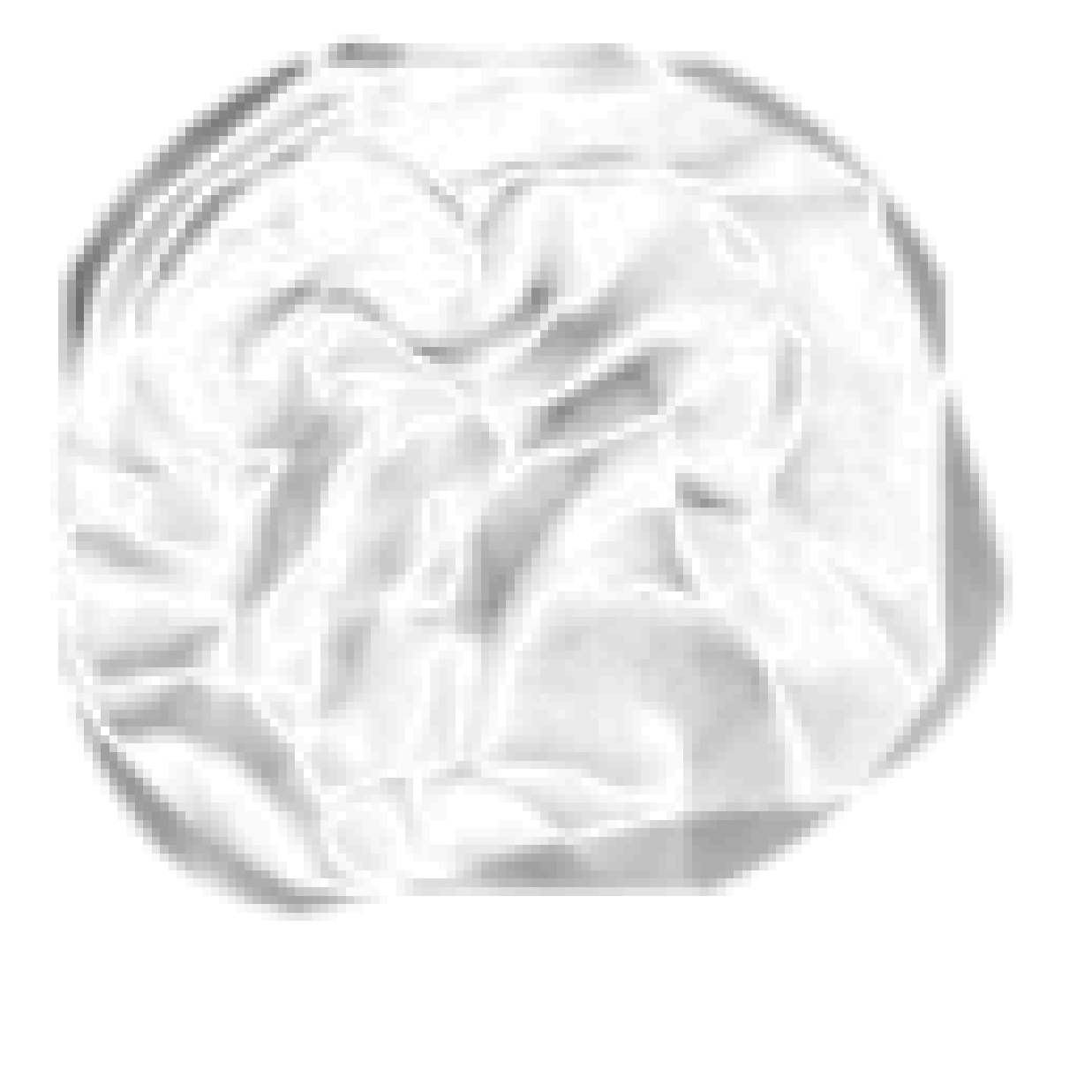}\hspace{2 mm}
\includegraphics[width=4.0cm,height=4.0cm]{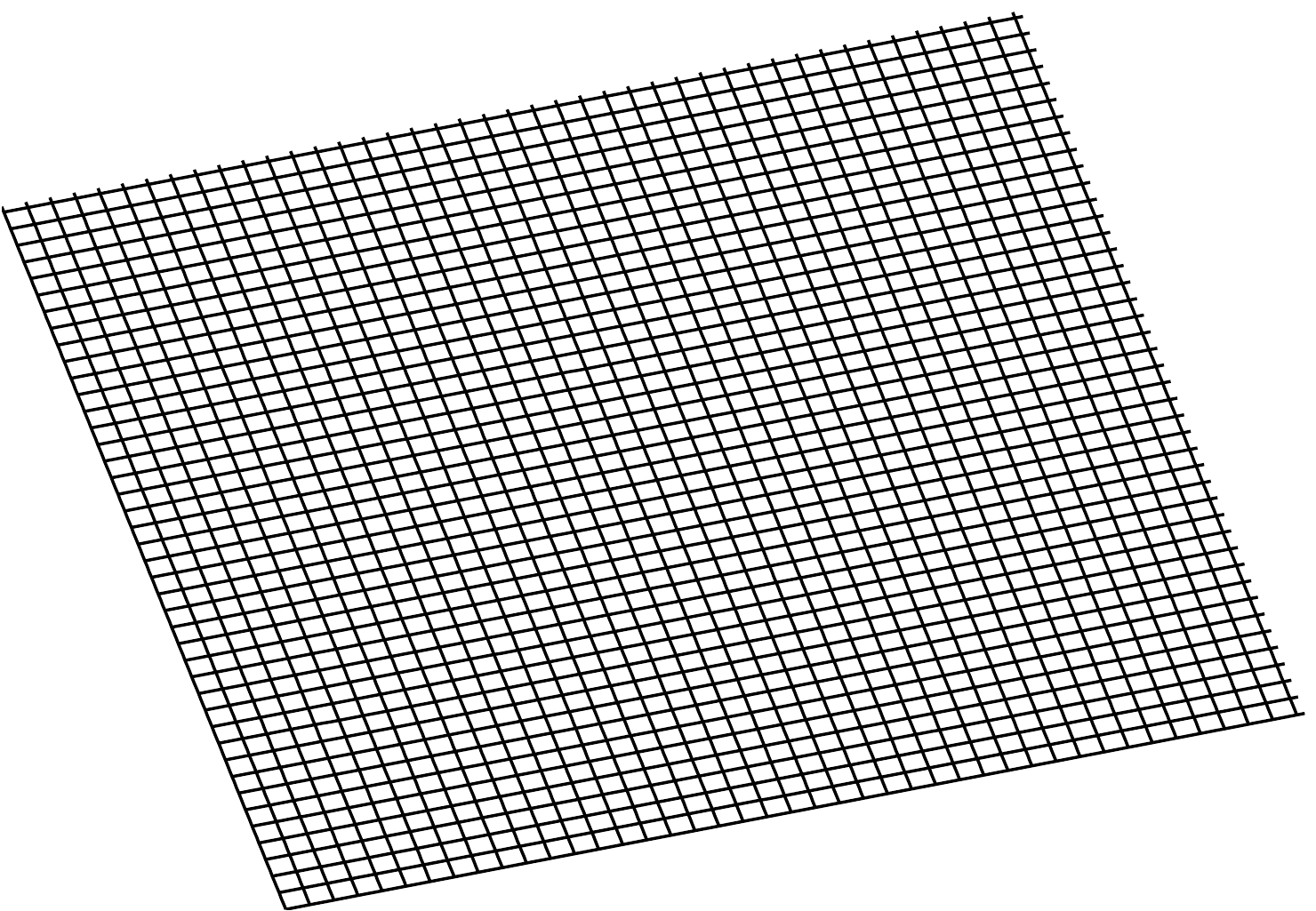}\\
\caption{First row (from left to right) : Original frame, grayscale reference $R$ and template $T$ ($T$ is a synthetic rotated and elastic deformed version  of $R$).
Second row (from left to right)
   MEIR results : $T(Id-u)$, difference between $R$ and $T(Id-u)$, transformation $Id-u$.
    Third row (from left to right)
   MPIR results : $T(\varphi)$, difference between $R$ and $T(\varphi)$, transformation $\varphi$.
   }\label{fig:1}
\end{figure}

In general an analytical solution to \eqref{eq:multisca} does not exist, and consequently the optimization problem \eqref{eq:multisca}  is then discretized  and gives rise to a finite dimensional problem. The numerical scheme used in this paper to solve the discretized version of \eqref{eq:multisca}  is a Gauss-Newton like method (with Armijo's line search), for which the starting guess is the solution of the registration problem  at the previous coarse scale $\theta_{i-1}$, that is,
$u_{\theta_{i-1}}$ solution of \eqref{eq:multisca} for $i\geq 1$, and   $\varphi$ the solution of the affine pre-registration \eqref{eq:opt} for  scale $\theta_0$.

Finally and for summarizing the MEIR approach consists in performing  firstly \eqref{eq:opt}, the affine registration  at a coarse scale,  and then  the multiscale elastic registration,  by solving \eqref{eq:multisca} for  each  scale (and using the solution of  each scale as the input for the next scale).

We note that in \eqref{eq:multisca}, if we consider  the regularizing parameter $\alpha=0$, and  search for an affine transformation $\varphi$  of the form \eqref{eq:rig} at each scale, then the proposed MEIR approach becomes a multiscale parametric (affine) image registration approach, hereafter denoted by MPIR.

We remark that in all the experiments described in Section \ref{sec:results} we further enrich the MEIR approach, by iterating it twice, and using the registered image as the input template for the second iterate. This means that the following
two  steps are performed.

\begin{itemize}

\item {\it Step 1} -  Registration of the pair $(R,T)$ with MEIR.

\item {\it Step 2} - Registration of the pair   $\big (R,T (Id-u^1)\big )$ with MEIR, where $u^1$ is the solution of  {\it Step 1}.

\item  The transformation which is the solution of the previous  {\it Step 2}, hereafter denoted by $u$, is the final result for the iterated MEIR.

\end{itemize}

 \begin{figure}[t!]
\centering
\includegraphics[width=3.5cm,height=3.5cm]{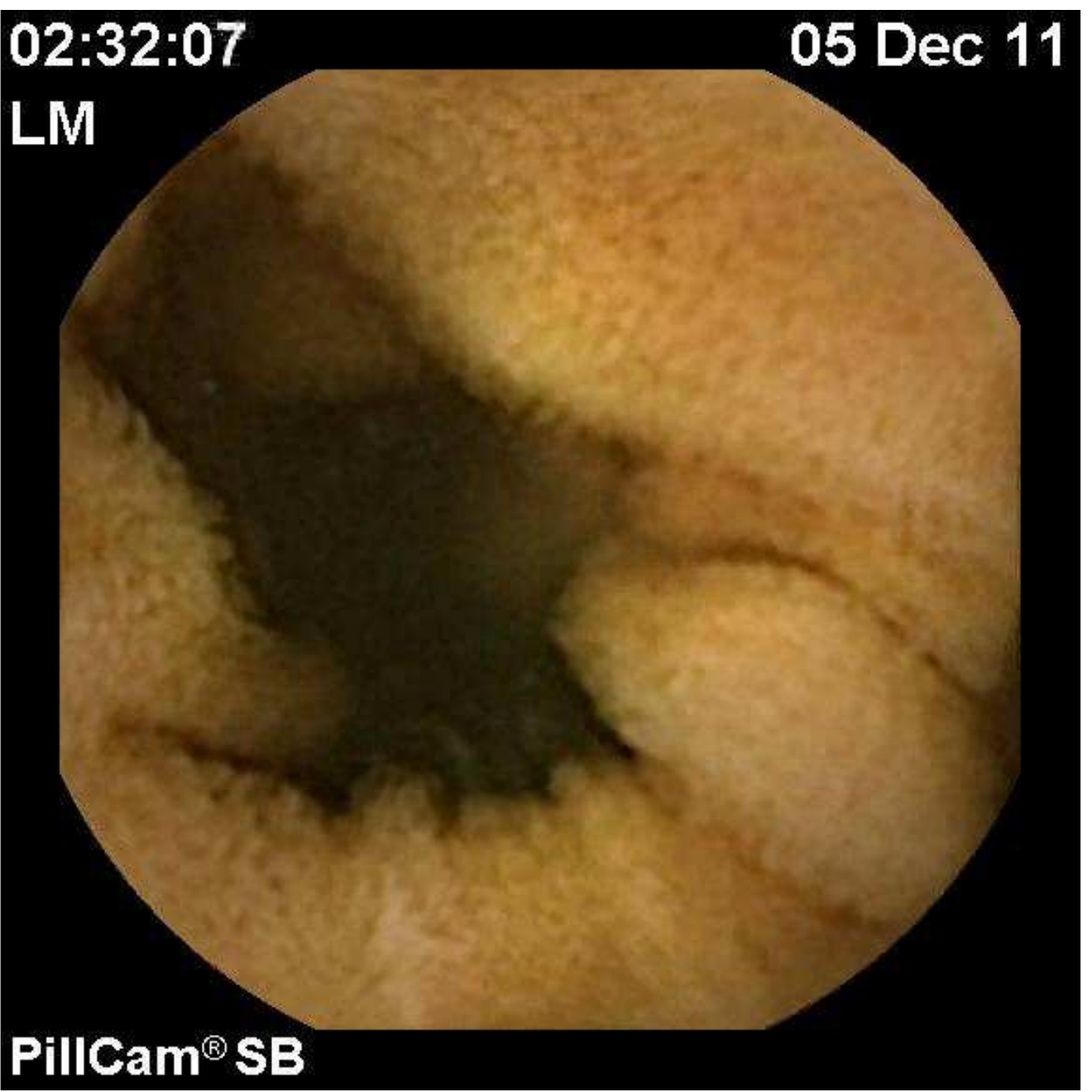}\hspace{2 mm}
\includegraphics[width=3.5cm,height=3.5cm]{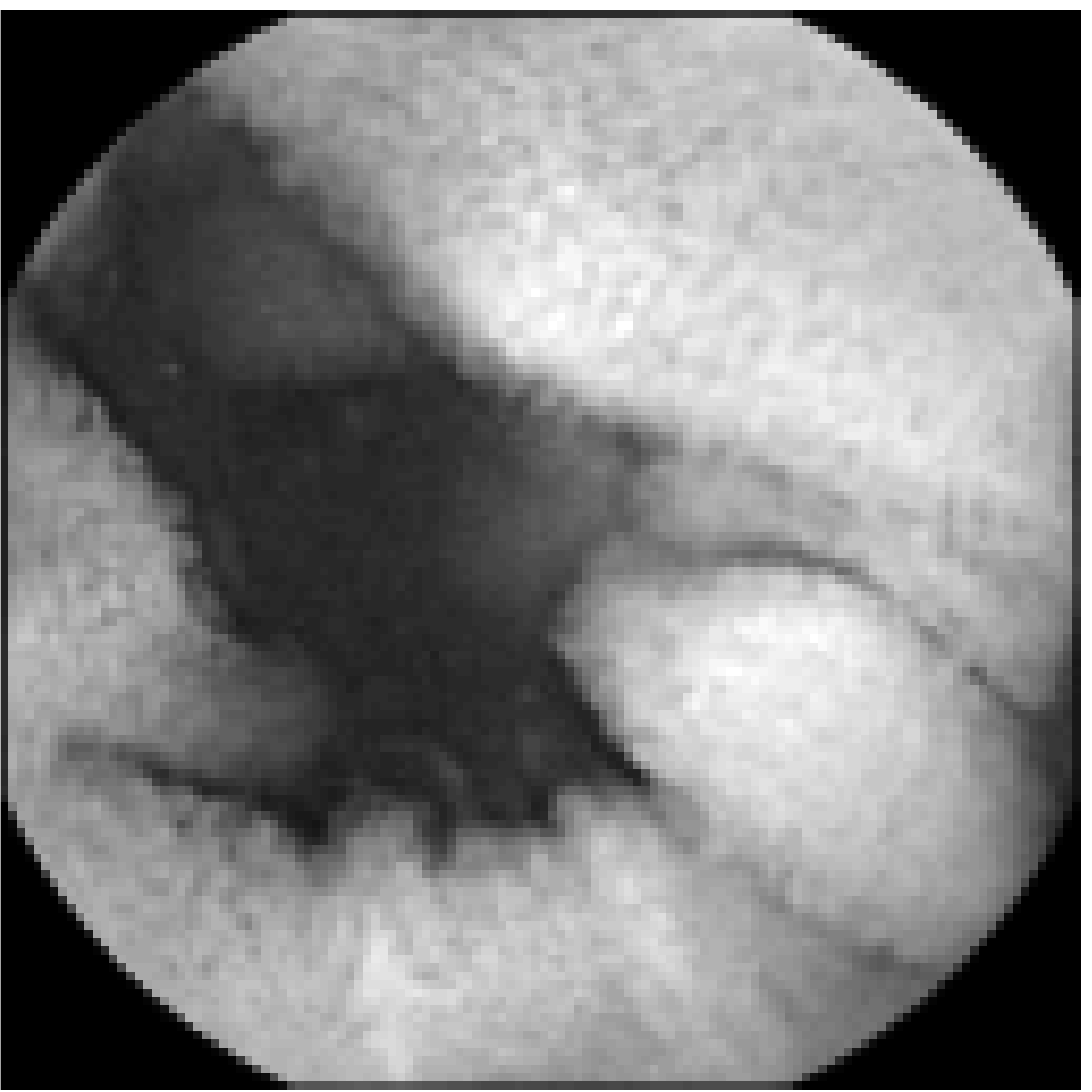}\hspace{2 mm}
\includegraphics[width=3.5cm,height=3.5cm]{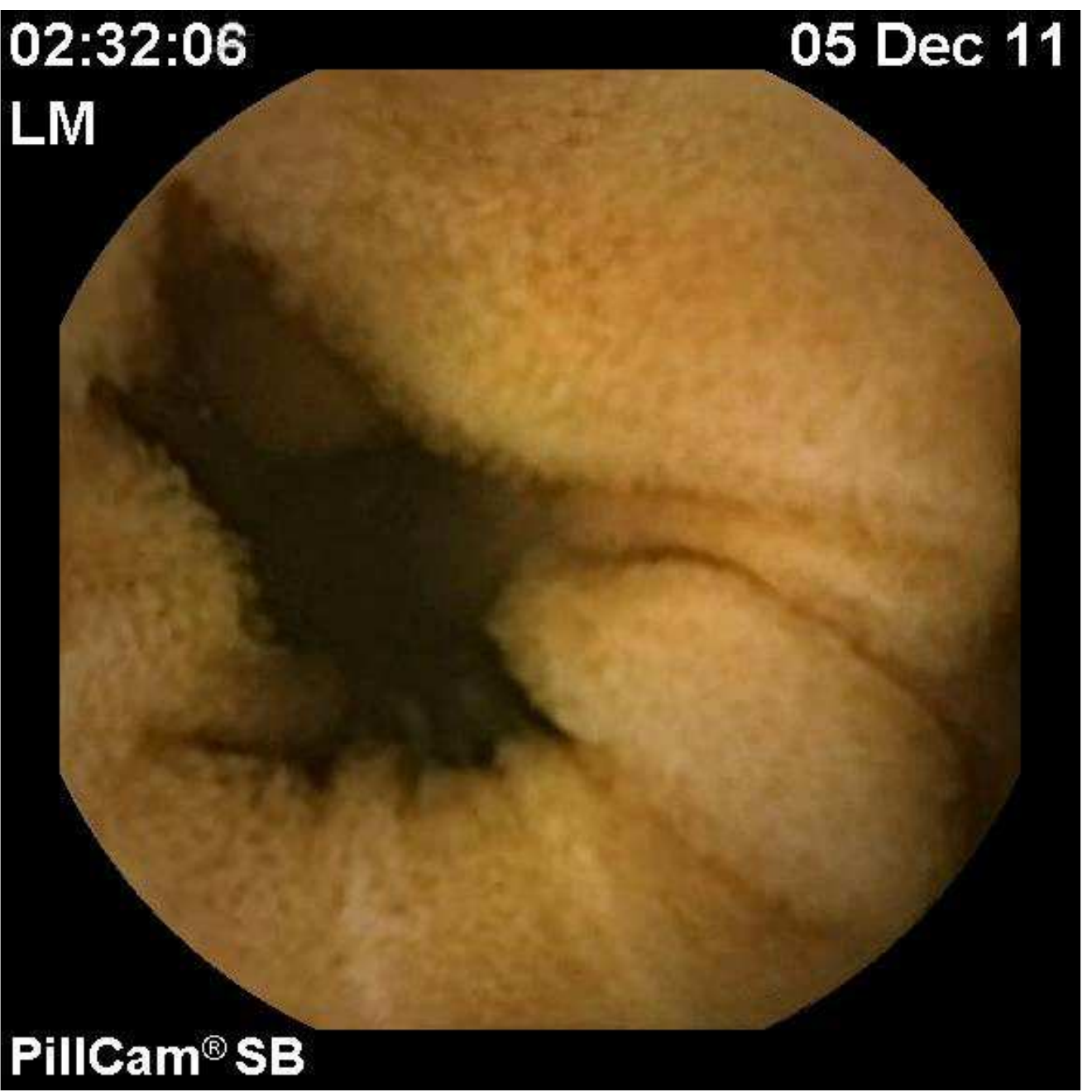}\hspace{2 mm}
\includegraphics[width=3.5cm,height=3.5cm]{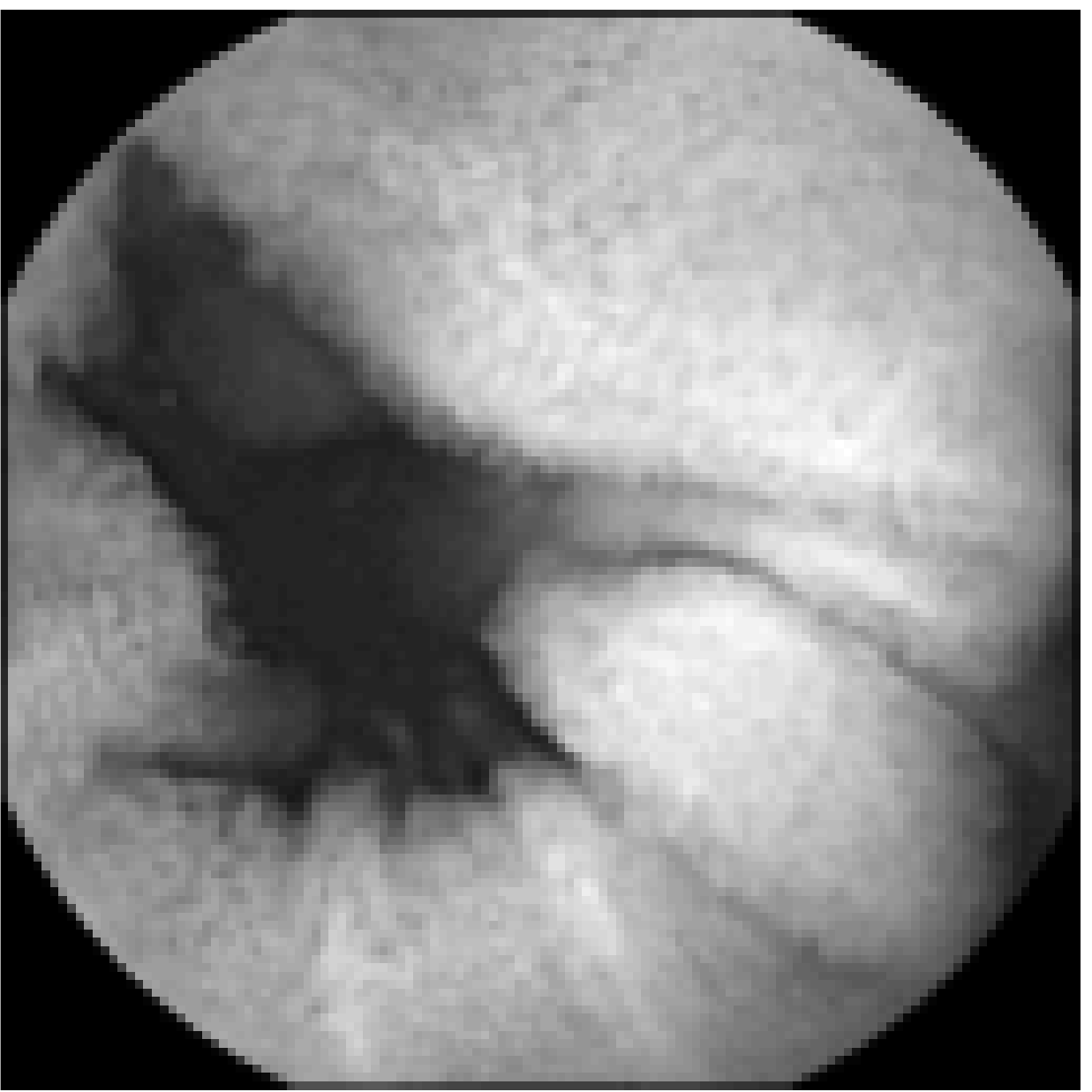}\\
\vspace{2 mm}
\includegraphics[width=3.5cm,height=3.5cm]{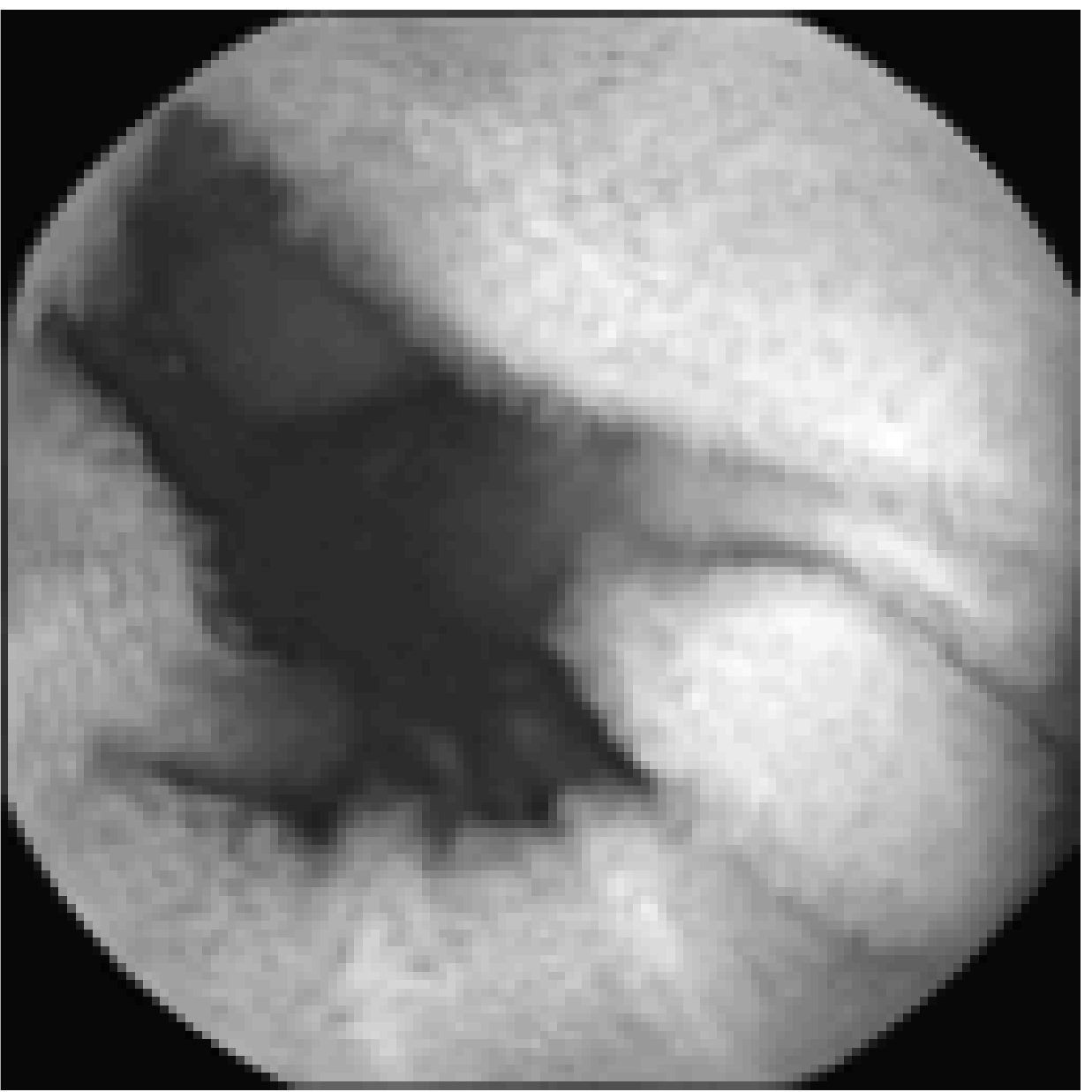}\hspace{2 mm}
\includegraphics[width=3.5cm,height=3.5cm]{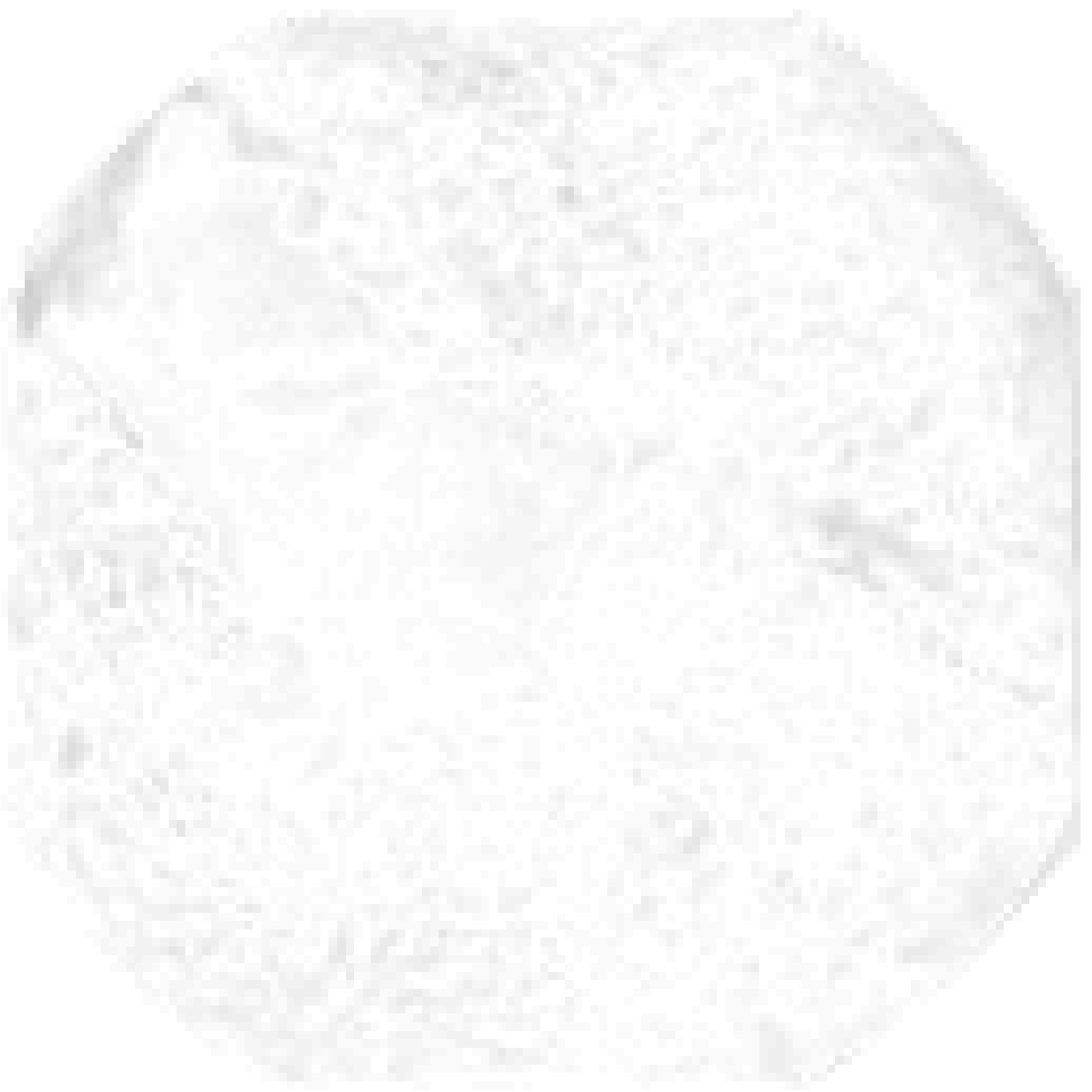}\hspace{2 mm}
\includegraphics[width=3.5cm,height=3.5cm]{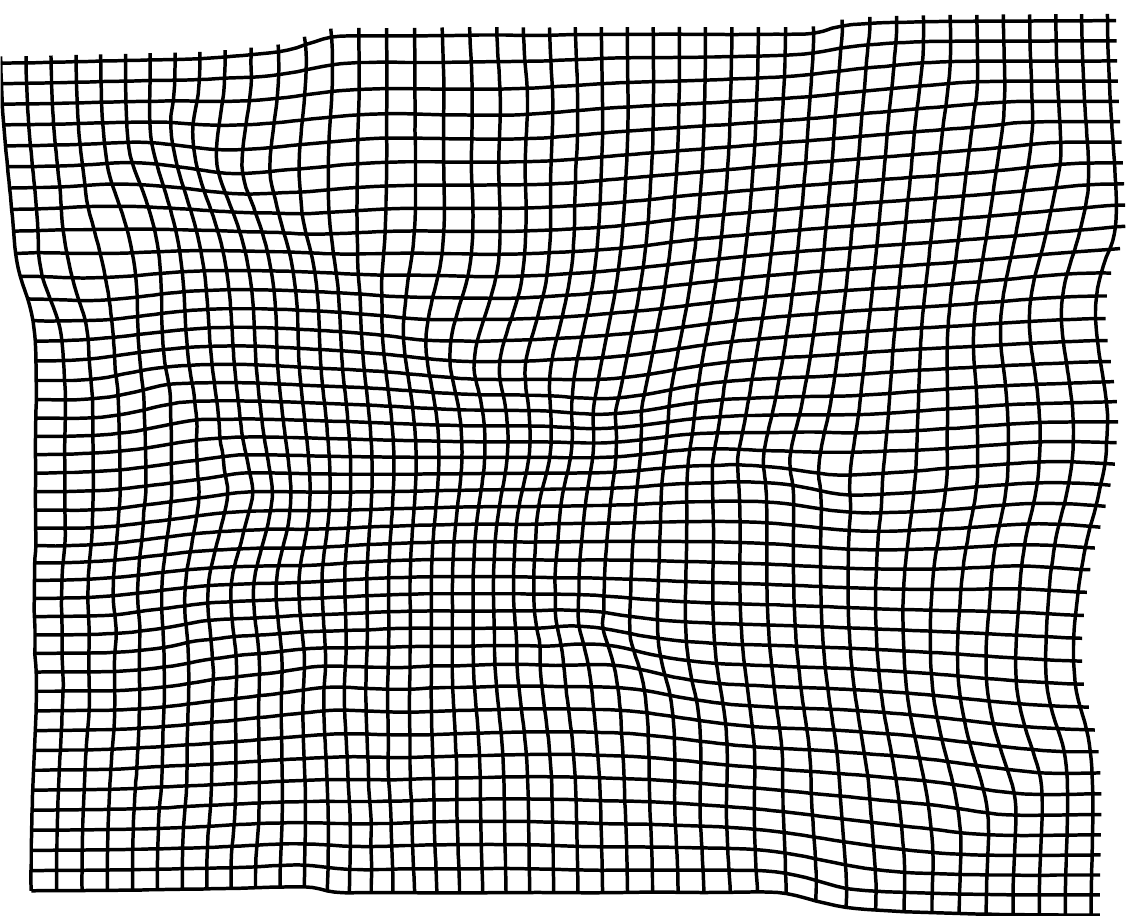}\\
\vspace{2 mm}
\includegraphics[width=3.5cm,height=3.5cm]{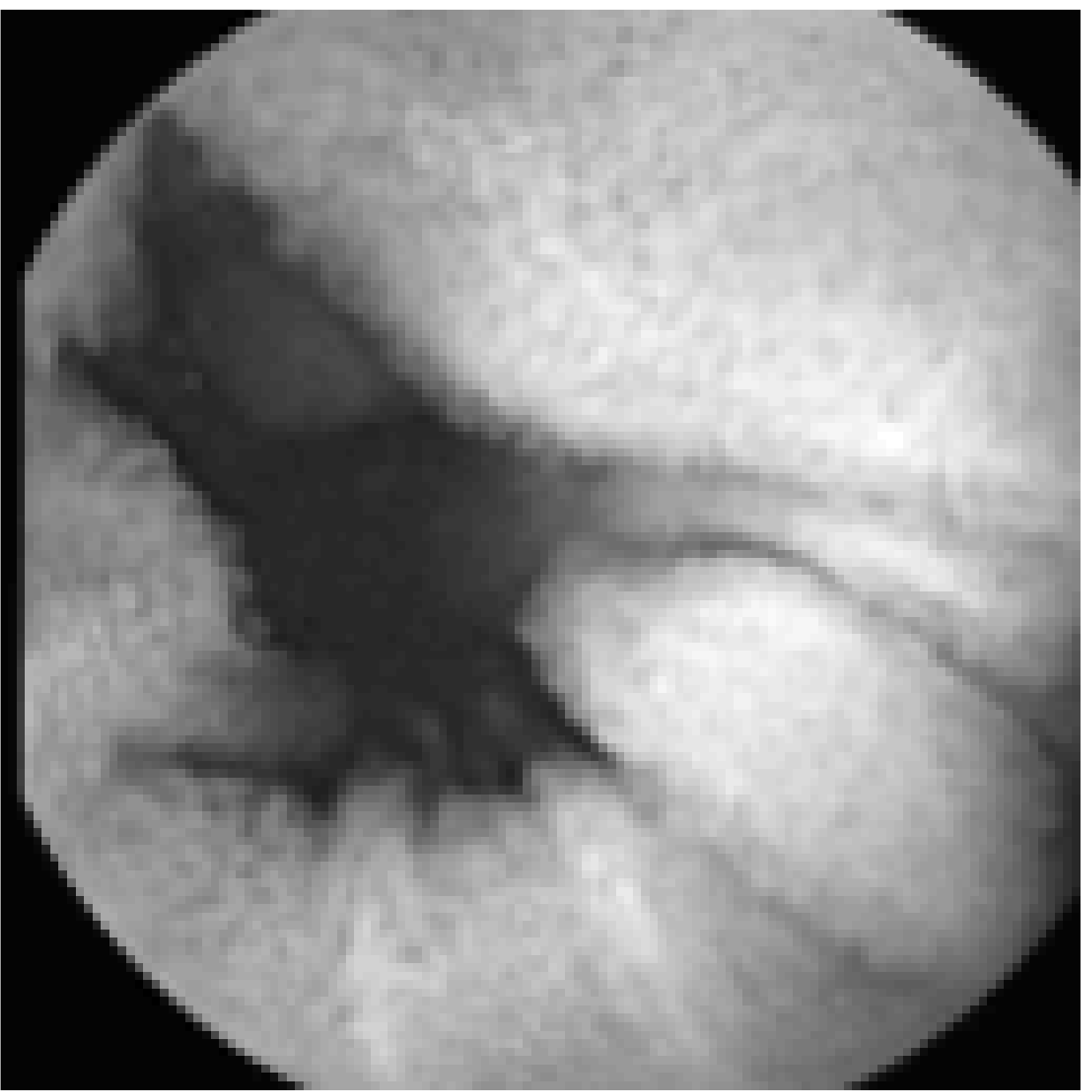}\hspace{2 mm}
\includegraphics[width=3.5cm,height=3.5cm]{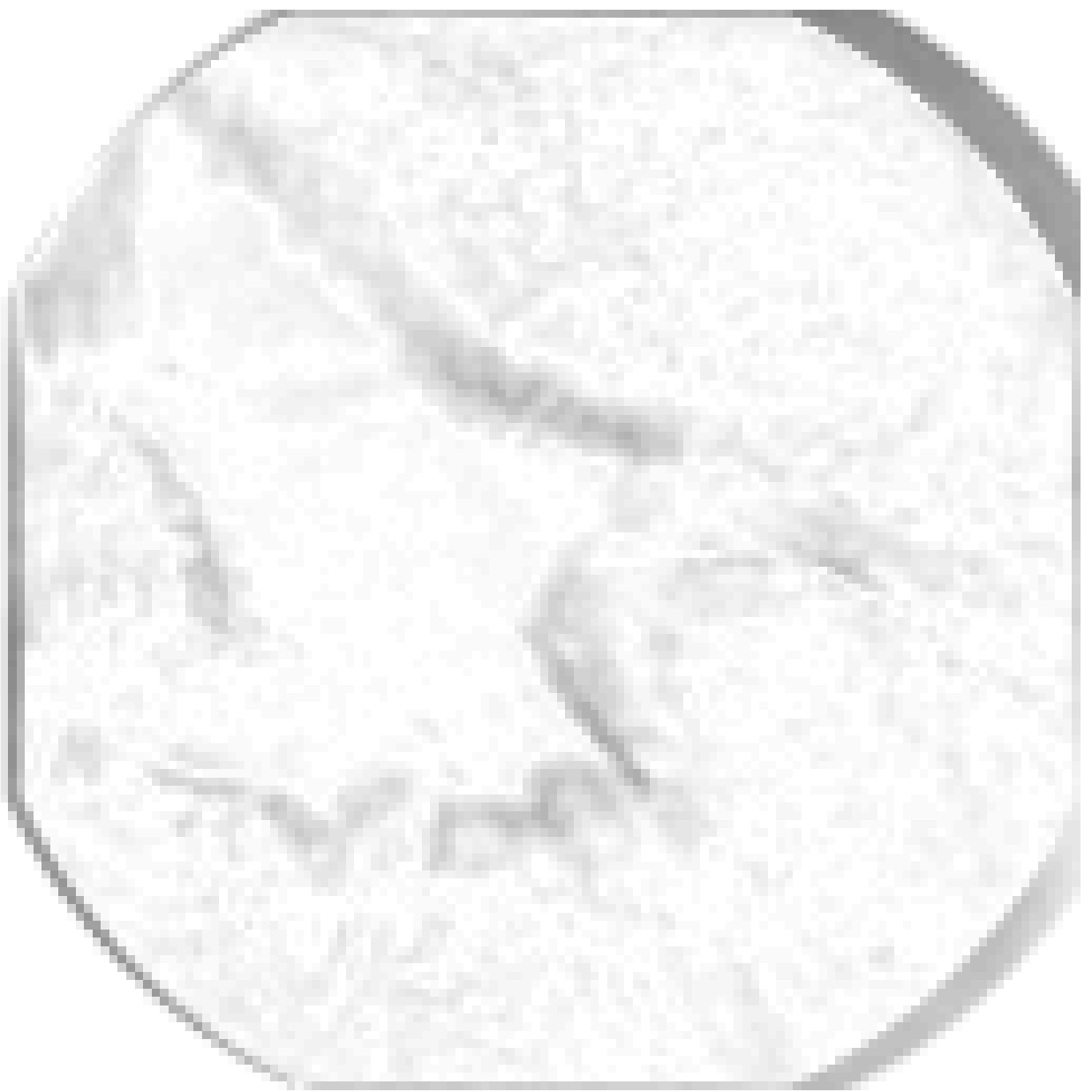}\hspace{2 mm}
\includegraphics[width=3.5cm,height=3.5cm]{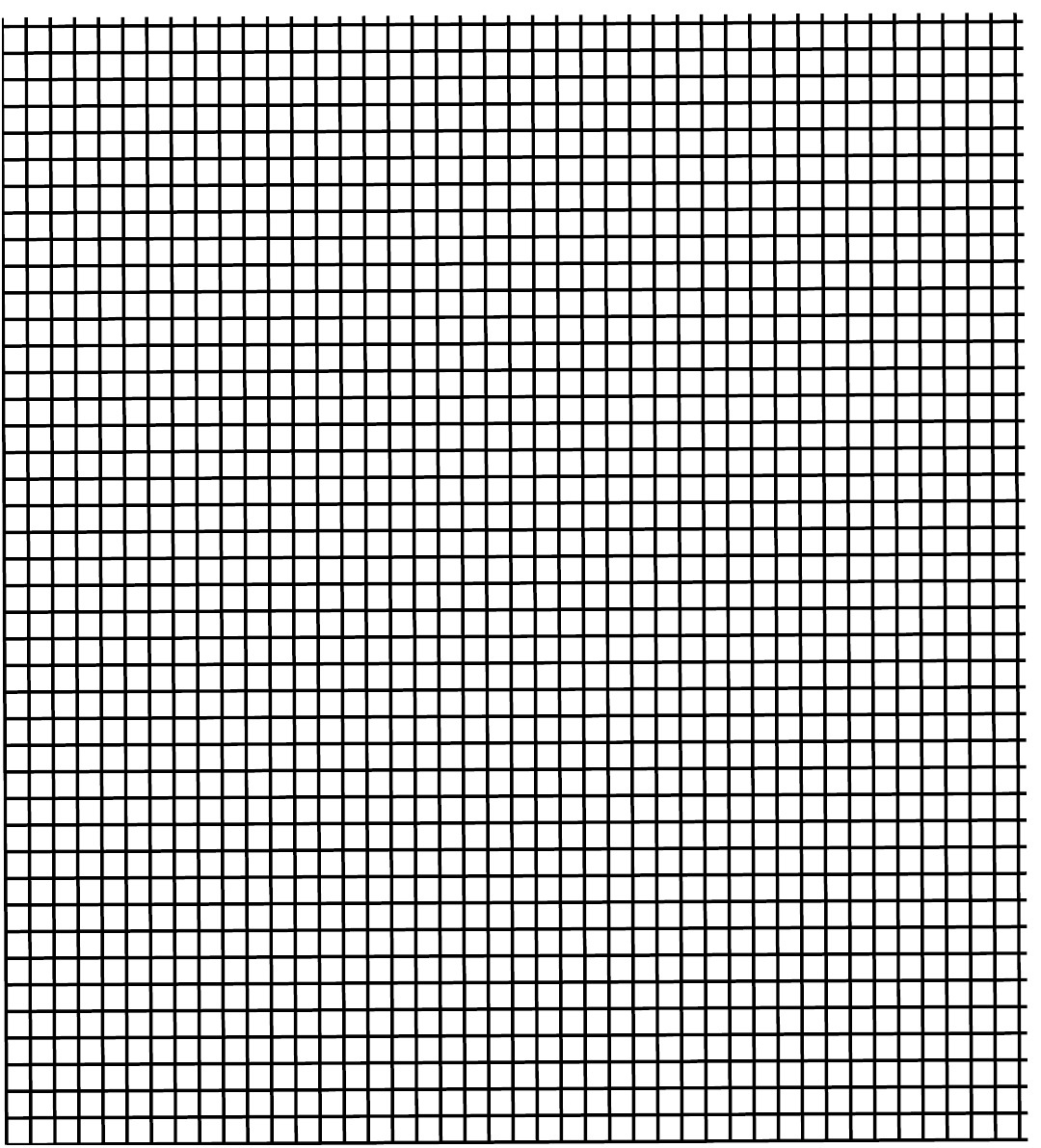}\\
\caption{First row (from left to right) :  original and grayscale  reference $R$  and original and grayscale template $T$ images ($T$ corresponds  to the   frame previous to $R$ in a WCE video).
Second row (from left to right)
   MEIR results: $T(Id-u)$, difference between $R$ and $T(Id-u)$, transformation $Id-u$.
   Third row (from left to right)
   MPIR results: $T(\varphi)$, difference between $R$ and $T(\varphi)$,  where $\varphi$ is the affine transformation   close to $Id-u$.}
\label{fig:2}
\end{figure}

The Figures \ref{fig:1}, \ref{fig:2} and \ref{fig:3}  illustrate  the results obtained with MEIR and MPIR,  for different pairs of images $(R,T)$, where $R$ is the reference and $T$ the template. We can visually compare in Figures  \ref{fig:1} and \ref{fig:2} the two registration approaches. In Figure  \ref{fig:1},  $T$ is a simulated  version  of $R$,   obtained by applying a rotation and an elastic deformation to $R$, and the result of MEIR, displayed in the second row, is clearly better than the MPIR result,  shown in third  row.
In  Figure  \ref{fig:2},  $R$ and $T$ are two consecutive frames of a WCE video: $R$ is the frame after $T$, in the video, and we can perceive an elastic deformation and a rotation in $R$.  Also in this case MEIR gives a better result than MPIR (compare the second and third  rows).
In Figure  \ref{fig:3},   $T$ is a rotated and scaled version of $R$, and the performance of both registration approaches are  visually very similar, that is the reason why we only show the results obtained with MEIR, and the MPIR results are omitted. Moreover in these three  figures the displayed grids for MEIR correspond to one iteration for MEIR; the grid obtained in the second iteration of MEIR only corrects minor differences.

 \begin{figure}[t!]
\centering
\includegraphics[width=4.0cm,height=4.0cm]{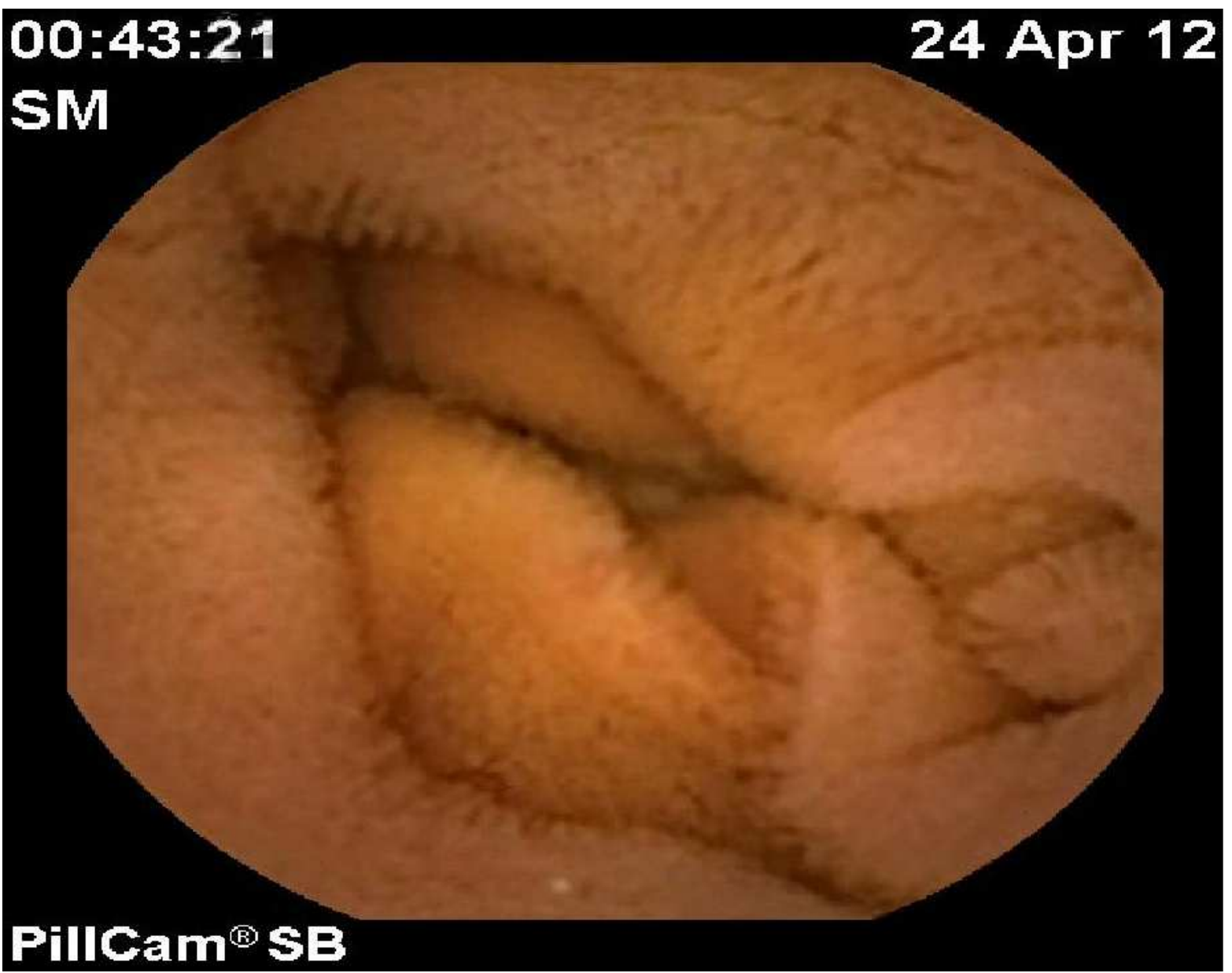}\hspace{2 mm}
\includegraphics[width=4.0cm,height=4.0cm]{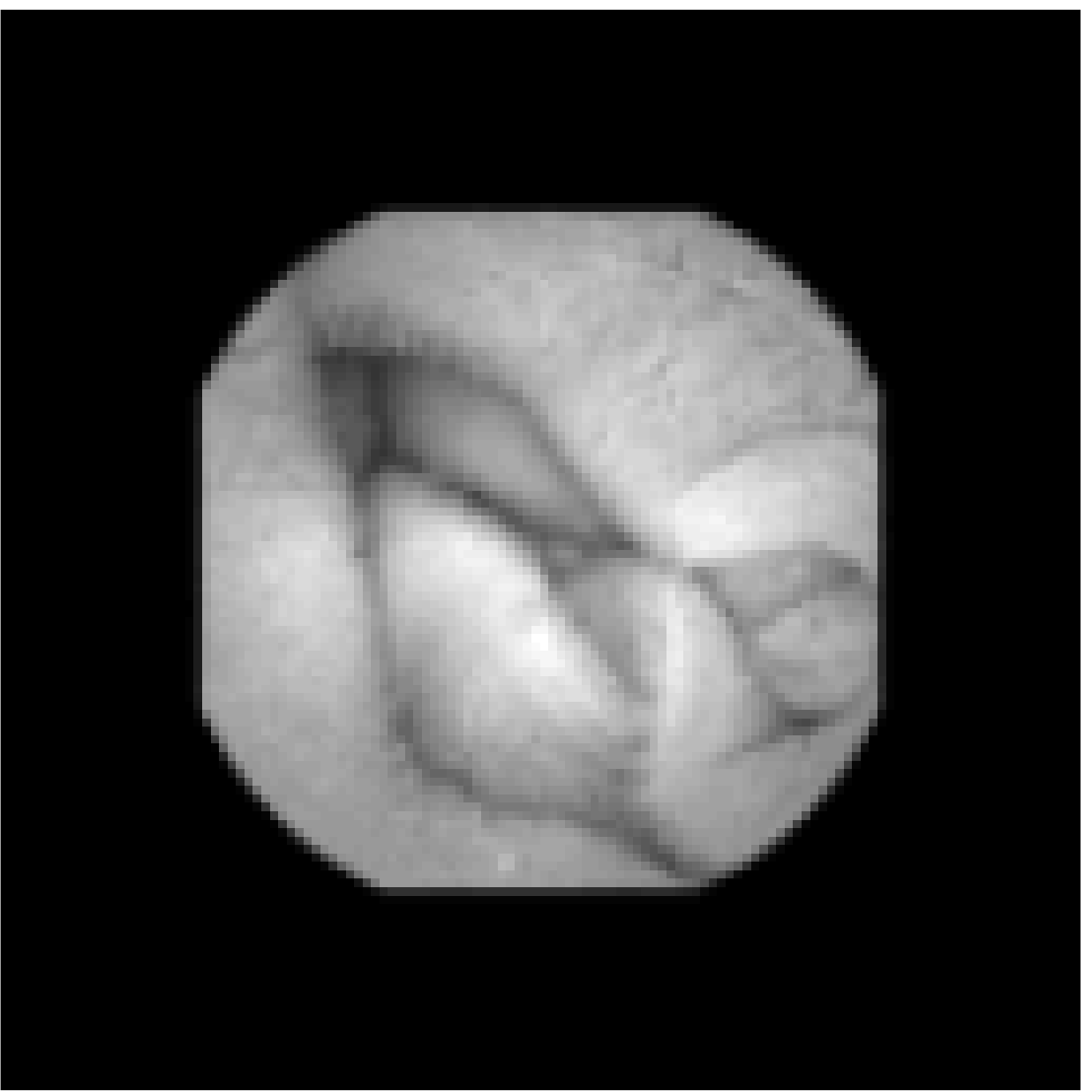}\hspace{2 mm}
\includegraphics[width=4.0cm,height=4.0cm]{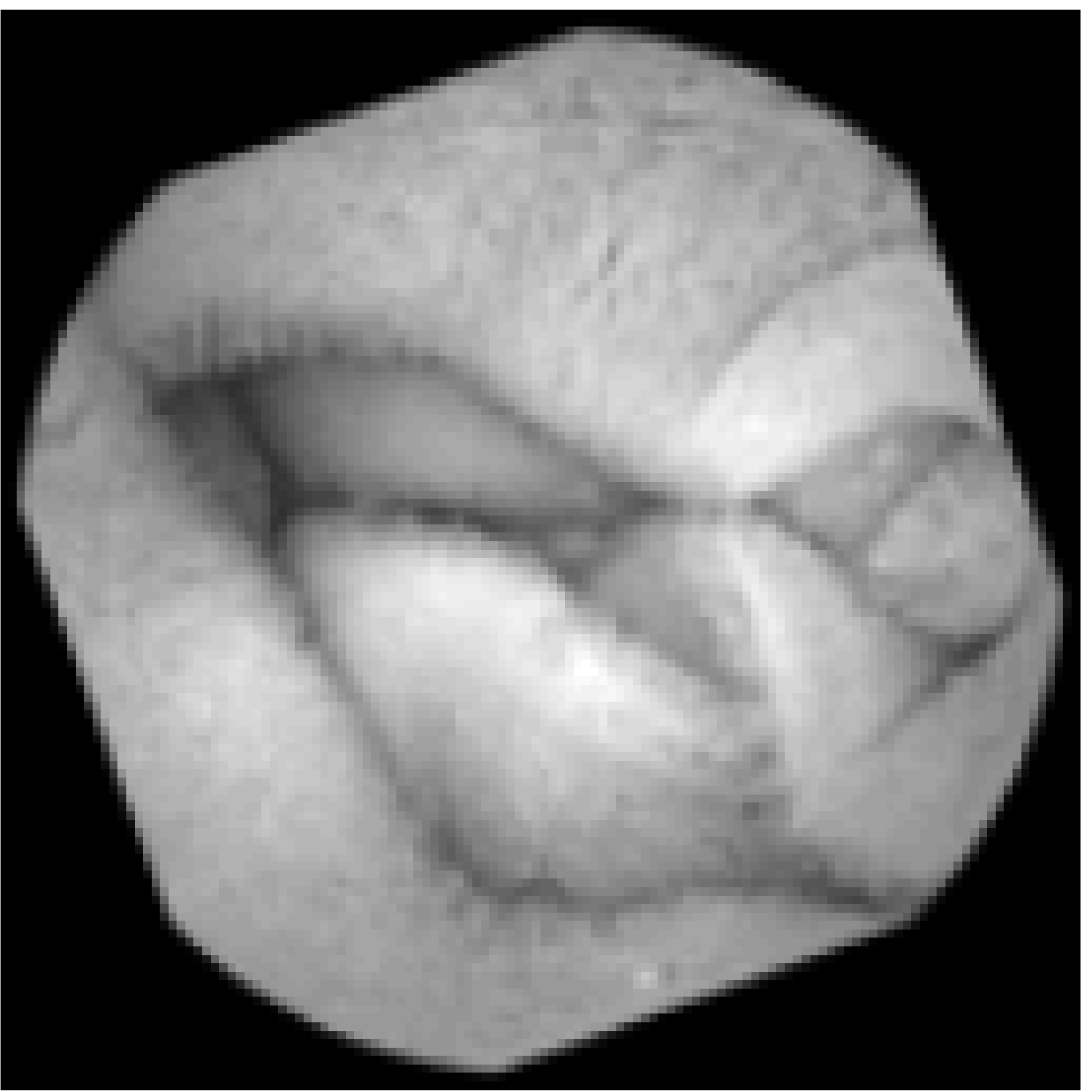}\\
\vspace{2 mm}
\includegraphics[width=4.0cm,height=4.0cm]{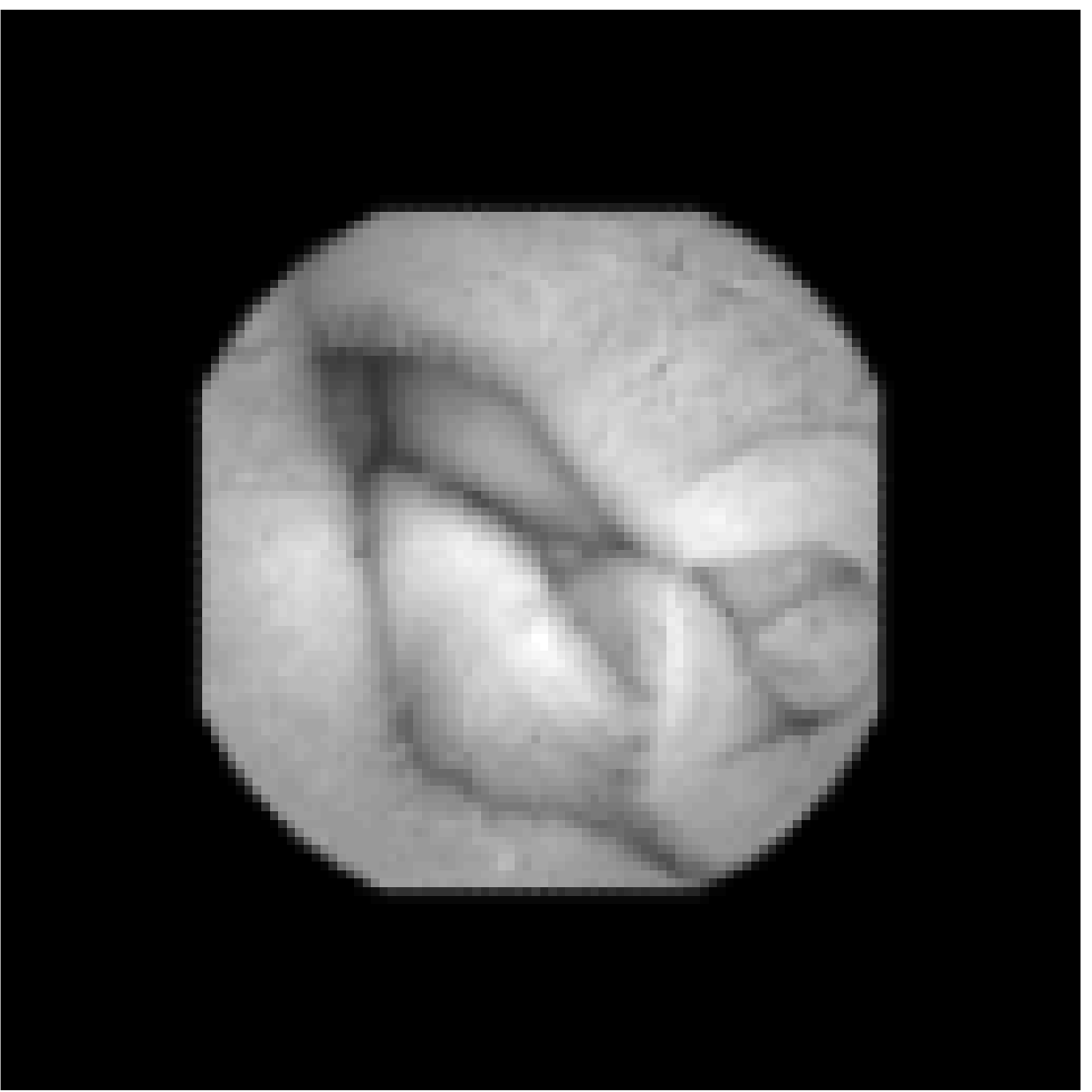}\hspace{10 mm}
\includegraphics[width=4.0cm,height=4.0cm]{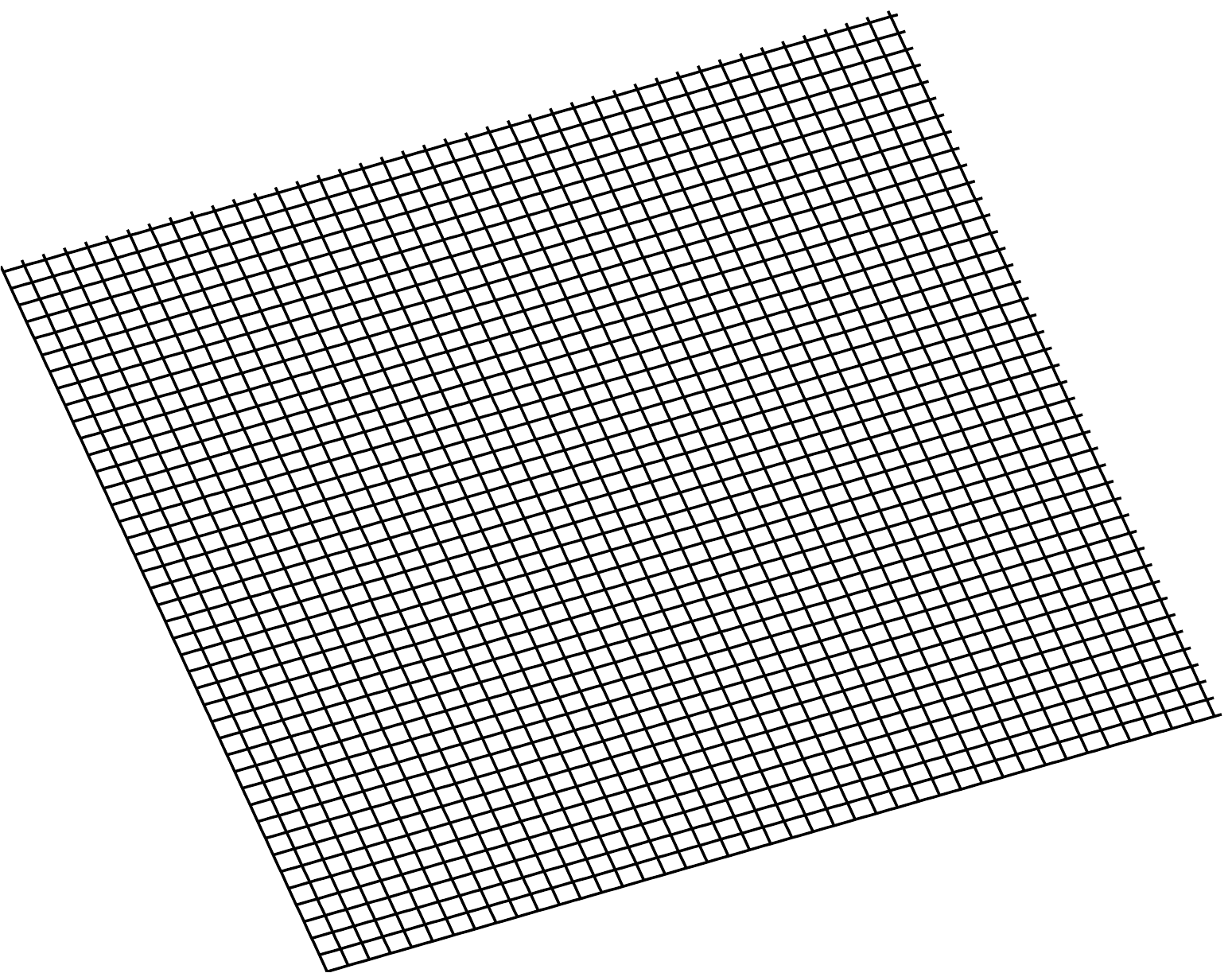}\\
 \caption{First row (from left to right) : original frame,  grayscale reference $R$  and template $T$ images ($T$ is an artificially  rotated  and scaled version  of $R$ - the rotation angle is 20 and scale factor is 1.4).
Second row (from left to right)
   MEIR results: transformed template $T(Id -u) $,  transformation $Id -u$.  }\label{fig:3}
\end{figure}

We can also quantitatively  compare the results obtained with MEIR and MPIR, displayed in  Figures \ref{fig:1} and \ref{fig:3}, where the template image $T$ is a  simulated version  of the reference image  $R$, by computing the following normalized dissimilarity measure ($NDM$)
\begin{equation}\label{eq:decision}
 NDM    :=   \displaystyle  \frac{\|   T(\varphi) -  R \|_{L^2(\Omega)}}{\|   R\|_{L^2(\Omega)}}  .
\end{equation}
This measure evaluates the accuracy of the registration approach. Here $\varphi$ denotes the final numerical solution of the registration  process ($\varphi $ of the form \eqref{eq:rig} for MPIR and $\varphi=Id-u$ for MEIR), and
 $L^2(\Omega)$  denotes the space of square-integrable  functions  in $\Omega$.
 We observe that the measure $NDM$ quantifies the  similarity between the reference and transformed template images in the norm of $L^2(\Omega)$, normalized by  the  $L^2(\Omega)$ norm  of the reference image.
  Clearly,  for Figures  \ref{fig:1} and \ref{fig:3}, where $T$ is a  simulated version  of $R$,  the smaller $ NDM$  is, the  more  accurate is the
  registration approach. In Figure  \ref{fig:1}  we have that $ NDM=0.033455$ for MEIR and $ NDM=0.390690$ for MPIR, and in  Figure  \ref{fig:3}  we have that $NDM=0.012473$ for MEIR and $NDM= 0.019216$ for MPIR. So in  Figure  \ref{fig:1} MEIR has a better performance than MPIR and in  Figure  \ref{fig:3} the results of both approaches resemble  each other closely.

 \section{Experiments, Results and Analysis}\label{sec:results}

We have evaluated the two multiscale registration approaches on 39 WCE videos, recorded at the Department of Gastroenterology of Coimbra Hospital ({\it CHUC - Centro Hospitalar e Universit\'ario de Coimbra, Portugal}).  The videos were acquired with the capsule {\it PillCam SB}, a WCE for the small bowel, manufactured by {\it Given Imaging}, Yoqneam, Israel.
Each video clip has the duration of 20 seconds and 100 frames.  Each frame has a resolution of $576\times 576$ pixels. The 39 videos belong to 9 different patients.

All the experiments were implemented with the software {\sc MATLAB{\circledR}} R2013b (The Mathworks, Inc.)  and we have also used  {\it FAIR Software} \cite{modersitzki2009fair},  an image registration  package written in MATLAB, that can be freely downloaded from {\it www.siam.org/books/fa06}.

We have performed two types  of experiments. Firstly we use real consecutive images of WCE videos, for showing the potential of the proposed MEIR approach. Secondly,  since it is difficult to validate, at the moment, the approach in human bodies,  we consider artificially scaled, rotated and elastic transformations of video frames, for demonstrating the efficacy of the proposed MEIR approach and  for evidencing  its superiority with respect to the MPIR approach,  when elastic deformations are involved.

In the numerical tests, for both MEIR and MPIR  we identify the image domain with the set $\Omega =[0,1]\times [0,1]$,  and discretize it  with $128\times 128 = 2^7\times 2^7$  points for both the template and reference images, in each scale scale, thus creating a regular grid. We also consider four scales $\theta=[100, 10, 1, 0]$. Morevover,  in MEIR the value  for
 the regularization parameter  is $\alpha =10$, and  for the elasticity parameters the values are  $\lambda=0$,  $\mu=1$.

We also note, as it  can seen for example in Figures \ref{fig:1} and \ref{fig:3} (first row), for generating the synthetic frames, before applying the (scaled, rotated or elastic) transformation  the original  grayscale frame   is padded with zeros such that its artificial version  is still inside the domain  $\Omega =[0,1]^2$. In addition, for all the tests the $NDM$ is always computed in the domain $[0,1]^2$ and not in a sub-region.

 \subsection{Experiments with real successive frames}\label{sec:testreal}

In this section we describe several  results obtained in the experiments performed with real successive frames, namely the results in terms of the normalized dissimilarity measure $NDM$ for computing an  estimation of the WCE speed.

The Figure \ref{fig:speed1} shows (in the middle) the plot  of  the $NDM$ curve for the MEIR approach, for a WCE video clip
with 100 frames and with the duration of $20$ seconds. In the same fashion as is done  in \cite{bao2012modeling}, this  curve can thus be understood  as  a qualitative  capsule speed information, that is  based on the similarity between consecutive frames. We remark as well  that each  video frame  has  the information concerning its time acquisition, thus there is a direct correspondence between the frame number, that belongs to  the interval $[1,100]$, and its acquisition time, that belongs to the interval $[0,20]$ in seconds.
Low values for $NDM$ indicate similarity between frames (for example,  for the pair of frames 12 and 13  displayed on the left of Figure \ref{fig:speed1},   the corresponding point in the  $NDM$ curve is $(12, 0.05523)$), so the capsule is almost still or rotates/moves slowly, while high  values for $NDM$ indicate abrupt changes/dissimilarities  in the  corresponding consecutive frames  (for instance to the pair of frames 51 and 52,  shown on the right of  Figure \ref{fig:speed1}, it  corresponds the point  $(51,0.37815)$ in the  $NDM$ curve)  revealing that the capsule is moving fast.
In particular, we refer that  from the medical point of view the parts of a video with sudden changes of image content  are of special interest. Therefore the $NDM$ can help clinicians in identifying quickly  these changes (corresponding to the $NDM$ peak values) as well as the other parts with slow motion (corresponding to low $NDM$ values).

\begin{figure}[t!]
\centering
\begin{minipage}[c][7.5cm][c]{0.23\textwidth}
\centering
  \includegraphics[width=3.5cm,height=3.5cm]{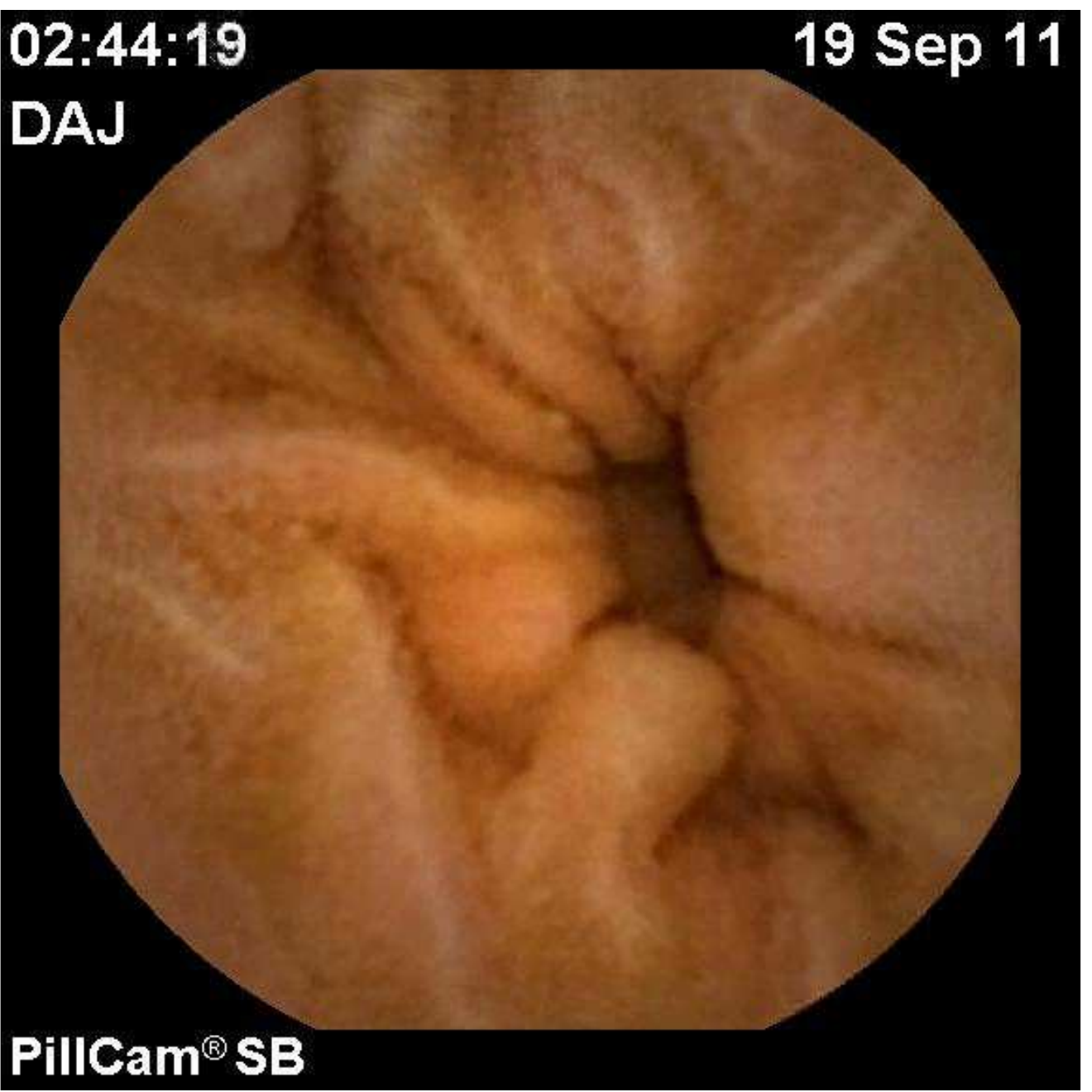}\par \vfill
  \includegraphics[width=3.5cm,height=3.5cm]{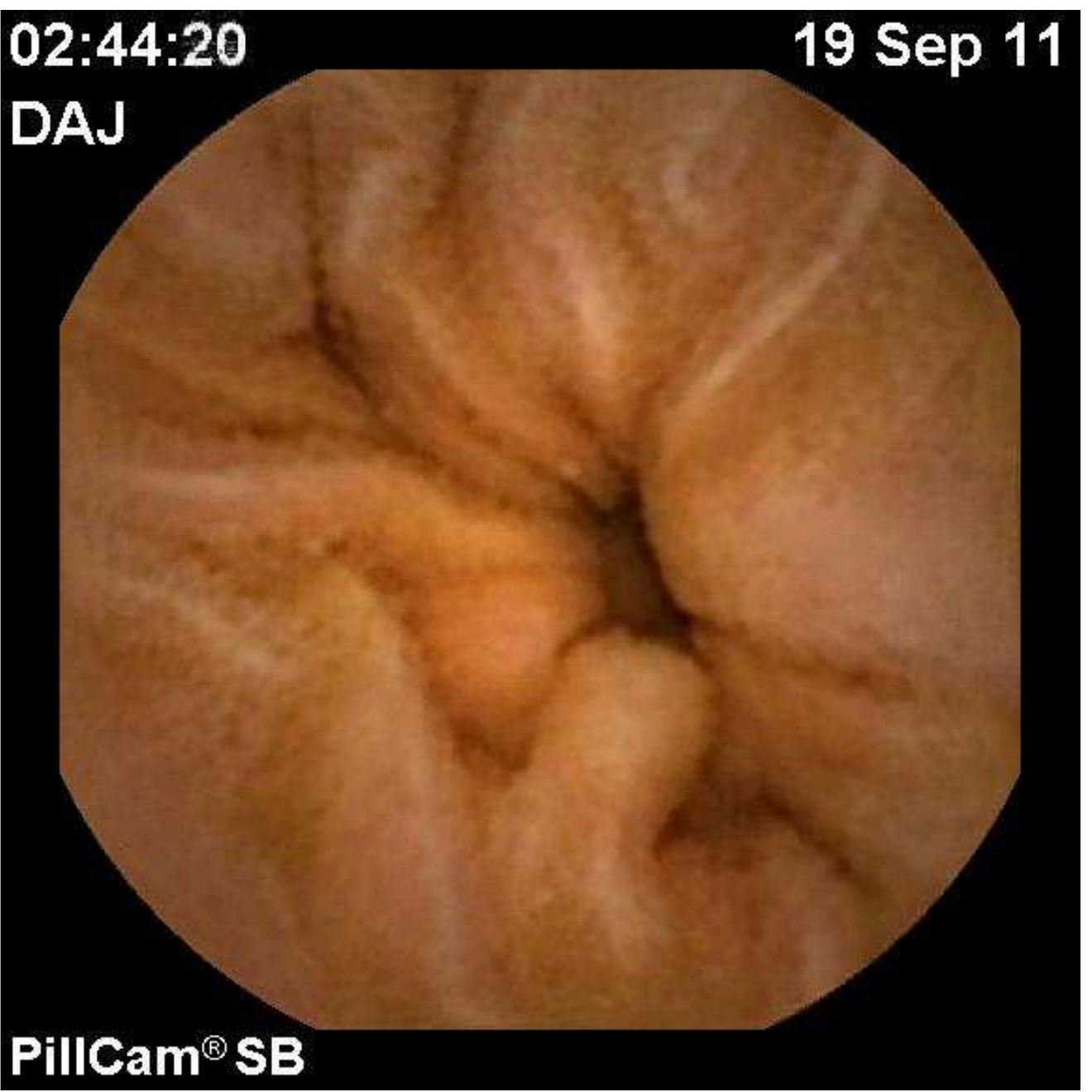}
\end{minipage}
\begin{minipage}[c][7.5cm][c]{0.5\textwidth}
\centering
  \includegraphics[width=7.5cm,height=7.5cm]{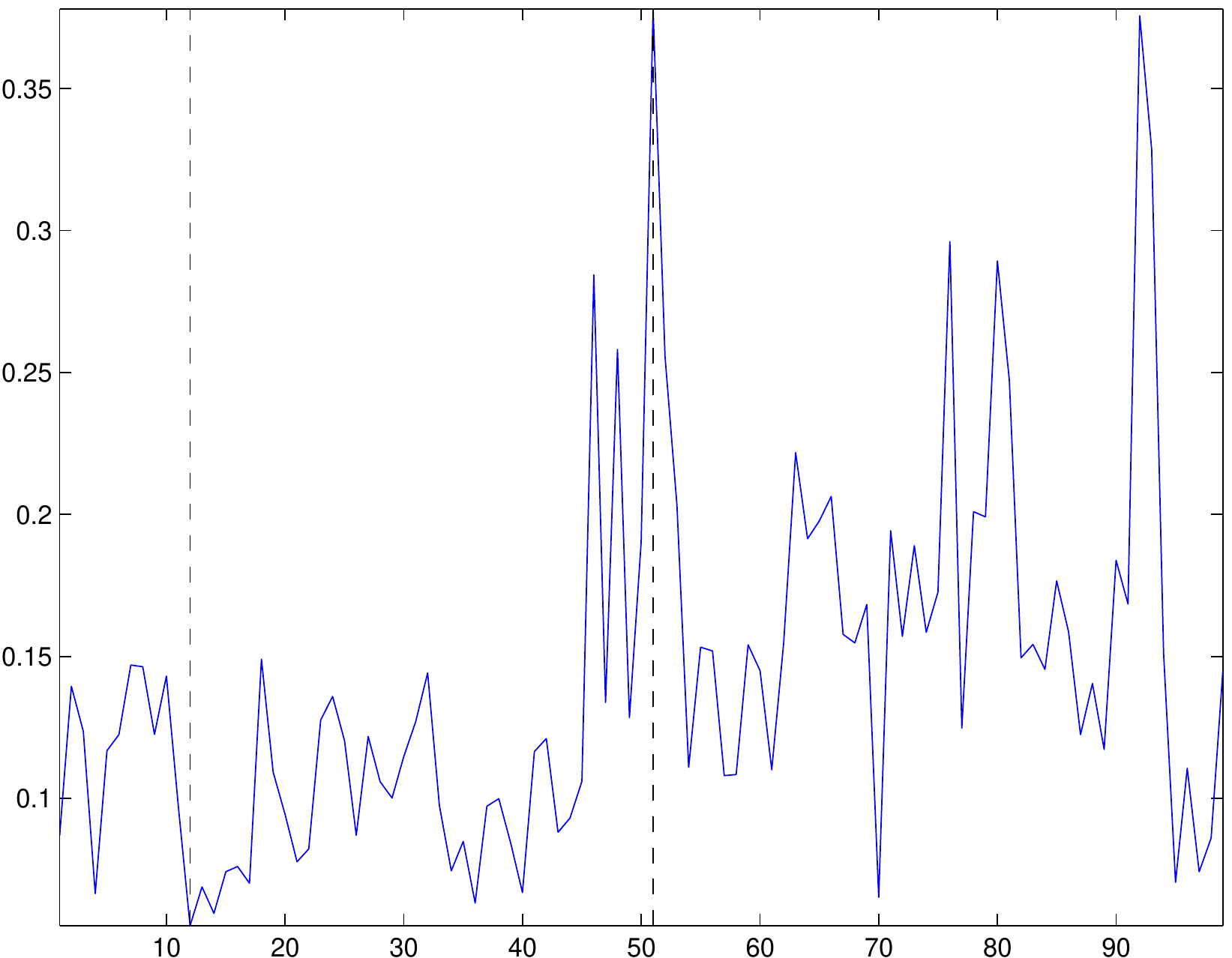}
\end{minipage}
\begin{minipage}[c][7.5cm][c]{0.23\textwidth}
\centering
  \includegraphics[width=3.5cm,height=3.5cm]{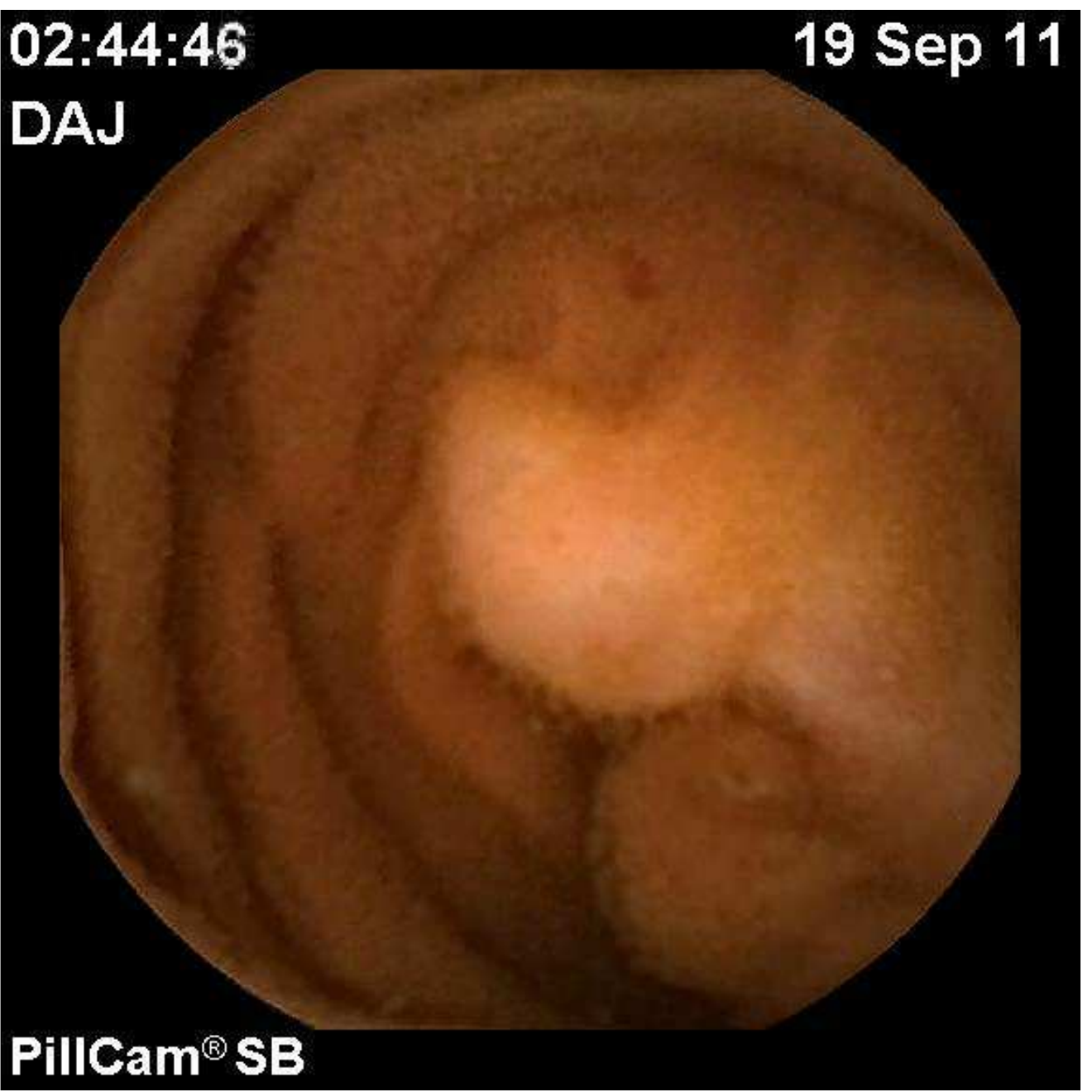}\par \vfill
  \includegraphics[width=3.5cm,height=3.5cm]{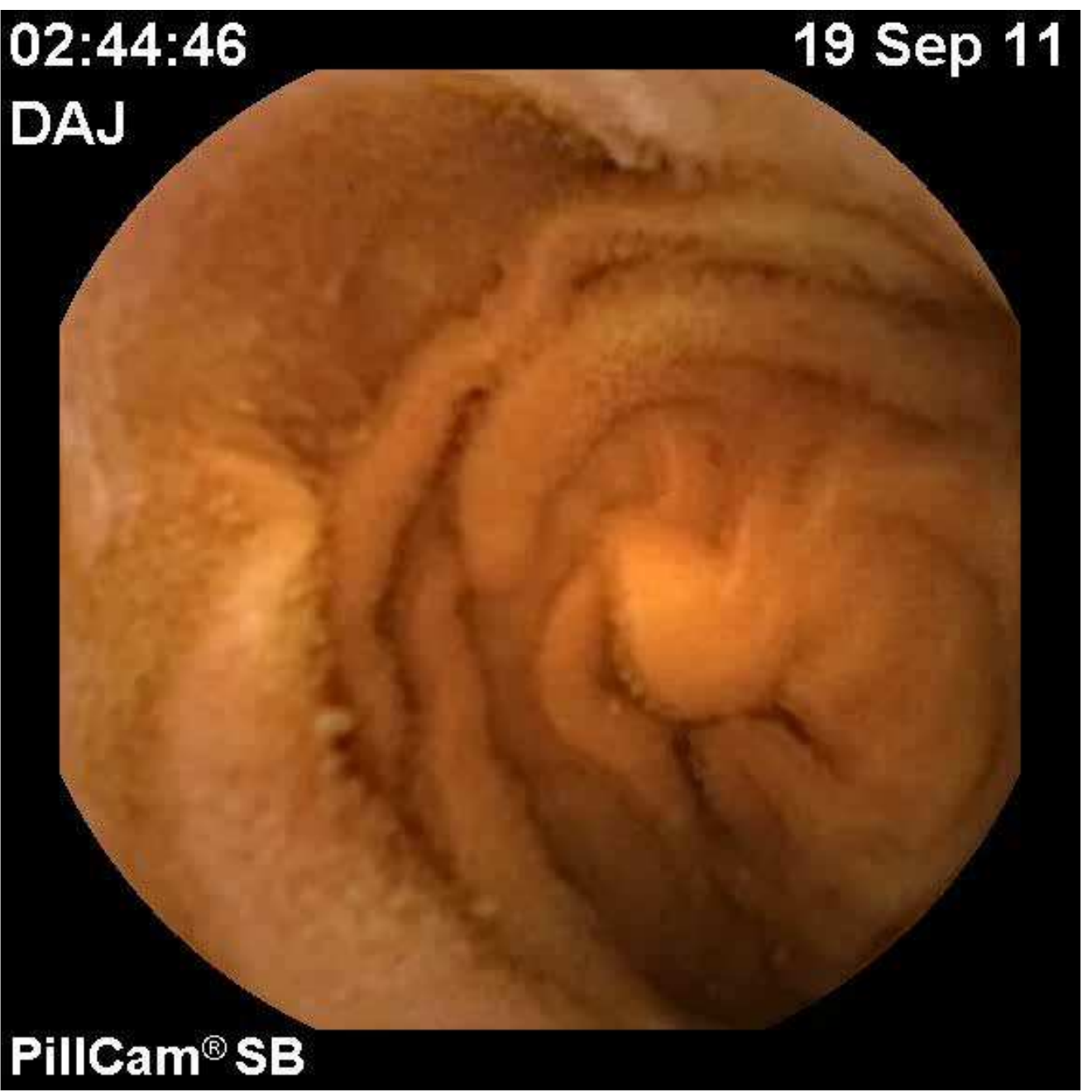}
\end{minipage}
\caption{Middle graphic: Qualitative speed estimation of the capsule in a  WCE video clip, with the duration of $20$ seconds and 100 frames, represented by the $NDM$  similarity curve between the consecutive frames, obtained with MEIR. First and Third columns:  Examples  of two pairs of consecutive frames of the video, registered with MEIR (the frames on the top are the templates and the references correspond to the bottom frames). The  pair on the left corresponds to the frames 12 and 13, exhibiting a big similarity, and for this pair the point in  the $NDM$ curve is   $(12,0.05523)$. The  pair on the right  displays the dissimilarity frames  51 and 52, and  the corresponding point   in the $NDM$ curve is $(51,0.37815)$. }\label{fig:speed1}
\end{figure}

 \begin{figure}[t!]
\centering
\includegraphics[width=7.2cm,height=6.2cm]{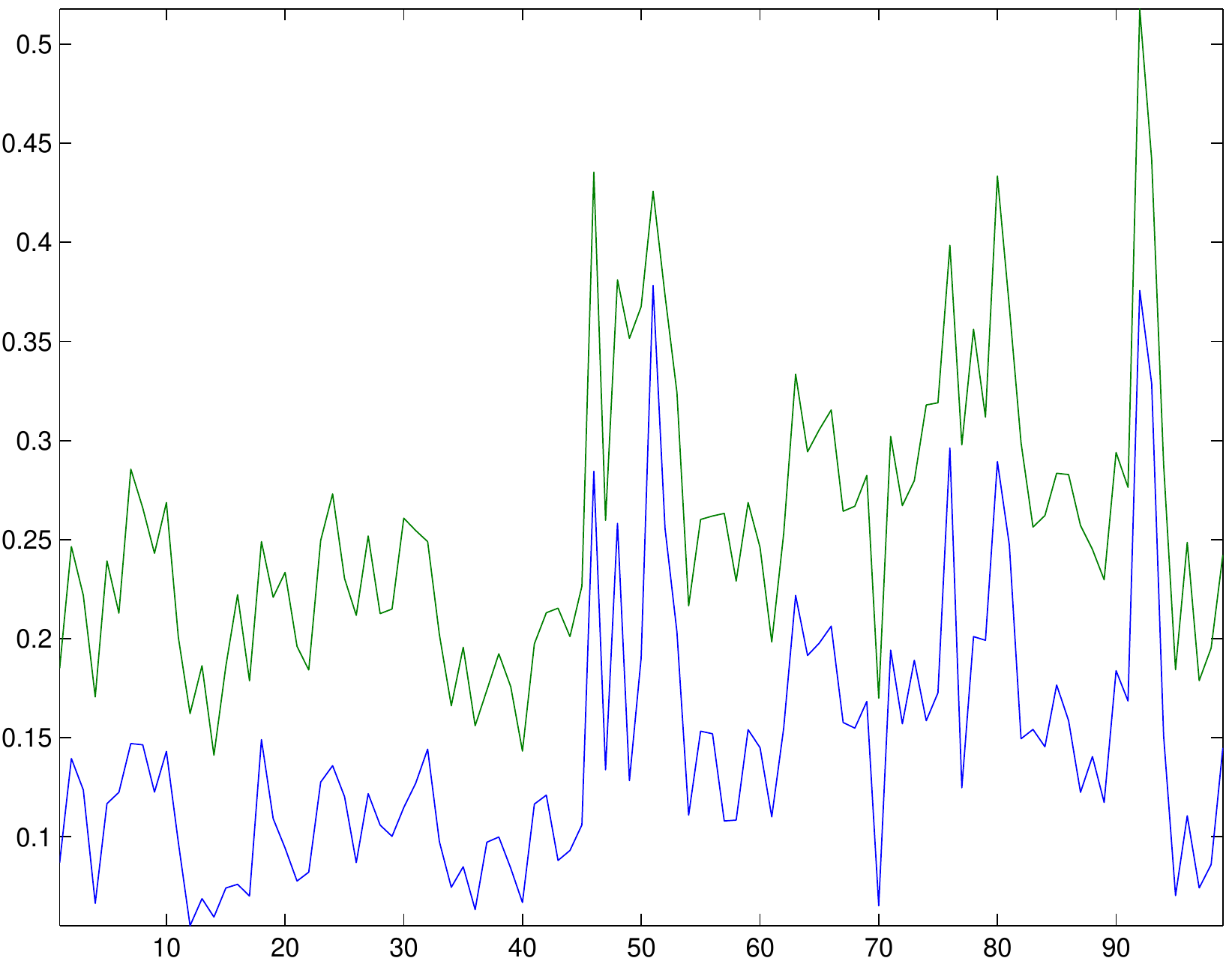}
\caption{Qualitative speed estimation of the capsule in a WCE video clip, with the duration of $20$ seconds and 100 frames, represented by the curves showing the similarity measure  $NDM$ between the frames, obtained with MEIR (blue curve) and MPIR (green curve). %
}\label{fig:speed2}
\end{figure}

Figure \ref{fig:speed2} displays the $NDM$ curves for the two approaches (MEIR and MPIR), for the same video considered in  Figure \ref{fig:speed1},
 and when the registration is done in the forward direction (starting from frame number 1 to  100).

 We also note that MEIR (and also MPIR) is a technique to match consecutive video frames, so it is particularly effective, when these frames have common regions, but not so effective when the frames are totally dissimilar.  The corresponding $NDM$ curve gives a valuable WCE speed information in regions where the WCE movement is continuous. When there are abrupt changes in consecutive  frames, the registration approaches lead to peaks in the $NDM$ curves, that accurately identify  the different pairs of consecutive frames where these peaks occur, however, the MEIR (or MPIR) approach, itself, is not very informative in these cases.

A comparison between the $NDM$ curves obtained with MEIR and MPIR reveals that there is a bigger gap between similar and dissimilar frames (respectively,   low and high values for $NDM$)  in the curve  generated with  MEIR  than with MPIR. This result evidences a better separation between similar/quite similar and different consecutive frames,  and thus a better performance of the MEIR  registration approach. This was somewhat expected, because  the small intestine is an elastic organ, and in motion due to peristalsis, therefore an elastic registration approach is  more suited than an affine one.
We refer as well  to Figure \ref{fig:incr_elast}  for a comparison,  for a single frame, between the $NDM$ curves, obtained with MEIR and MPIR, as the amount of elastic deformation increases.

Figure \ref{fig:3real} exhibits 3 different pairs $(R,T)$ of consecutive frames in WCE videos. For each pair we can perceive an elastic deformation and/or a rotation and/or a change in scale while passing from the previous frame $T$ to the following one $R$.
 Figure \ref{fig:3real_results} shows the results obtained with MEIR,  for each pair in Figure \ref{fig:3real}.  The grids correspond to the transformations obtained with one MEIR iteration. Clearly the transformed templates $T(Id-u)$, displayed on the first row of  Figure \ref{fig:3real_results}, demonstrate  the elastic matching of  these there pairs of consecutive video frames.

Finally,   we note that in order to improve  the efficiency of the  MEIR approach,  the affine  pre-registration problem \eqref{eq:opt} can be solved by a multi-level strategy by considering down-sampled images. Using a two-level approach for solving  \eqref{eq:opt}, first with $64\times 64 = 2^6\times 2^6$  and then  with $128\times 128 = 2^7\times 2^7$  points, for both the template and reference images, we have observed a reduction of $9\%$ in the overall  MEIR computation  time.

\subsection{Experiments with artificial frames}\label{sec:testart}

To evaluate   the performance of the  proposed multiscale approach  (elastic with affine pre-registration, MEIR) and also for a comparison with the  multiscale  fully parametric registration approach, MPIR (that  is similar to many other existing approaches that rely only on affine correspondences between frames) we  start by  simulating transformations of video frames. Secondly we register the originals and corresponding simulated frames with the proposed MEIR and MPIR registration procedures, and finally  we compare the results. More specifically, we proceed in the following way:

\begin{enumerate}

\item For each  small bowel video, 20 frames are selected, by sampling the video every 1 second. Thus there is a total of 780 frames.

\item For each sampled video frame we build a synthetically elastic deformed frame,  together with  a scaled or/and rotated  deformed version of it (either separately  or in a collective, {\it i.e.} using two or more transformations simultaneously). Figures \ref{fig:rot_sca} and \ref{fig:elas} show examples of synthetic frames.

\item We   register the original video frame and the corresponding modified version of it, using the two multiscale approaches, MEIR and MPIR.

\item We use the normalized dissimilarity measure $NDM$ introduced in \eqref{eq:decision} to assess and compare the accuracy of the registration approaches MEIR and MPIR,  for all the tests.

\item We further assess and compare the performance of MEIR and MPIR, for tracking the capsule within the body, by using the idea described in  \cite{spyrou2014video}   for estimating the displacement and  orientation of the WCE.  In  fact, in \cite{spyrou2014video}  the scale and rotation parameters, resulting from an affine registration scheme (that involves the algorithms SURF and RANSAC),  are identified  with the capsule displacement and orientation using a projective transformation and the pinhole camera model.  Here we use the  scale and and rotation parameters resulting from MEIR and MPIR approaches, for inferring the displacement and  orientation of the WCE as in  \cite{spyrou2014video}.

The solution of MPIR corresponds to an affine transformation of the type \eqref{eq:rig} and gives immediately the scale $\omega_0$ and rotation $\omega_1$ needed for WCE localization and orientation, following  \cite{spyrou2014video}.  When the MEIR approach is used,    we need to consider the affine transformation of the form \eqref{eq:rig}   closest to the solution of the MEIR approach (iterated twice), in the least-squares sense,   to deduce  the   WCE localization and orientation as in \cite{spyrou2014video}.

\begin{figure}[t!]
\centering
\includegraphics[width=3.25cm,height=3.25cm]{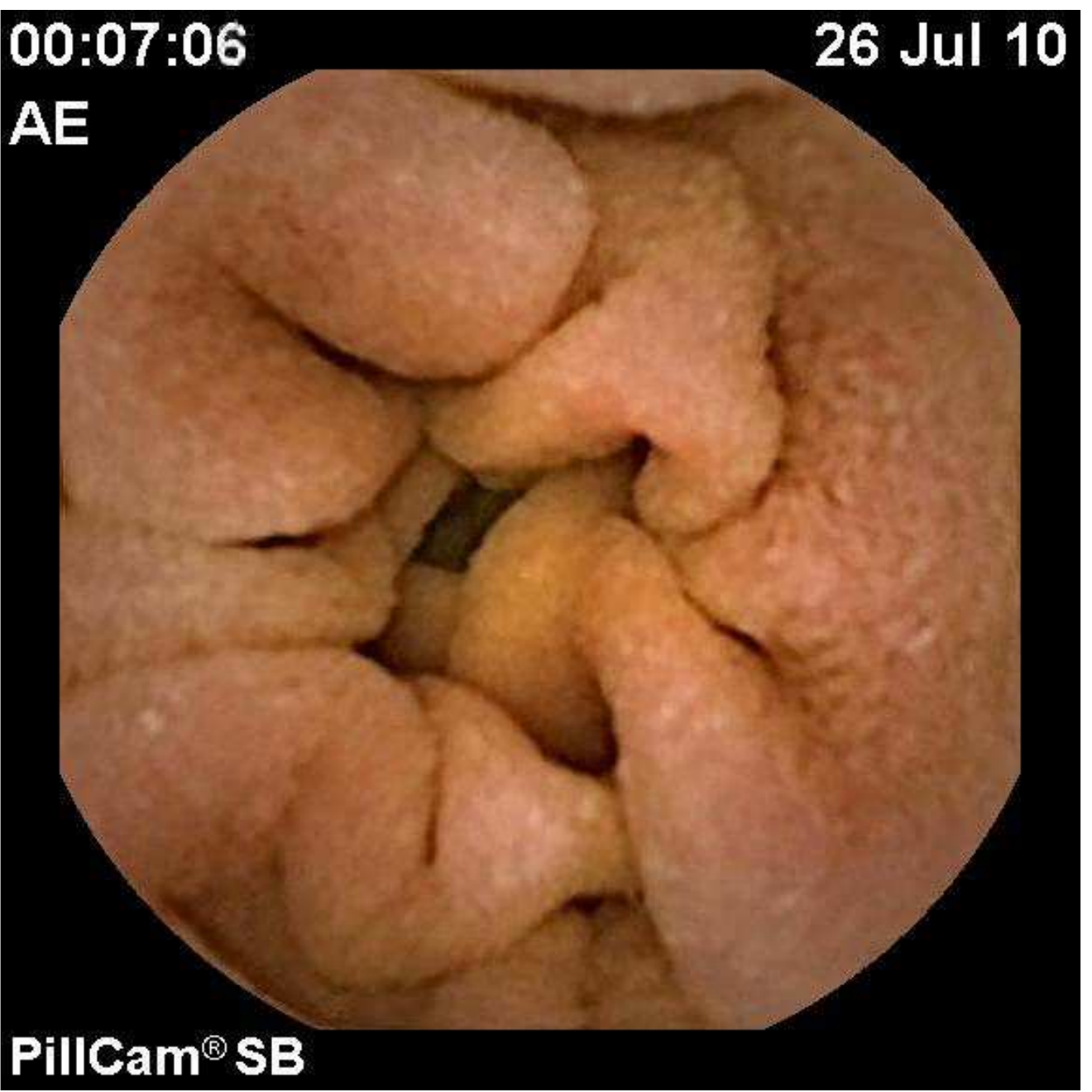}\hspace{2 mm}
\includegraphics[width=3.25cm,height=3.25cm]{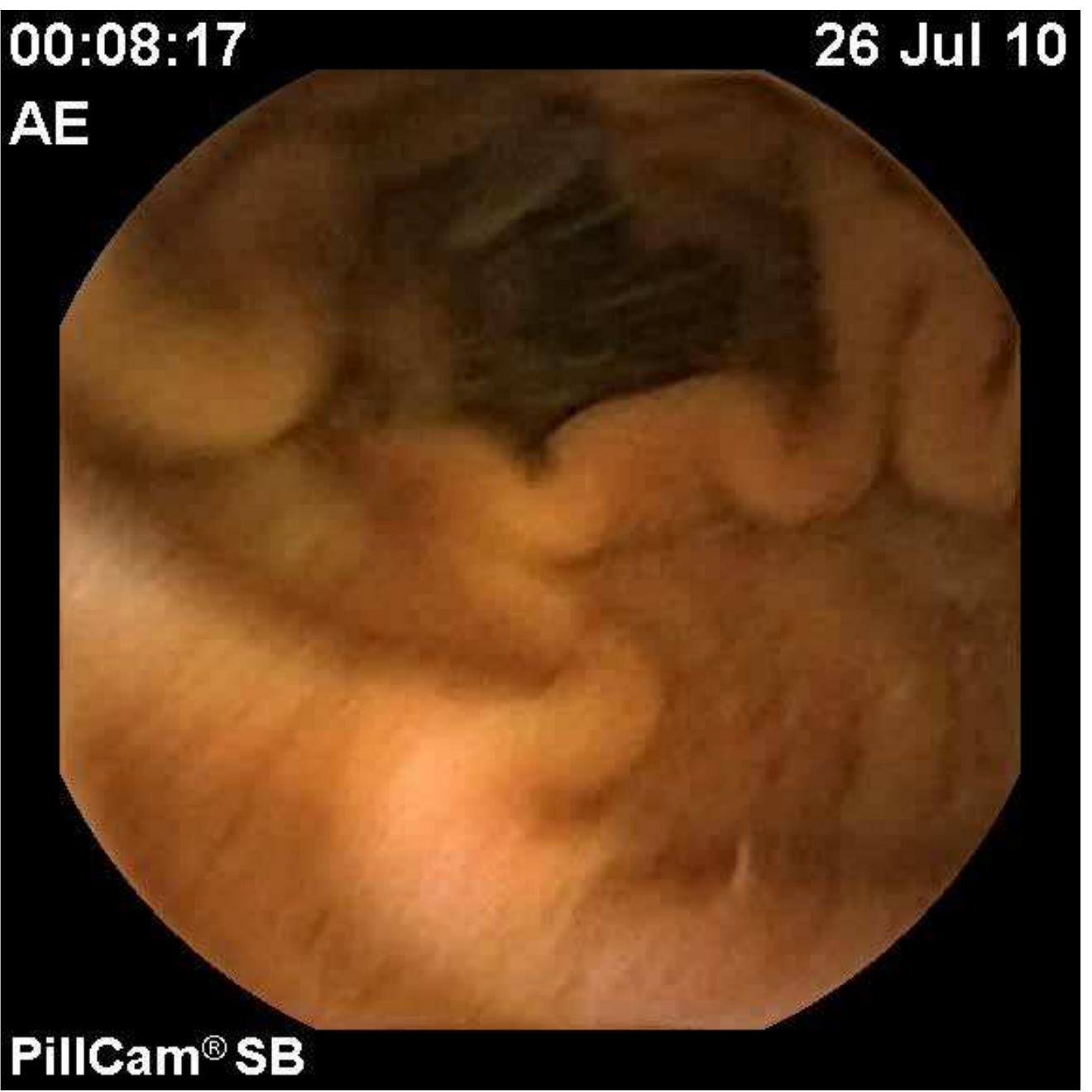}\hspace{2 mm}
\includegraphics[width=3.25cm,height=3.25cm]{Figs/Fig7_2185/Fig7_3_R.pdf}\\
\vspace{2 mm}
\includegraphics[width=3.25cm,height=3.25cm]{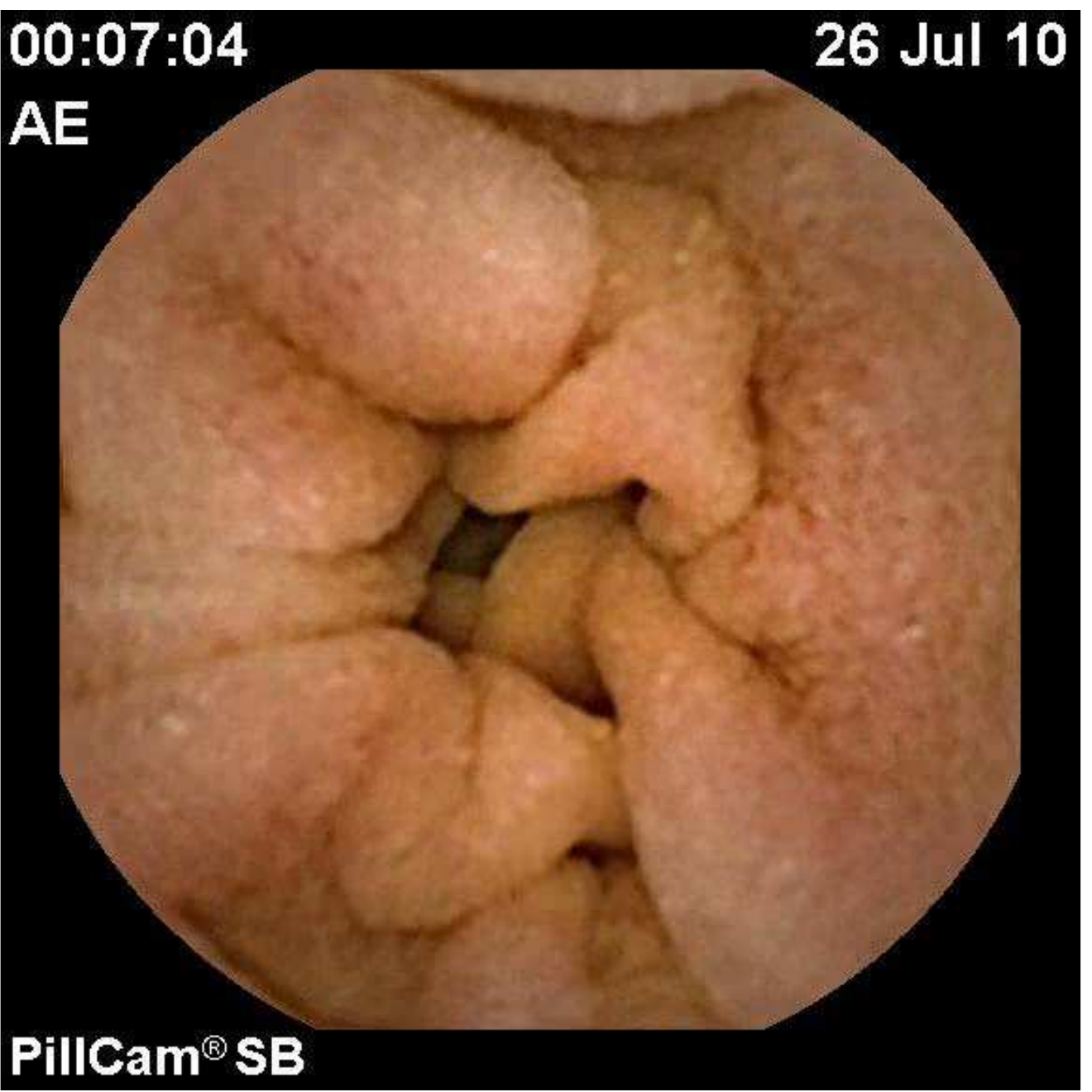}\hspace{2 mm}
\includegraphics[width=3.25cm,height=3.25cm]{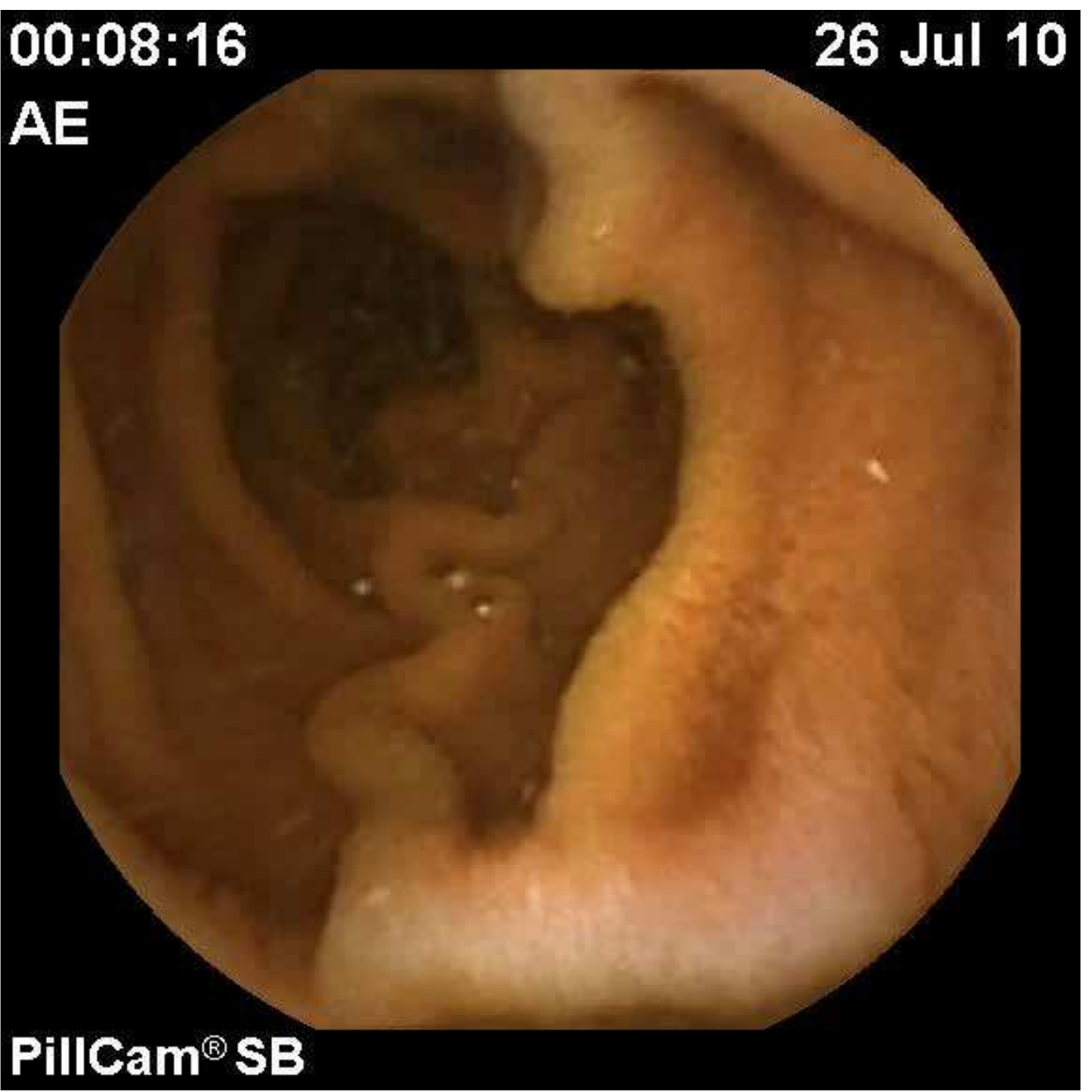}\hspace{2 mm}
\includegraphics[width=3.25cm,height=3.25cm]{Figs/Fig7_2185/Fig7_3_T.pdf}\\
\caption{Three  columns showing three different pairs $(R,T)$ of consecutive   frames  in WCE videos (original frames). The first line shows the reference images  $R$ and the bottom line the template  images $T$. Image $R$ follows $T$ in the video.}\label{fig:3real}
\end{figure}

 \begin{figure}[h!]
\centering
\includegraphics[width=3.25cm,height=3.25cm]{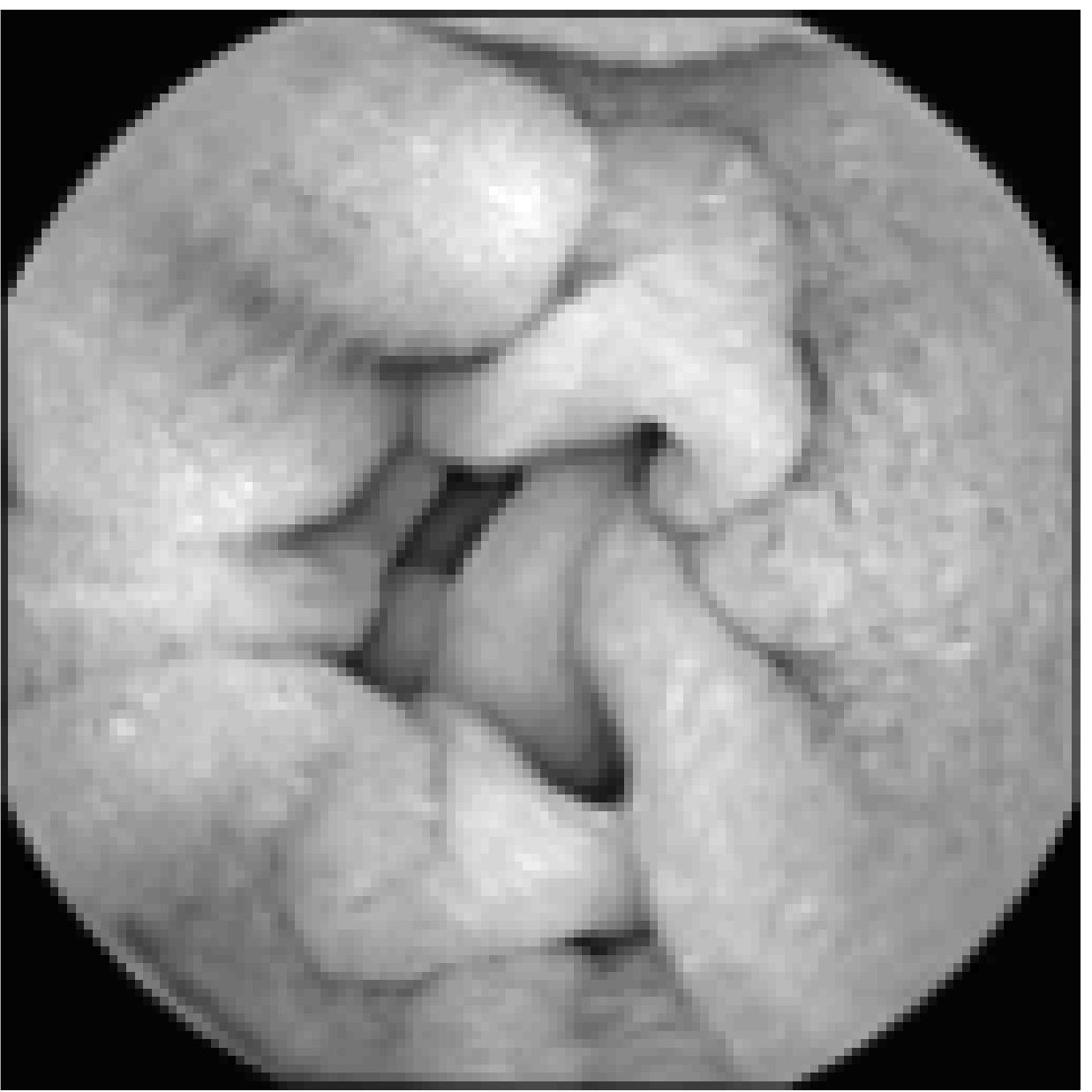}\hspace{2 mm}
\includegraphics[width=3.25cm,height=3.25cm]{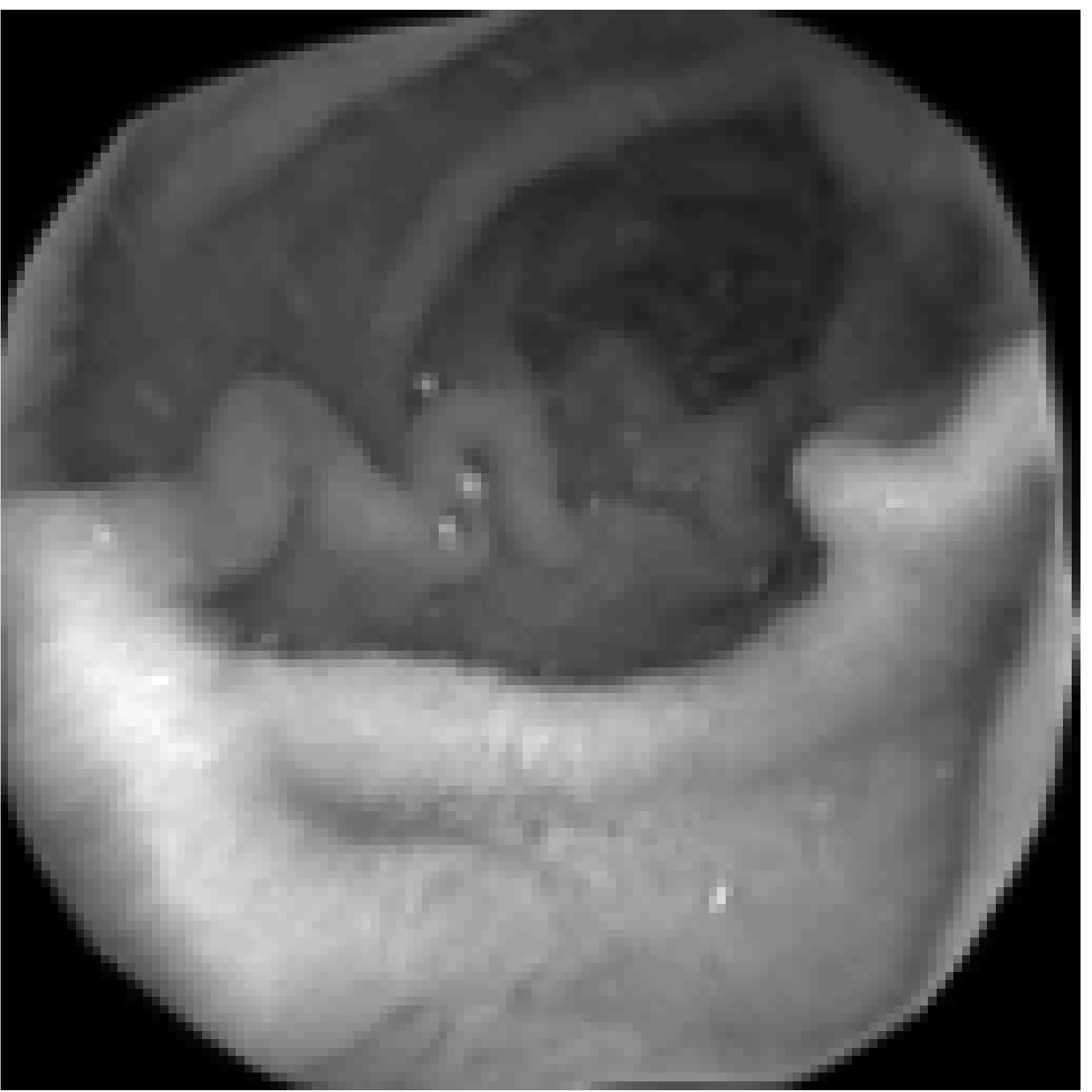}\hspace{2 mm}
\includegraphics[width=3.25cm,height=3.25cm]{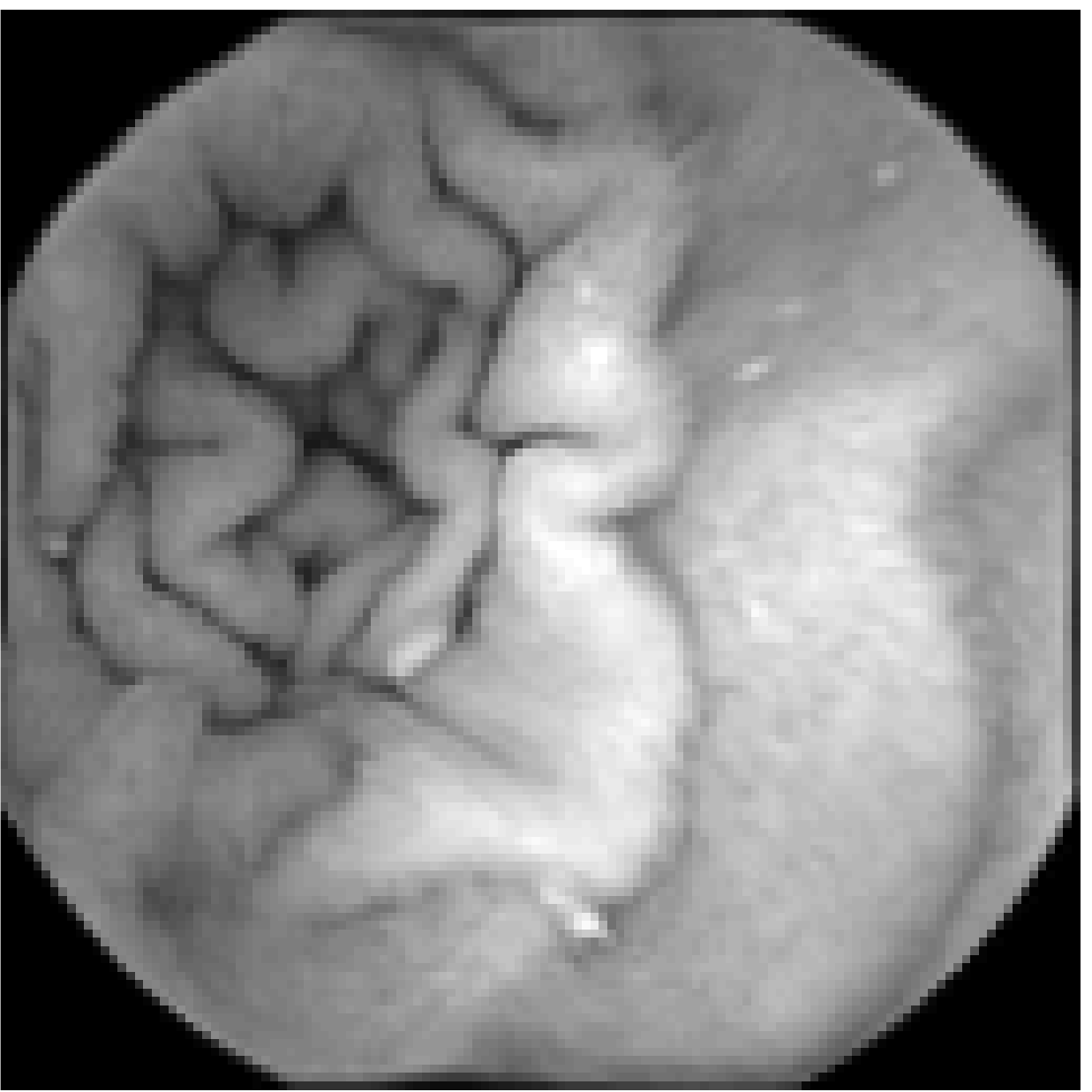}\\
\vspace{2 mm}
\includegraphics[width=3.25cm,height=3.25cm]{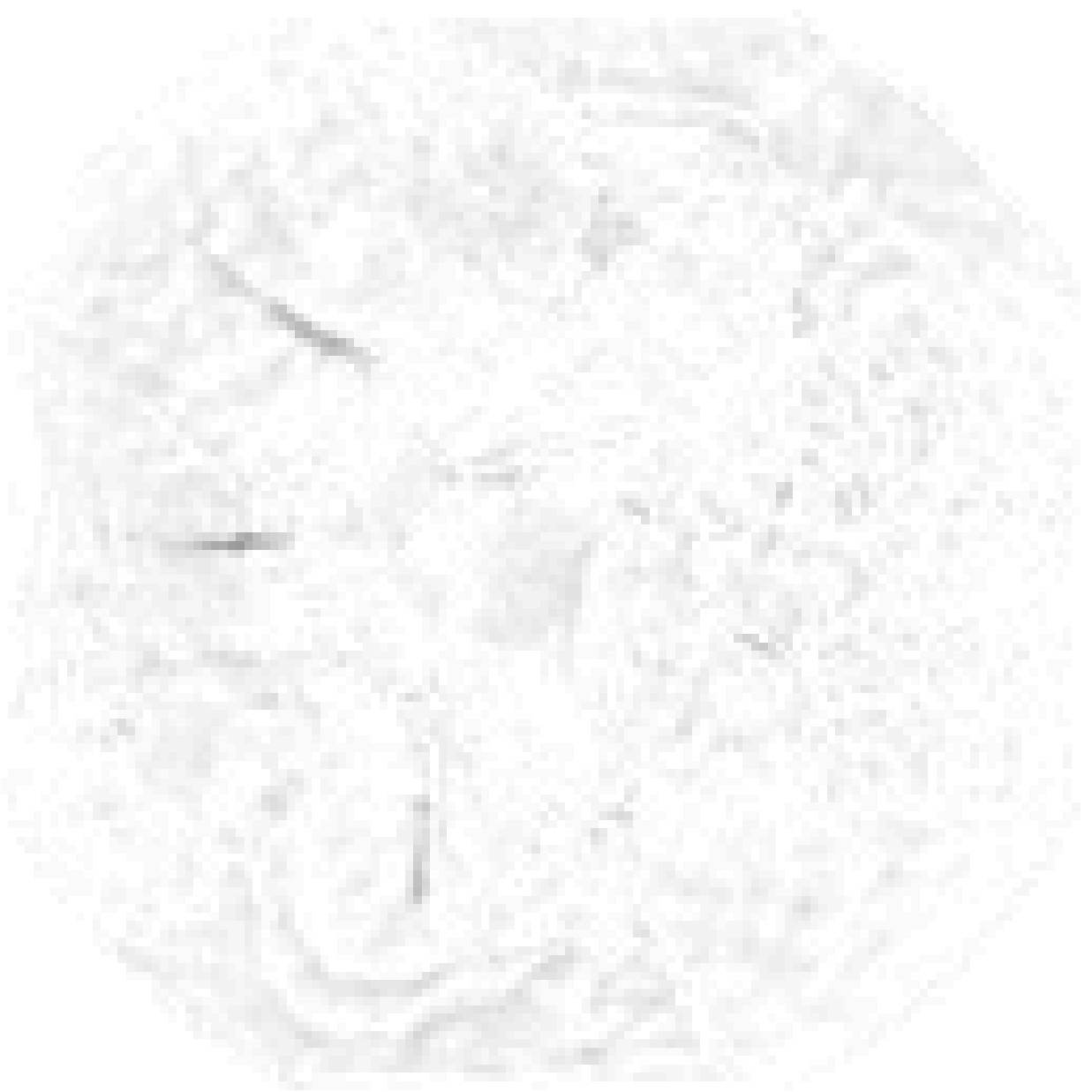}\hspace{2 mm}
\includegraphics[width=3.25cm,height=3.25cm]{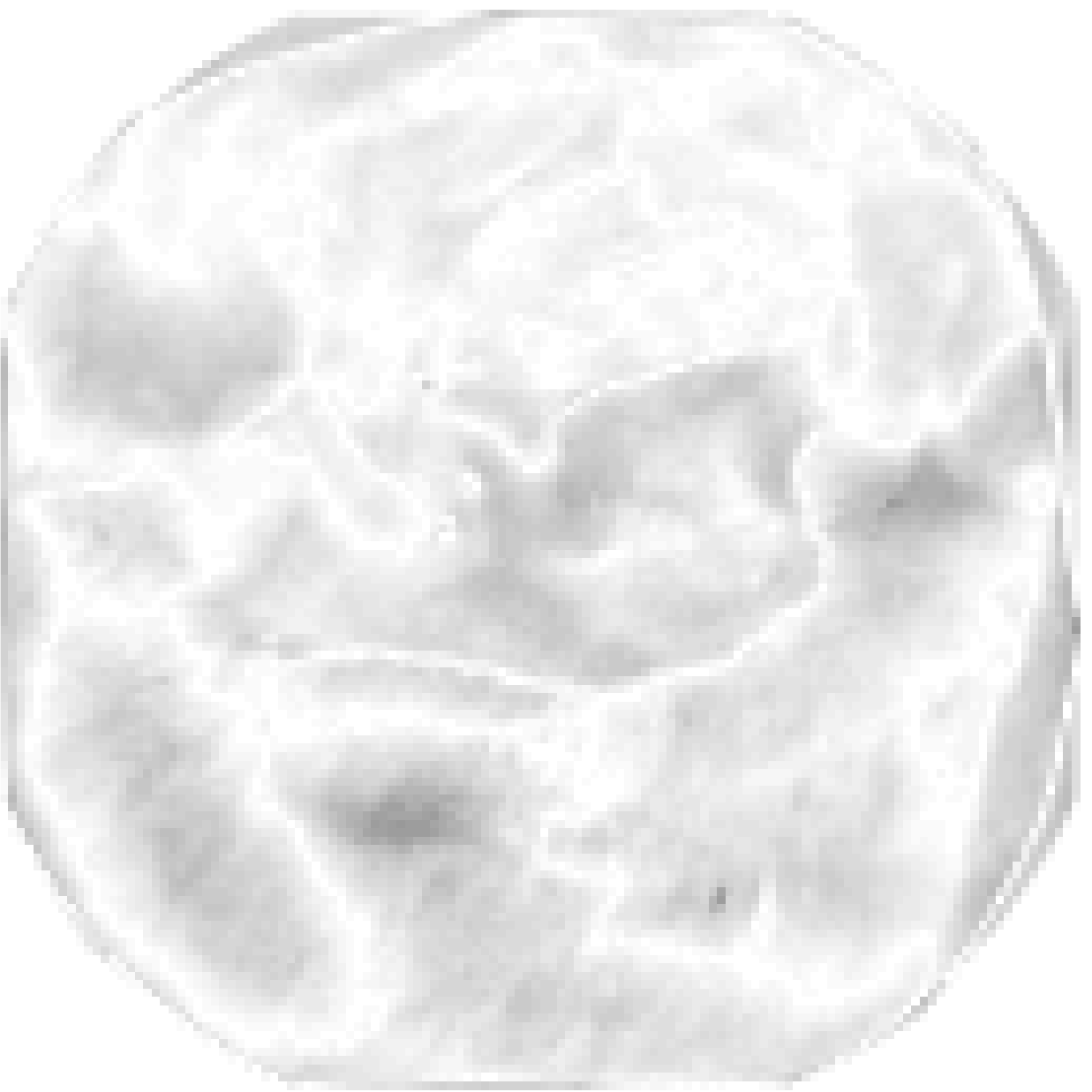}\hspace{2 mm}
\includegraphics[width=3.25cm,height=3.25cm]{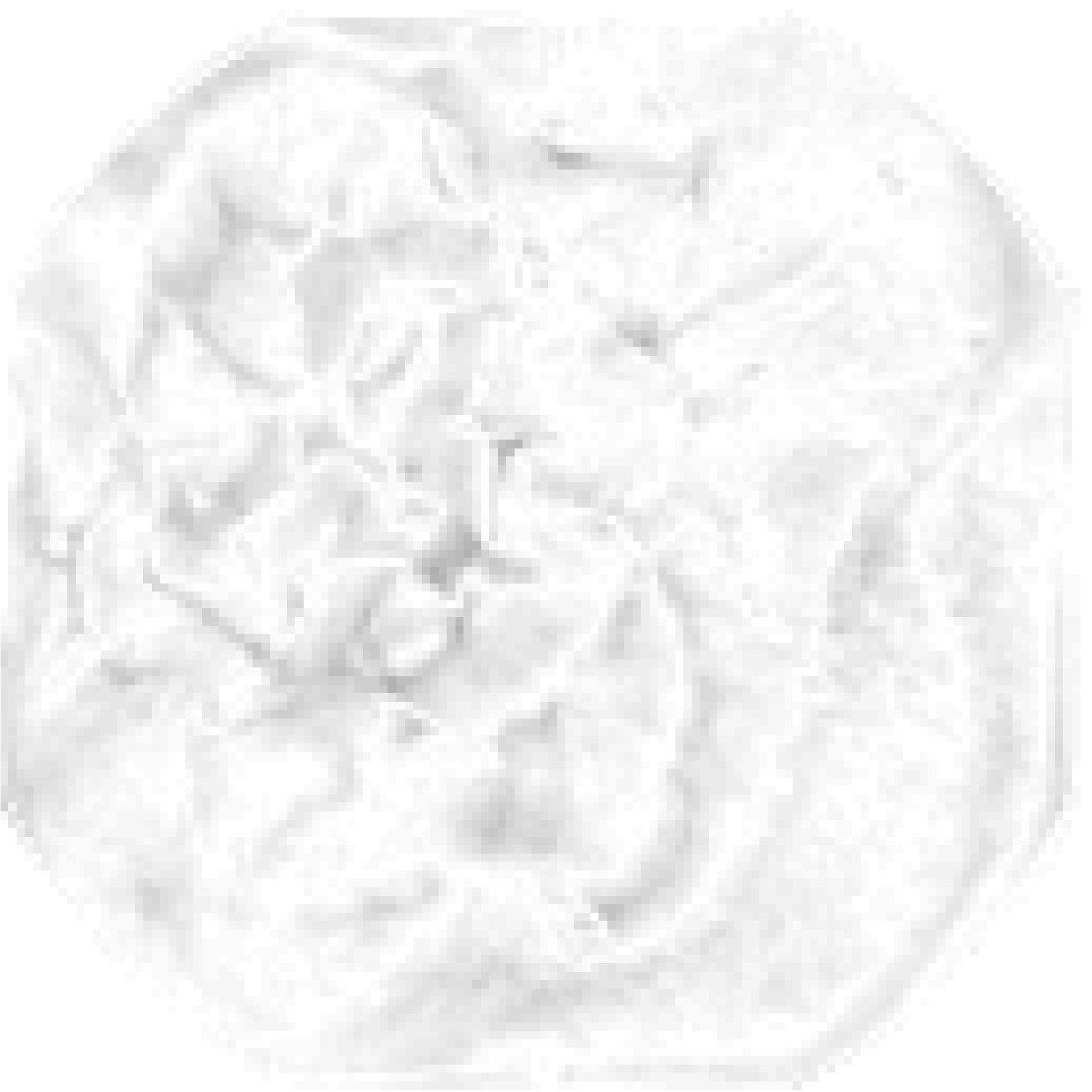}\\
\vspace{2 mm}
\includegraphics[width=3.25cm,height=3.25cm]{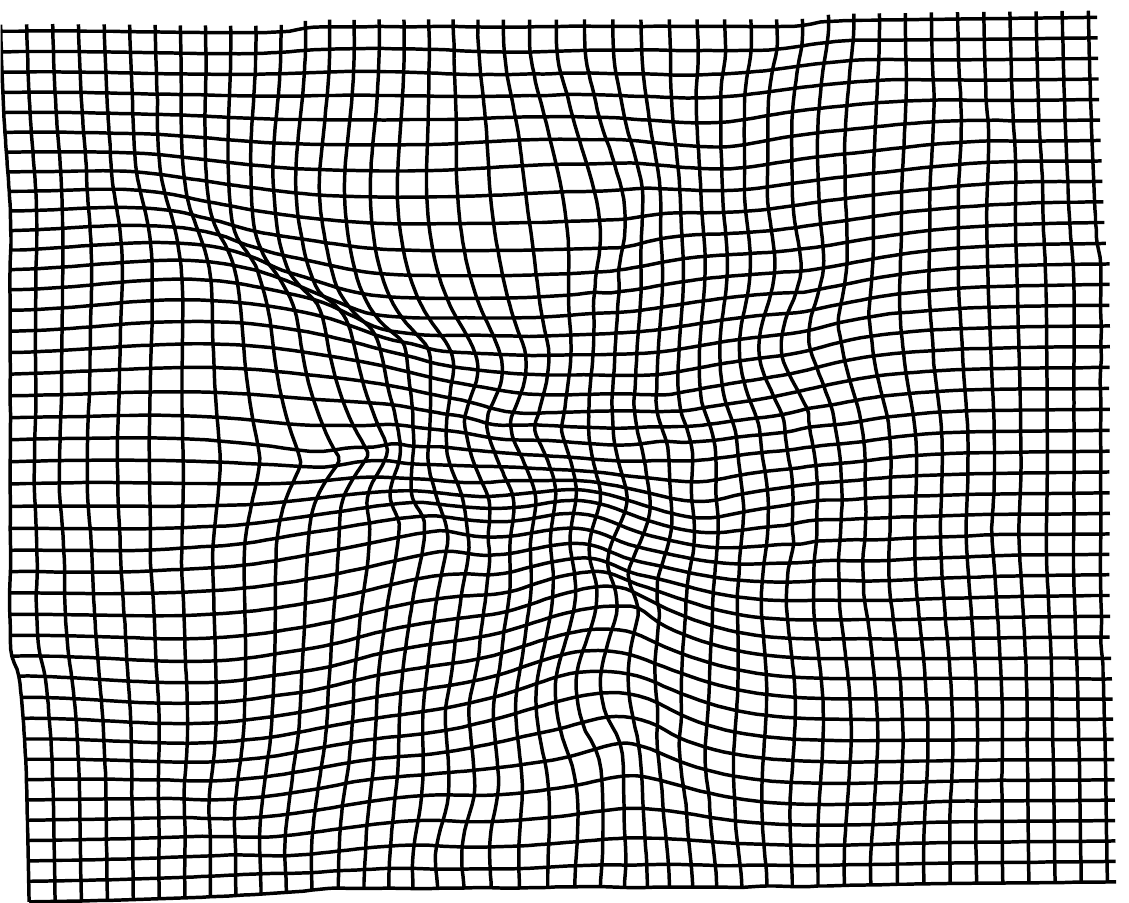}\hspace{2 mm}
\includegraphics[width=3.25cm,height=3.25cm]{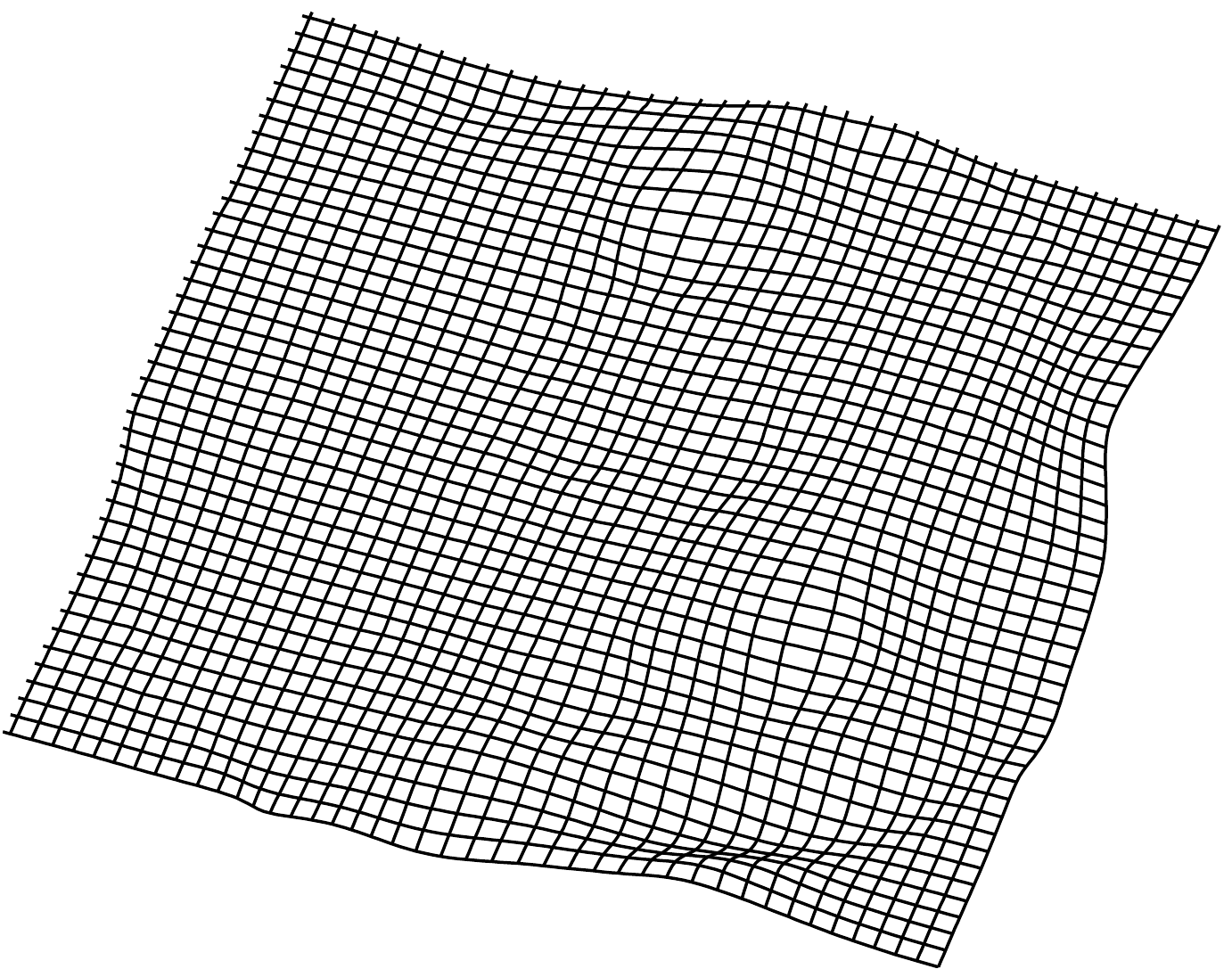}\hspace{2 mm}
\includegraphics[width=3.25cm,height=3.25cm]{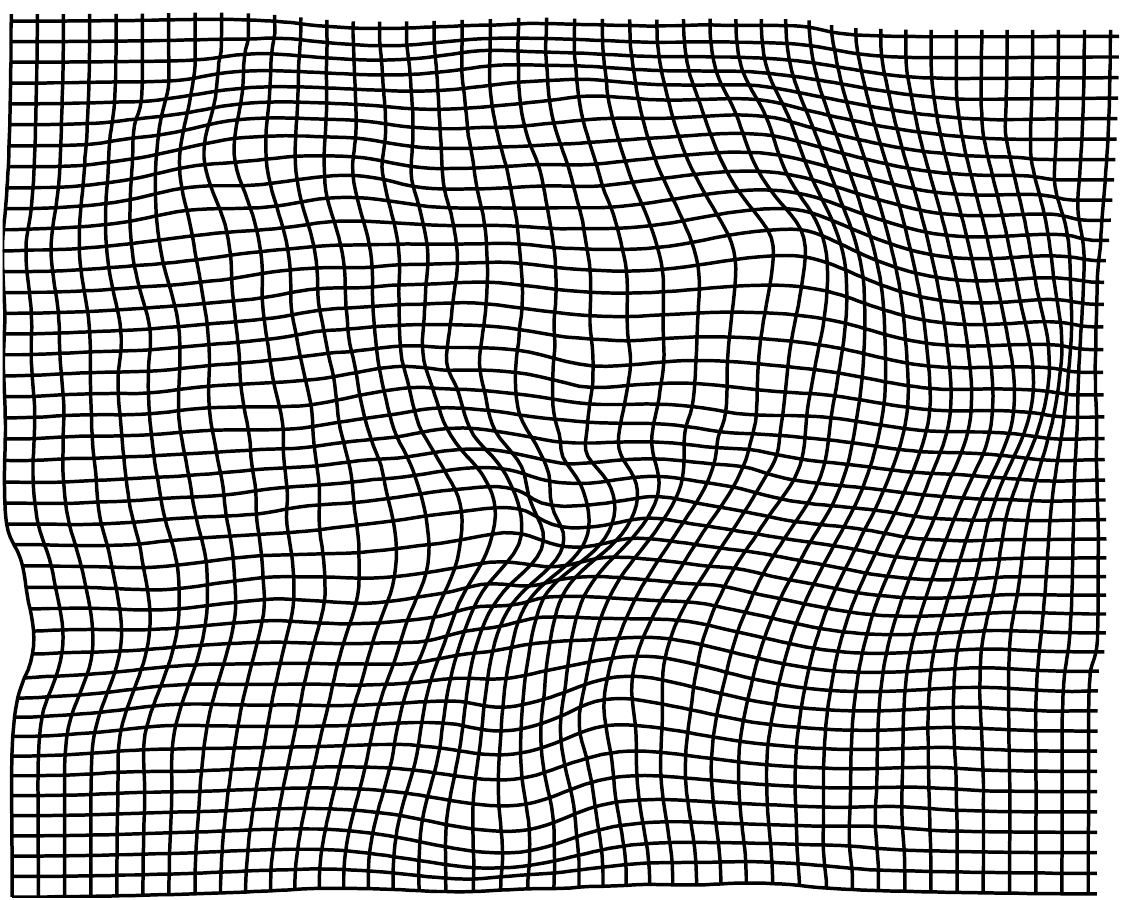}\\
\caption{Results obtained with MEIR for the three  $(R,T)$ pairs  of Figure \ref{fig:3real}. Each column shows (from top to bottom) :  the transformed template image $T(Id-u)$ to compare with $R$, the difference between the reference and the transformed template images, and finally the  deformed mesh $Id-u$  corresponding to the solution of MEIR approach.}\label{fig:3real_results}
\end{figure}

Finally,  for the all the tests  involving the frames synthetically generated, we estimate the scale or/and rotation errors for MEIR and MPIR,  by comparing the obtained scale and rotation parameters, $\omega_0$ and $\omega_1$,   with the {\it a priori} known scale  and rotation values used to built the  synthetically scaled or/and rotated frames.

\end{enumerate}

 \begin{figure}[t!]
\centering
\includegraphics[width=3.5cm,height=3.5cm]{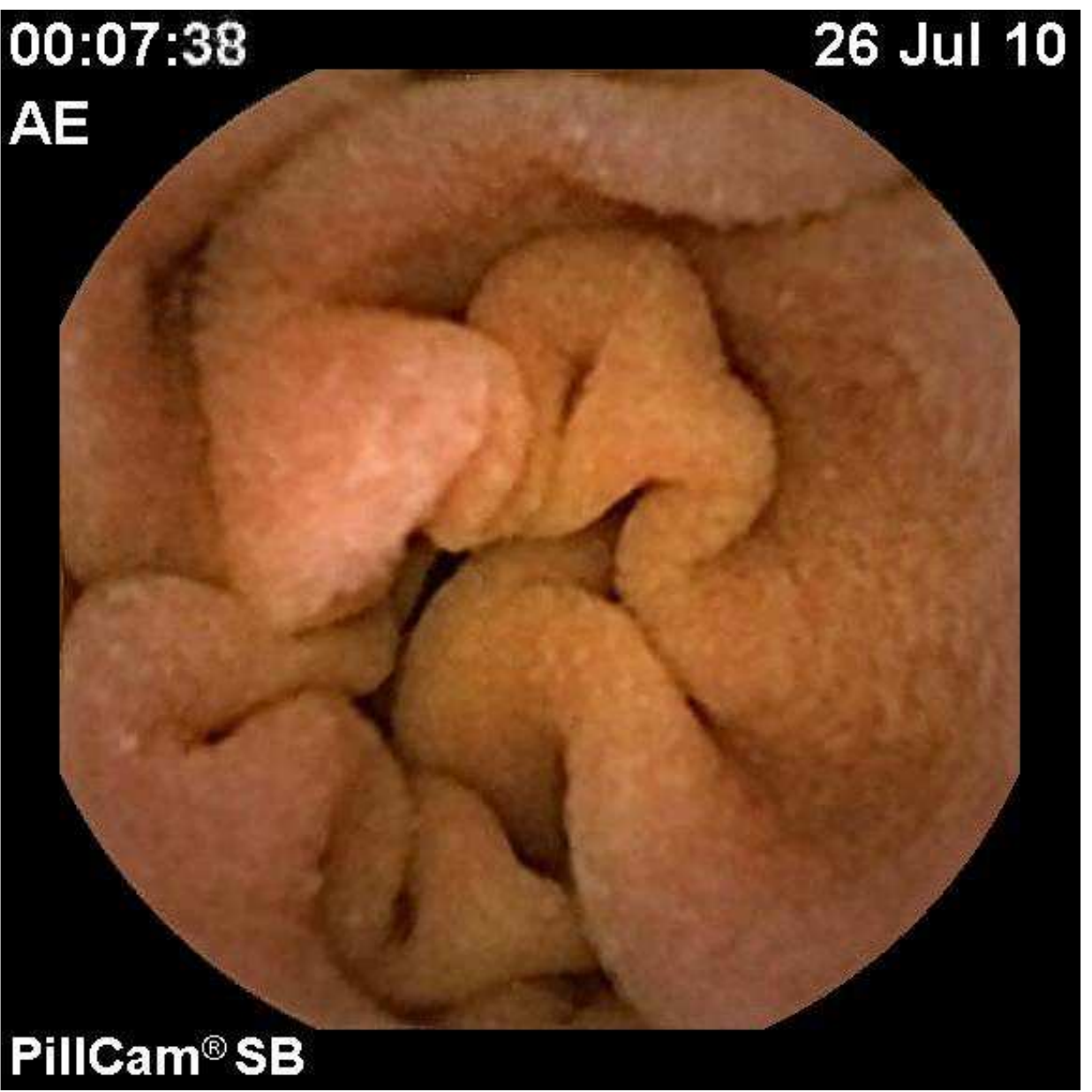}\hspace{2 mm}
\includegraphics[width=3.5cm,height=3.5cm]{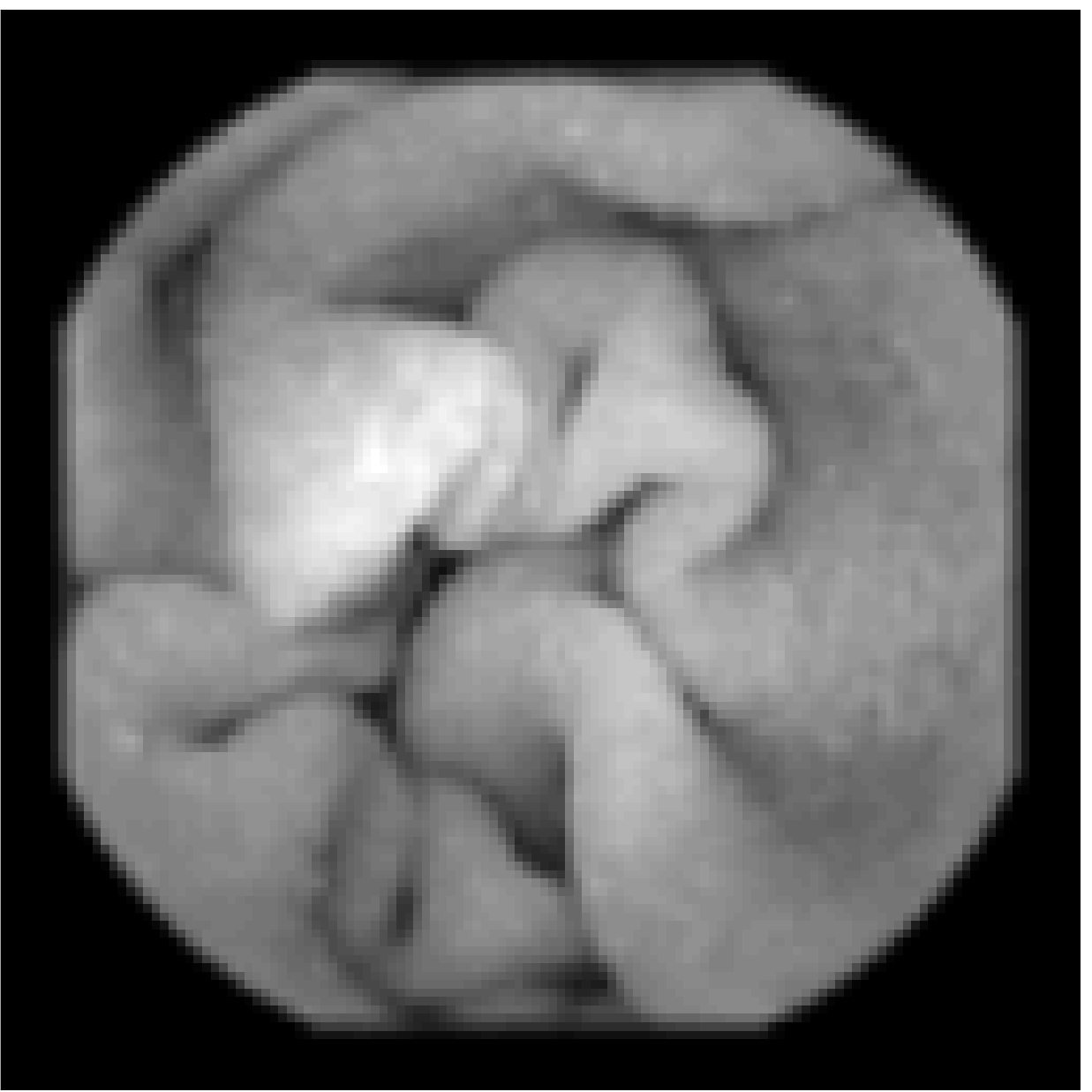}\hspace{2 mm}
\includegraphics[width=3.5cm,height=3.5cm]{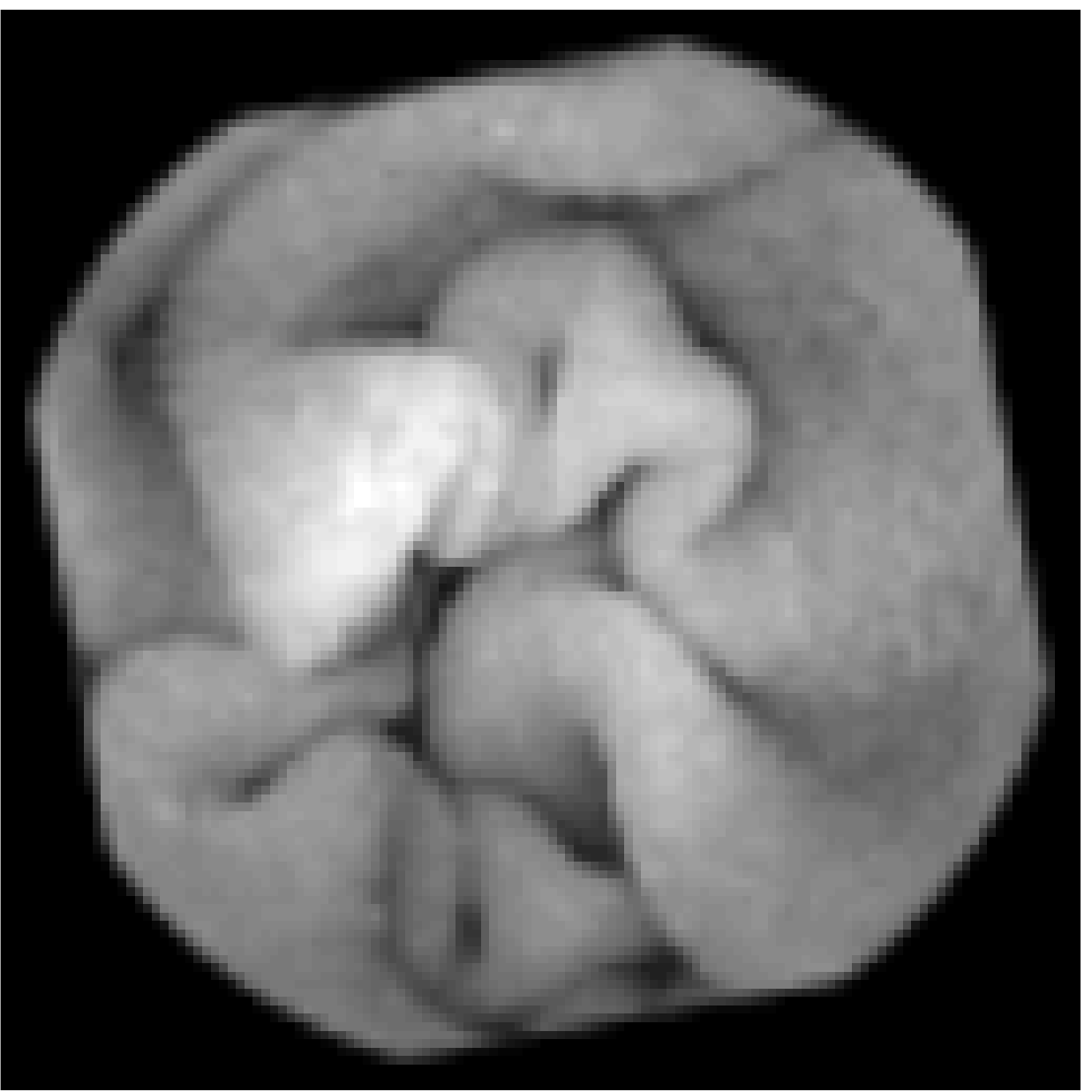}\hspace{2 mm}
\includegraphics[width=3.5cm,height=3.5cm]{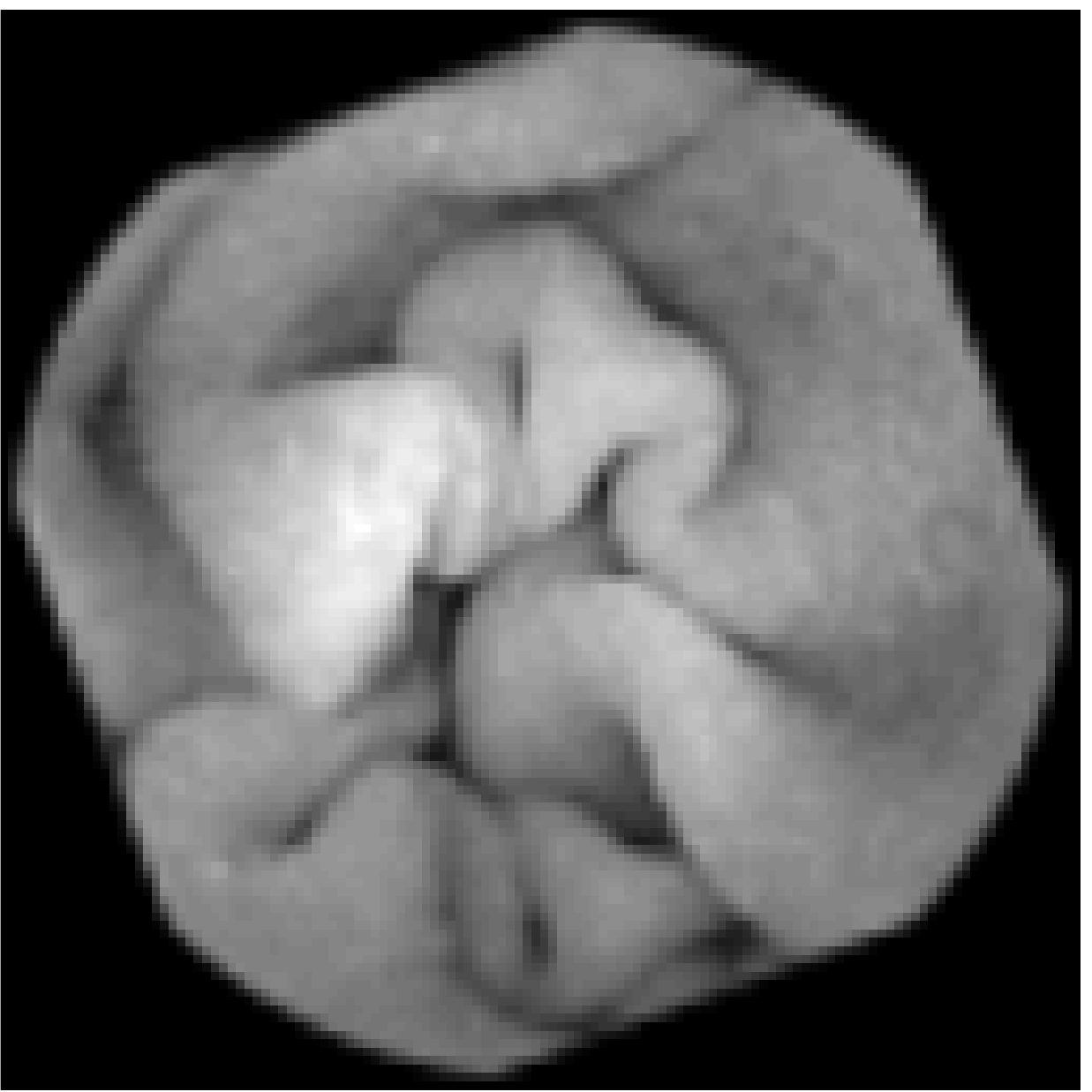}\\
\vspace{2 mm}
\includegraphics[width=3.5cm,height=3.5cm]{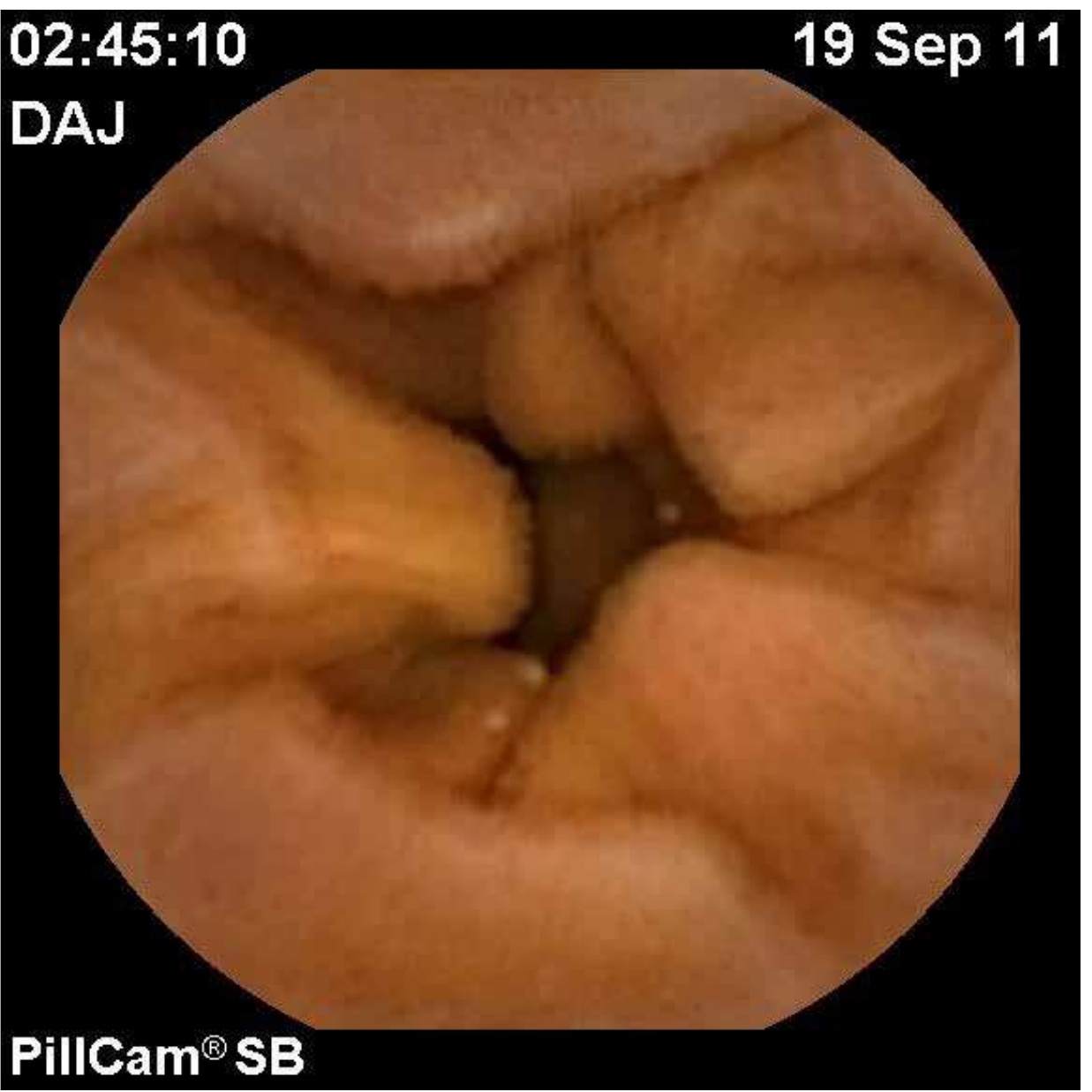}\hspace{2 mm}
\includegraphics[width=3.5cm,height=3.5cm]{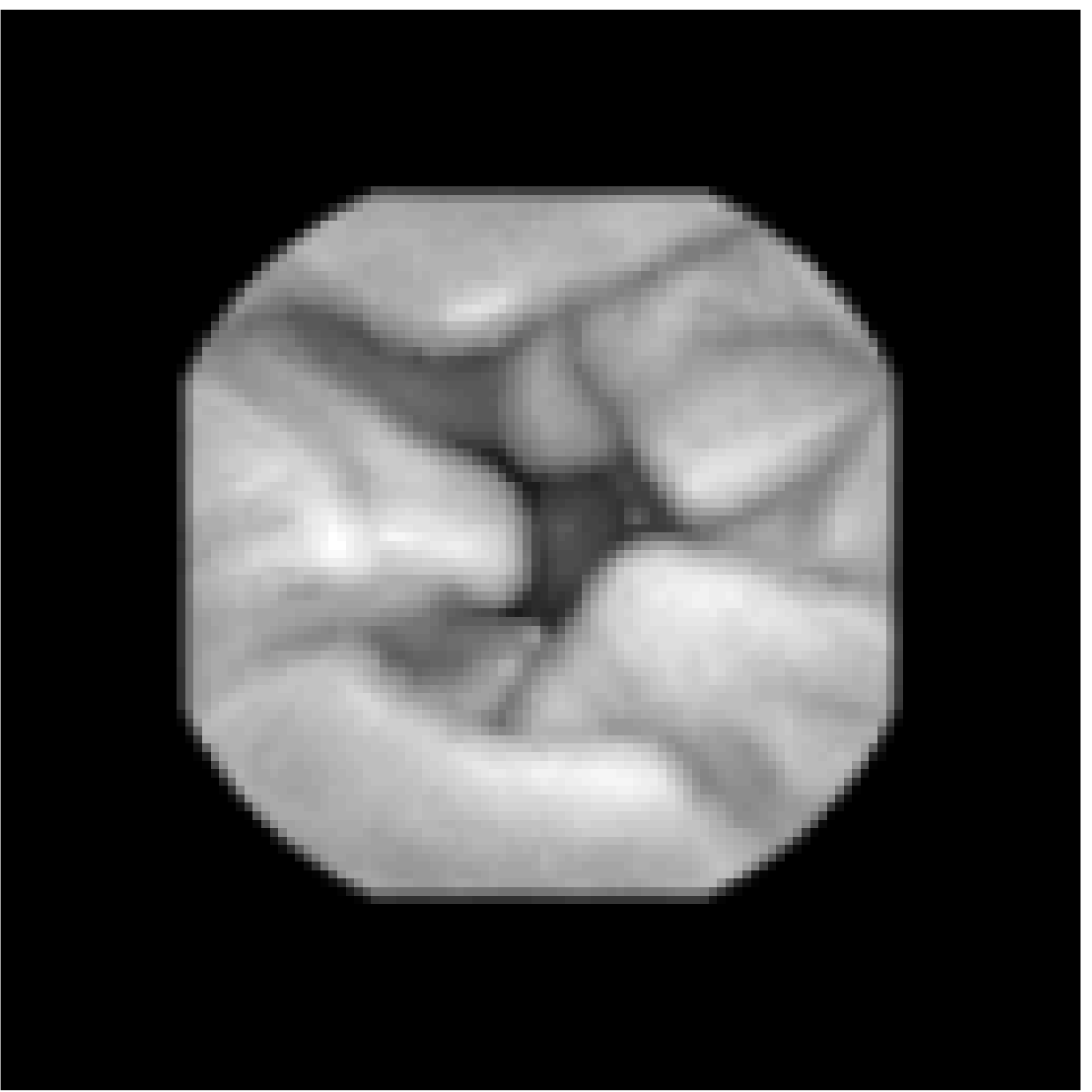}\hspace{2 mm}
\includegraphics[width=3.5cm,height=3.5cm]{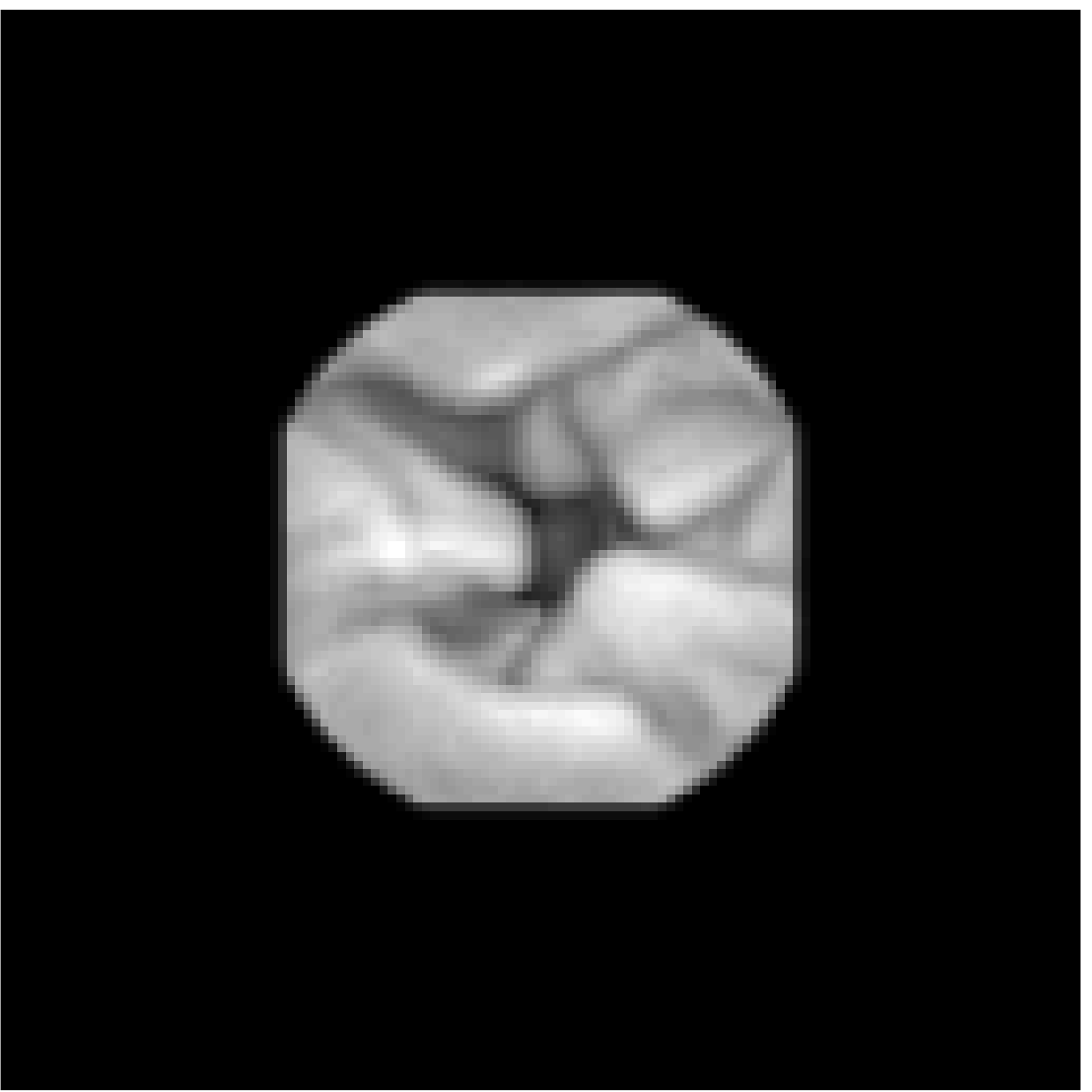}\hspace{2 mm}
\includegraphics[width=3.5cm,height=3.5cm]{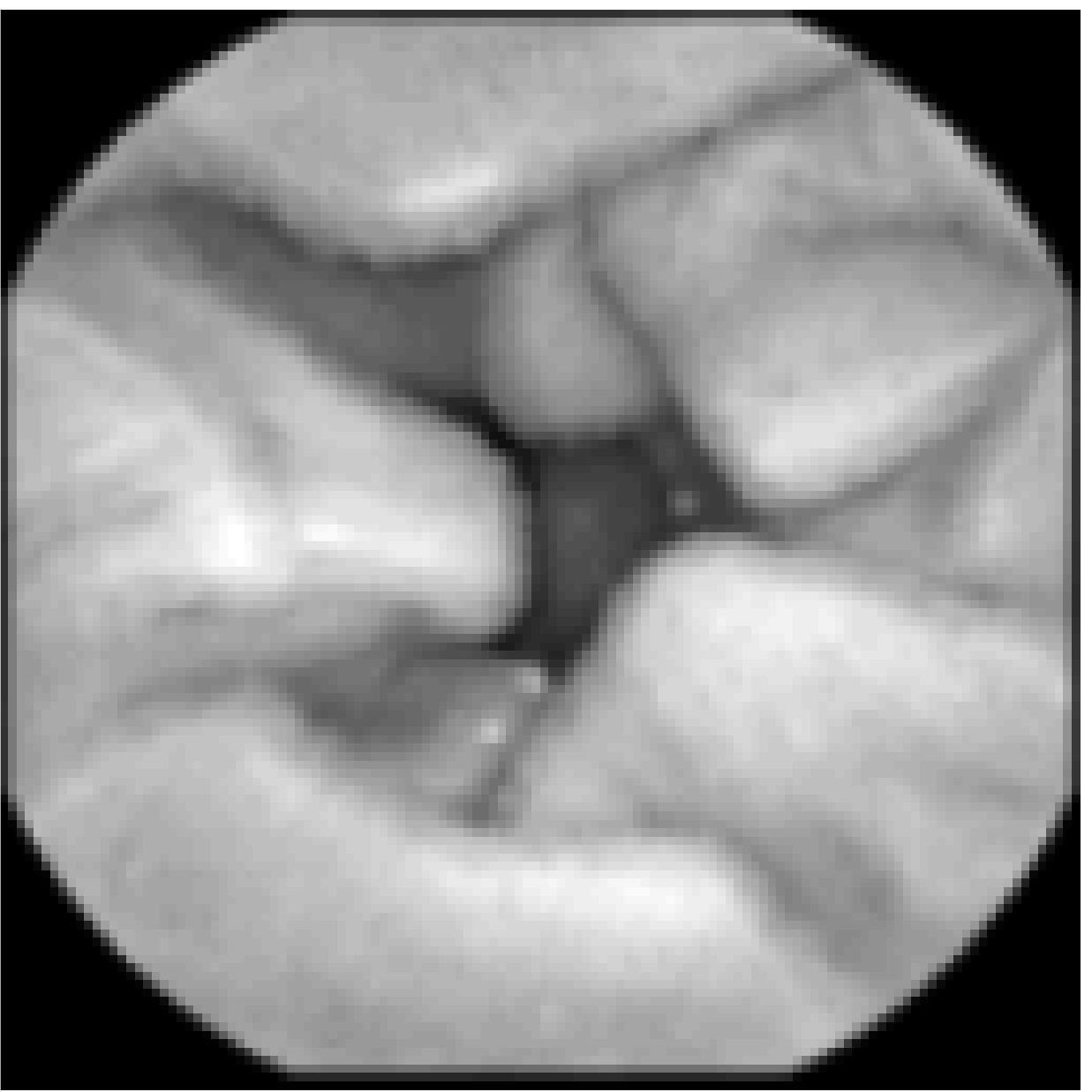}\\
\caption{First row (from left to right) : original frame, grayscale frame and its synthetic  rotated versions with rotation angles $\omega_1=10$ and $\omega_1=20$.
Second row (from left to right) : original frame and its synthetic scaled versions with scale factors $\omega_0=0.5$ and $\omega_0=1.5$.}\label{fig:rot_sca}
\end{figure}

\begin{figure}[t!]
\centering
\includegraphics[width=4.0cm,height=4.0cm]{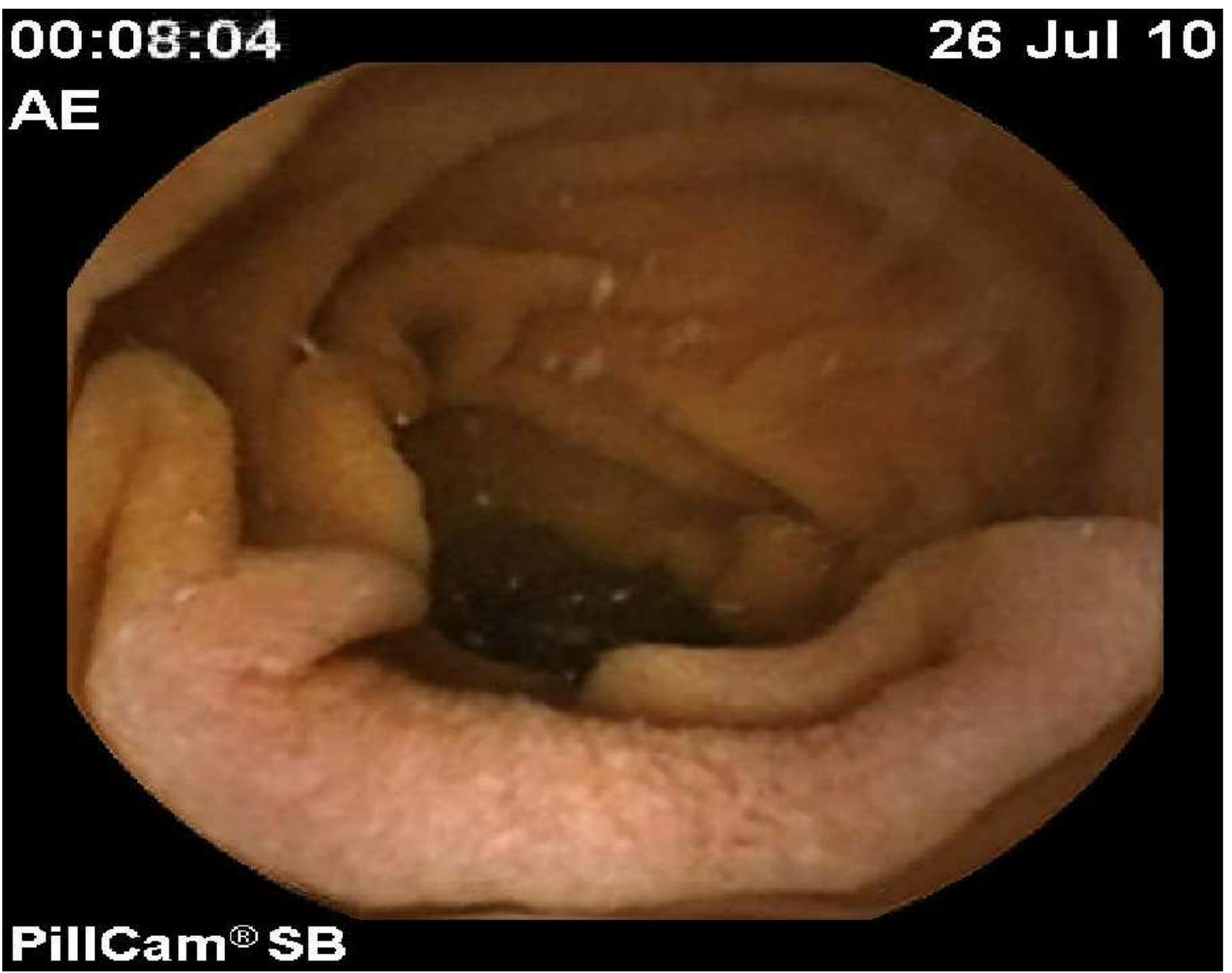}\hspace{2 mm}
\includegraphics[width=4.0cm,height=4.0cm]{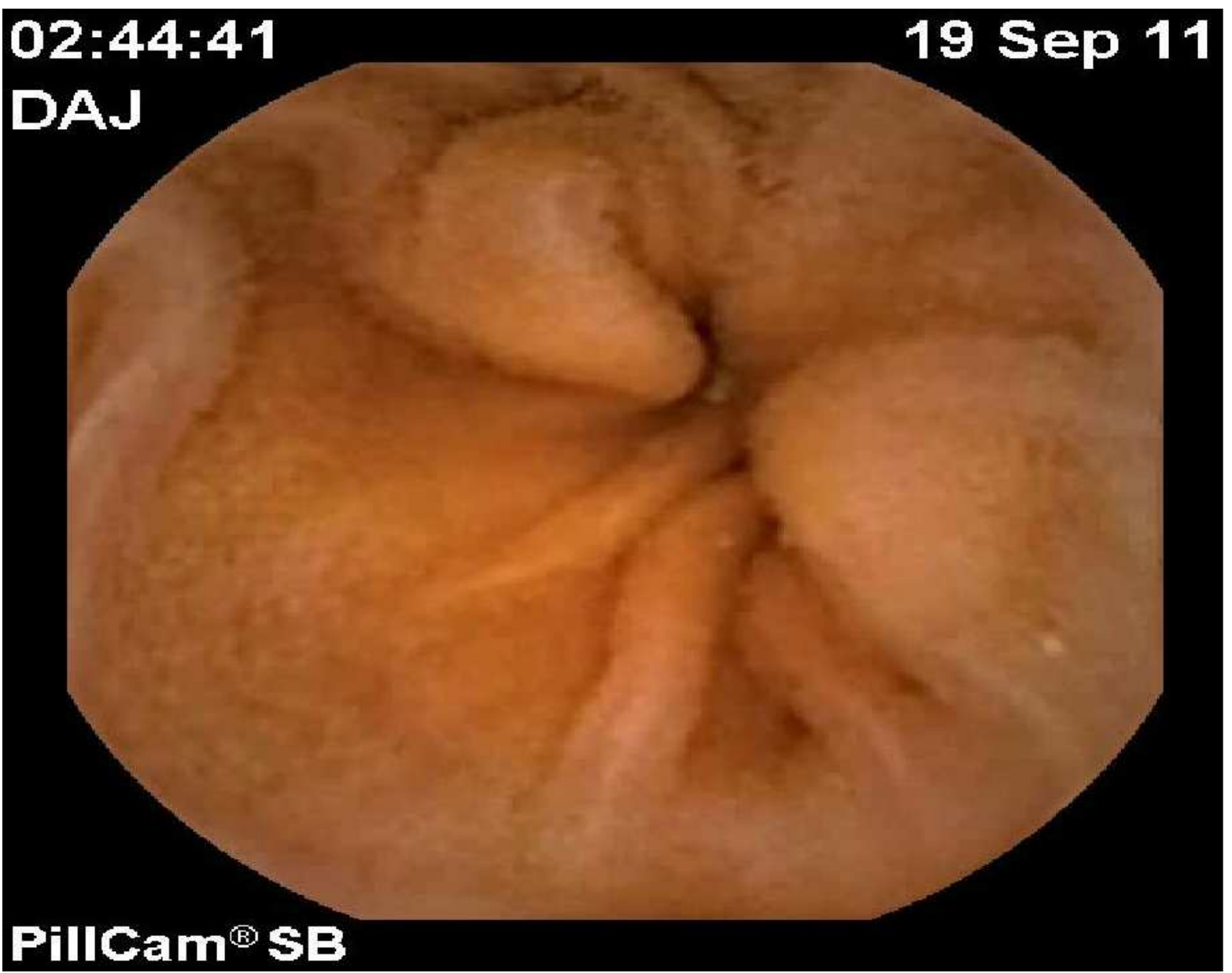}\hspace{2 mm}
\includegraphics[width=4.0cm,height=4.0cm]{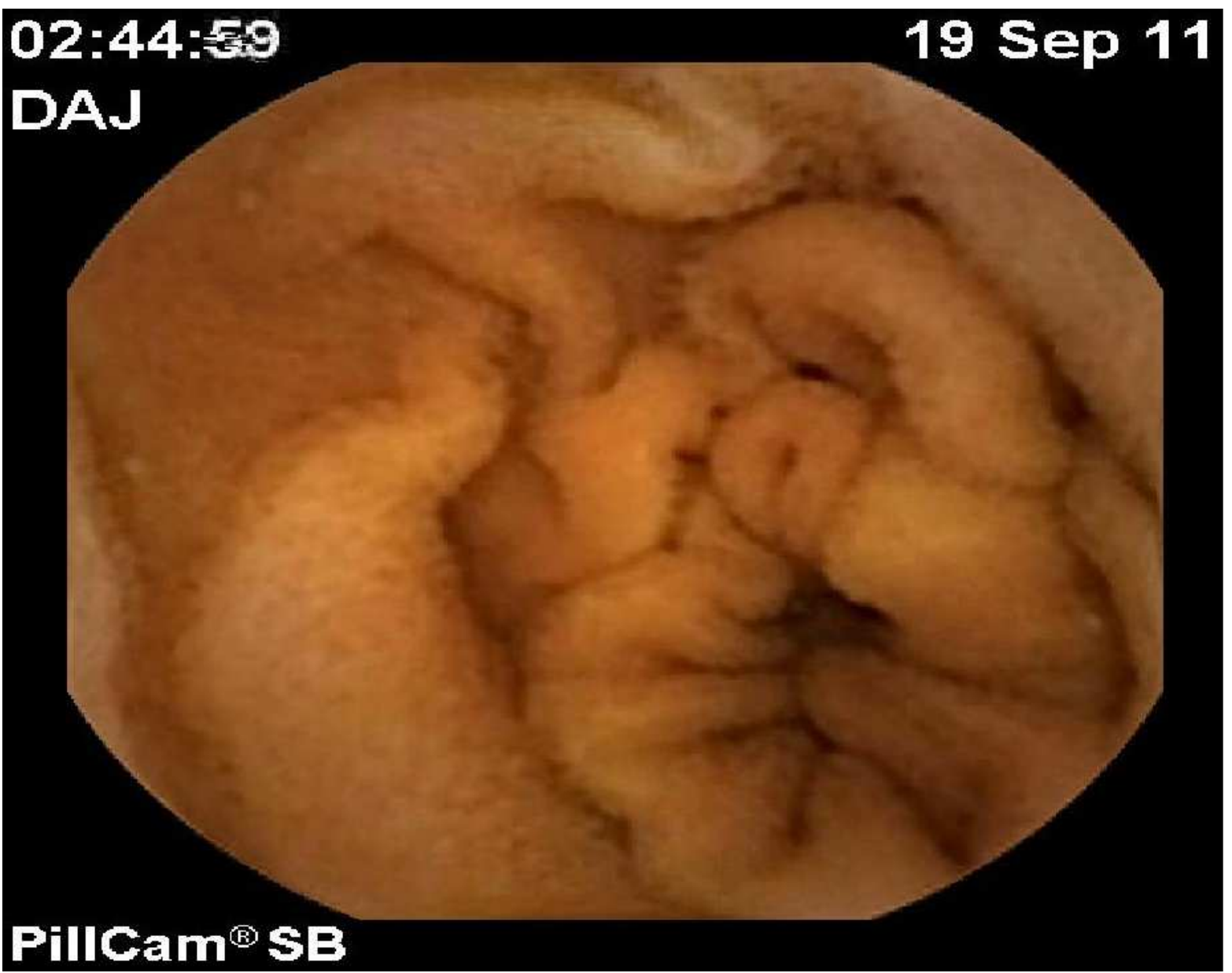}\\
\vspace{2 mm}
\includegraphics[width=4.0cm,height=4.0cm]{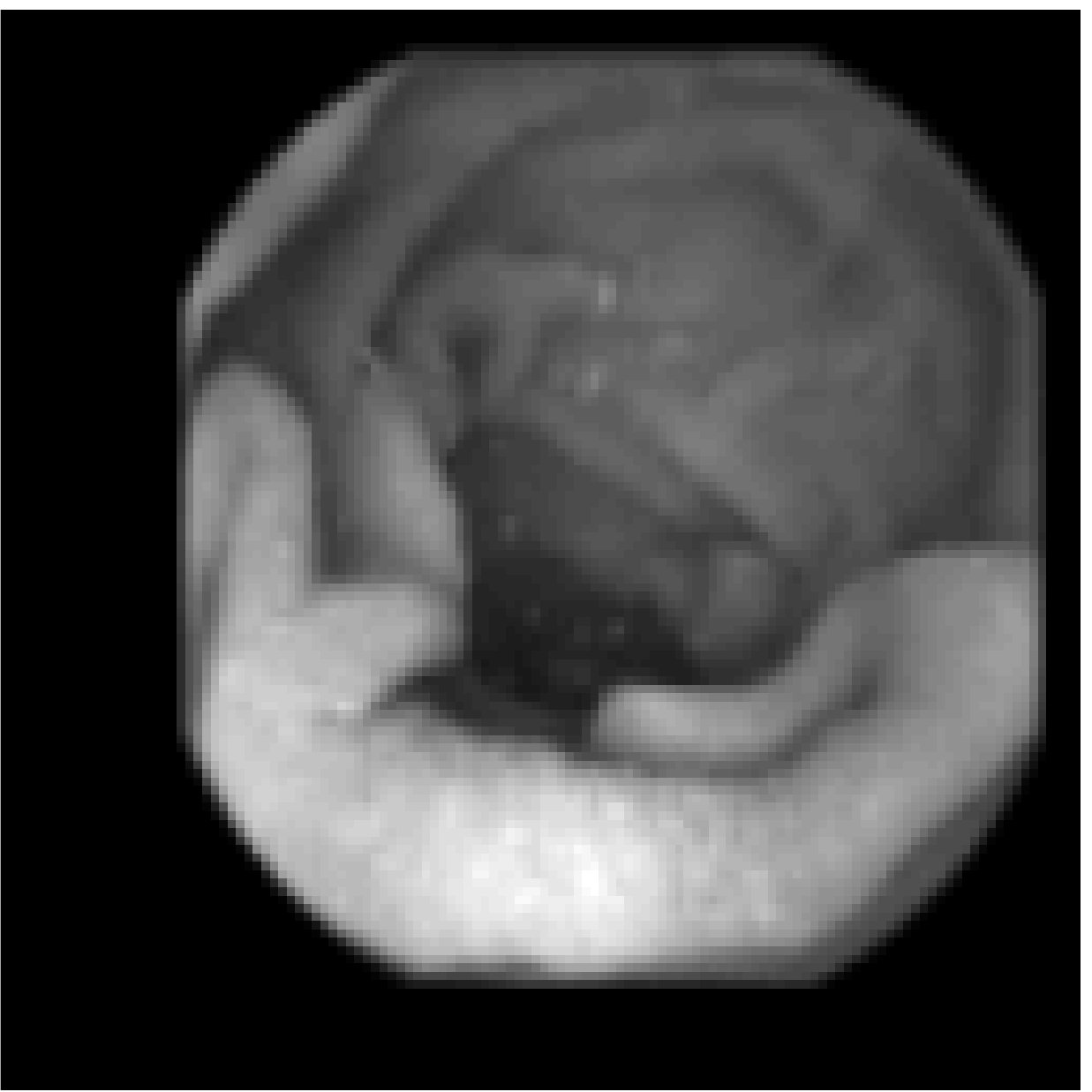}\hspace{2 mm}
\includegraphics[width=4.0cm,height=4.0cm]{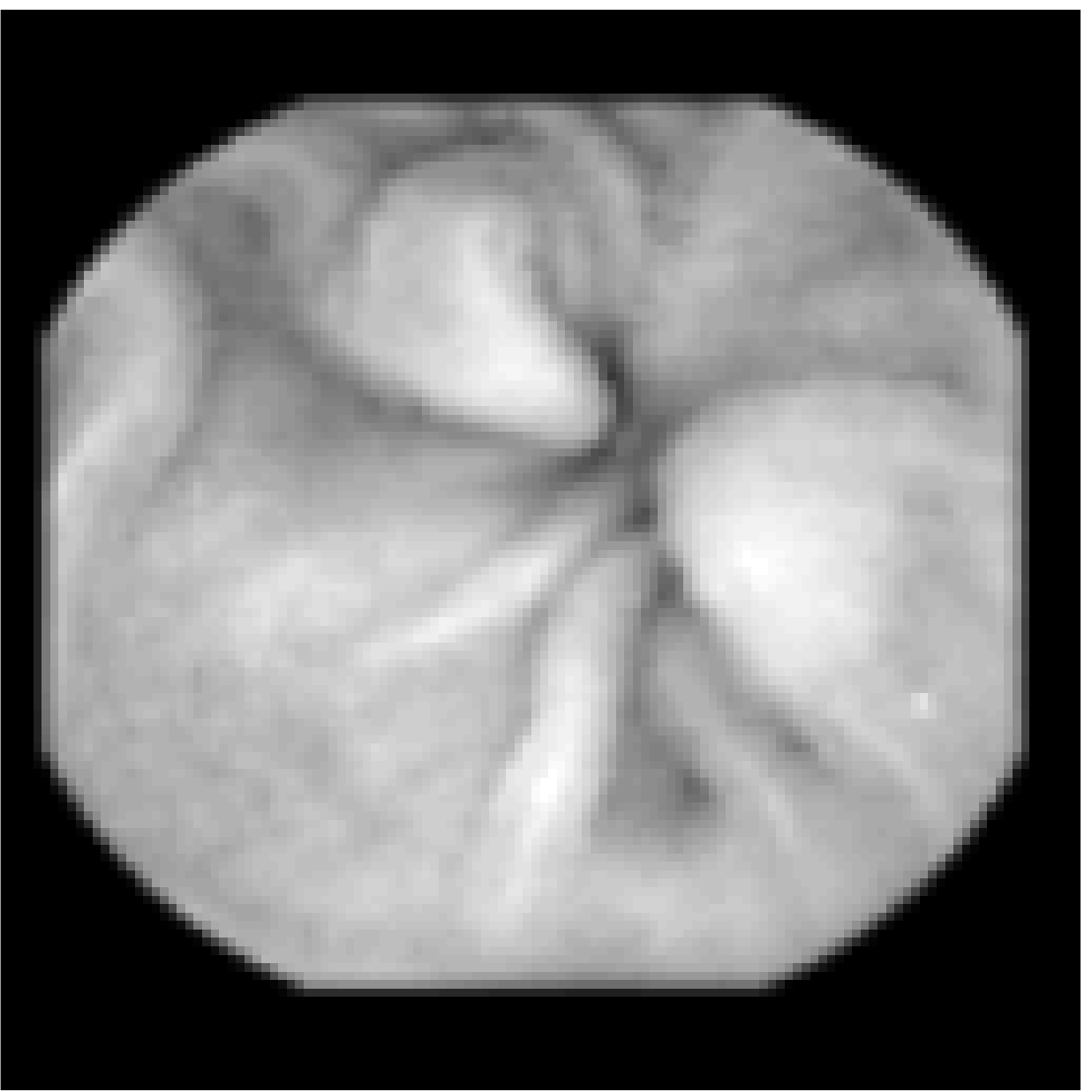}\hspace{2 mm}
\includegraphics[width=4.0cm,height=4.0cm]{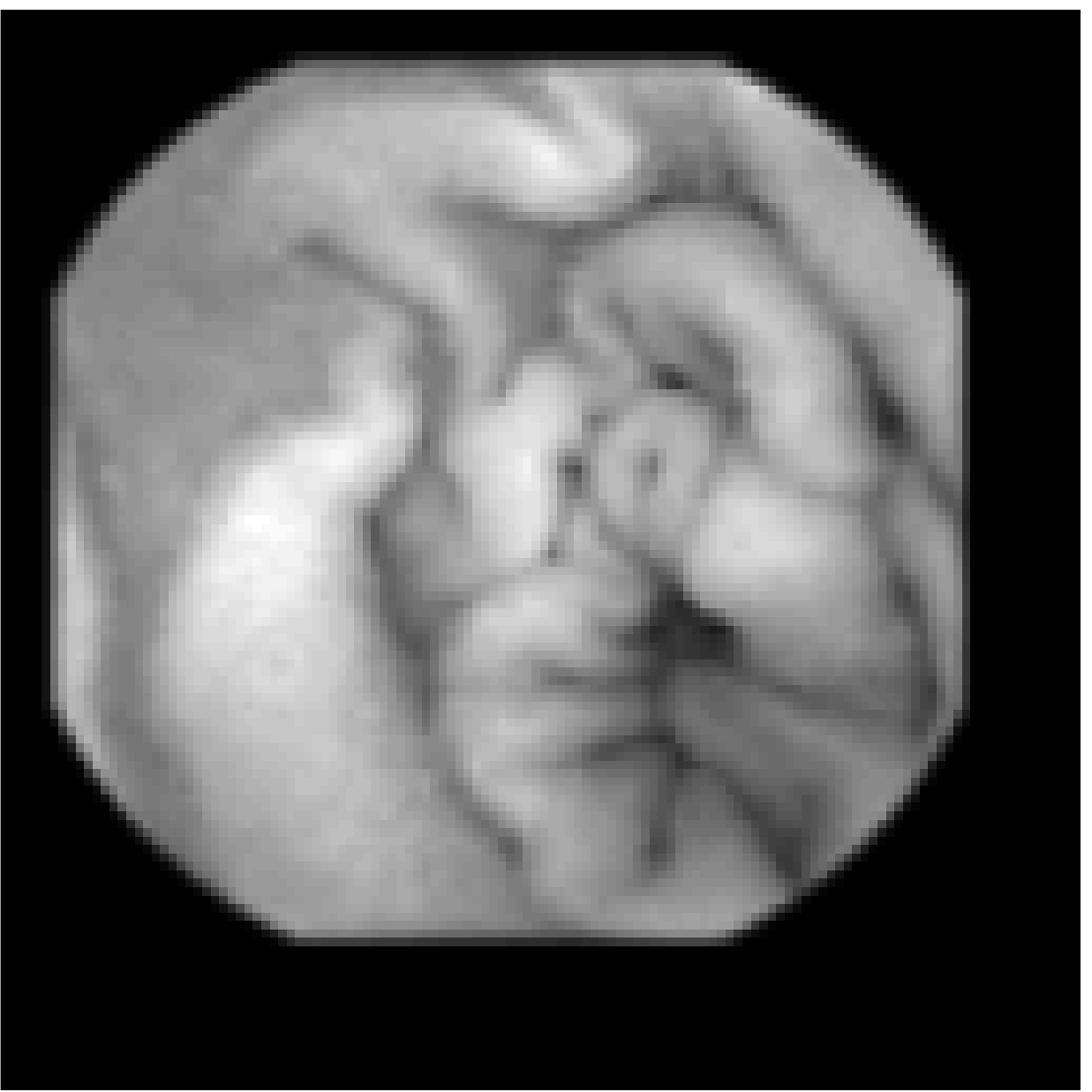}\\
\vspace{2 mm}
\includegraphics[width=4.0cm,height=4.0cm]{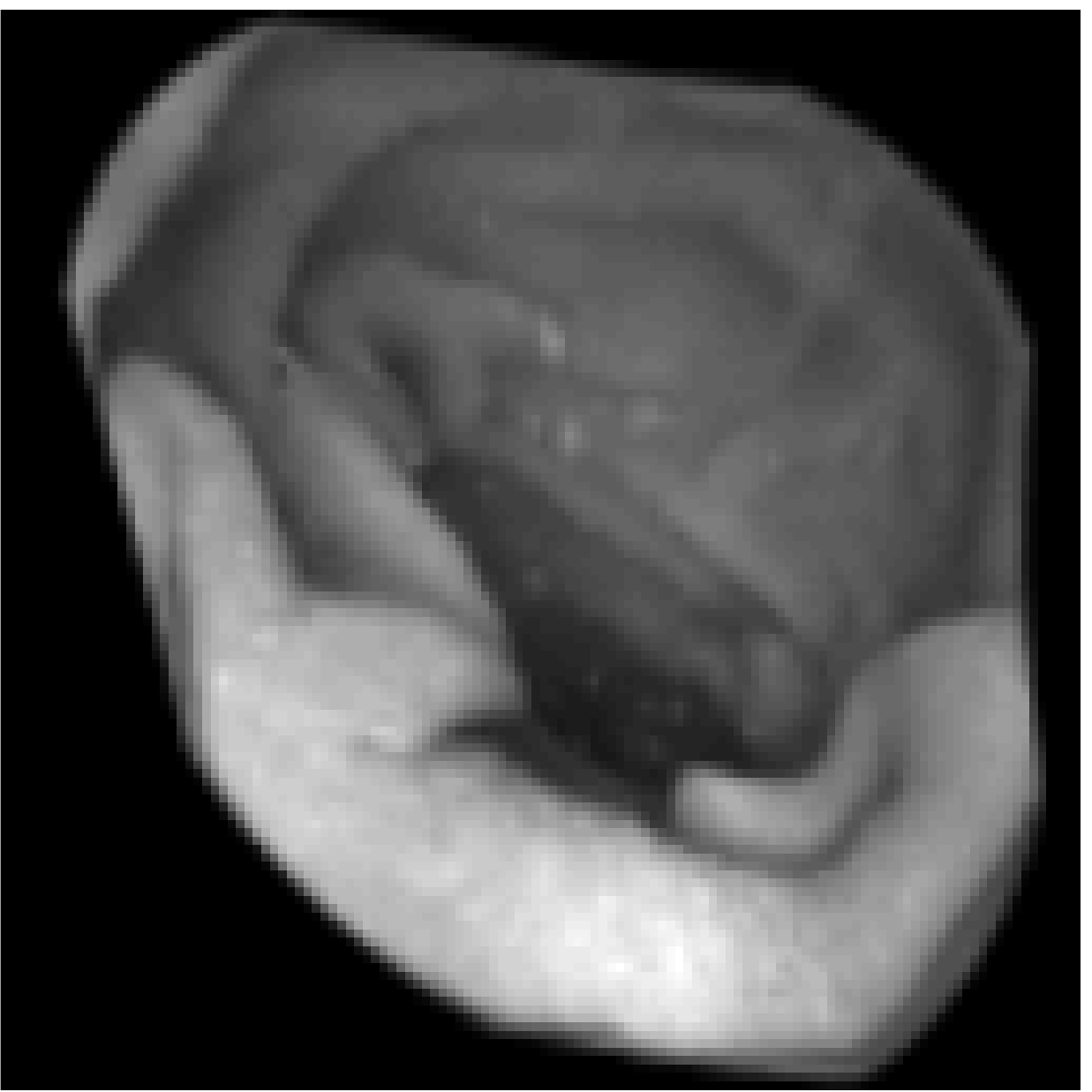}\hspace{2 mm}
\includegraphics[width=4.0cm,height=4.0cm]{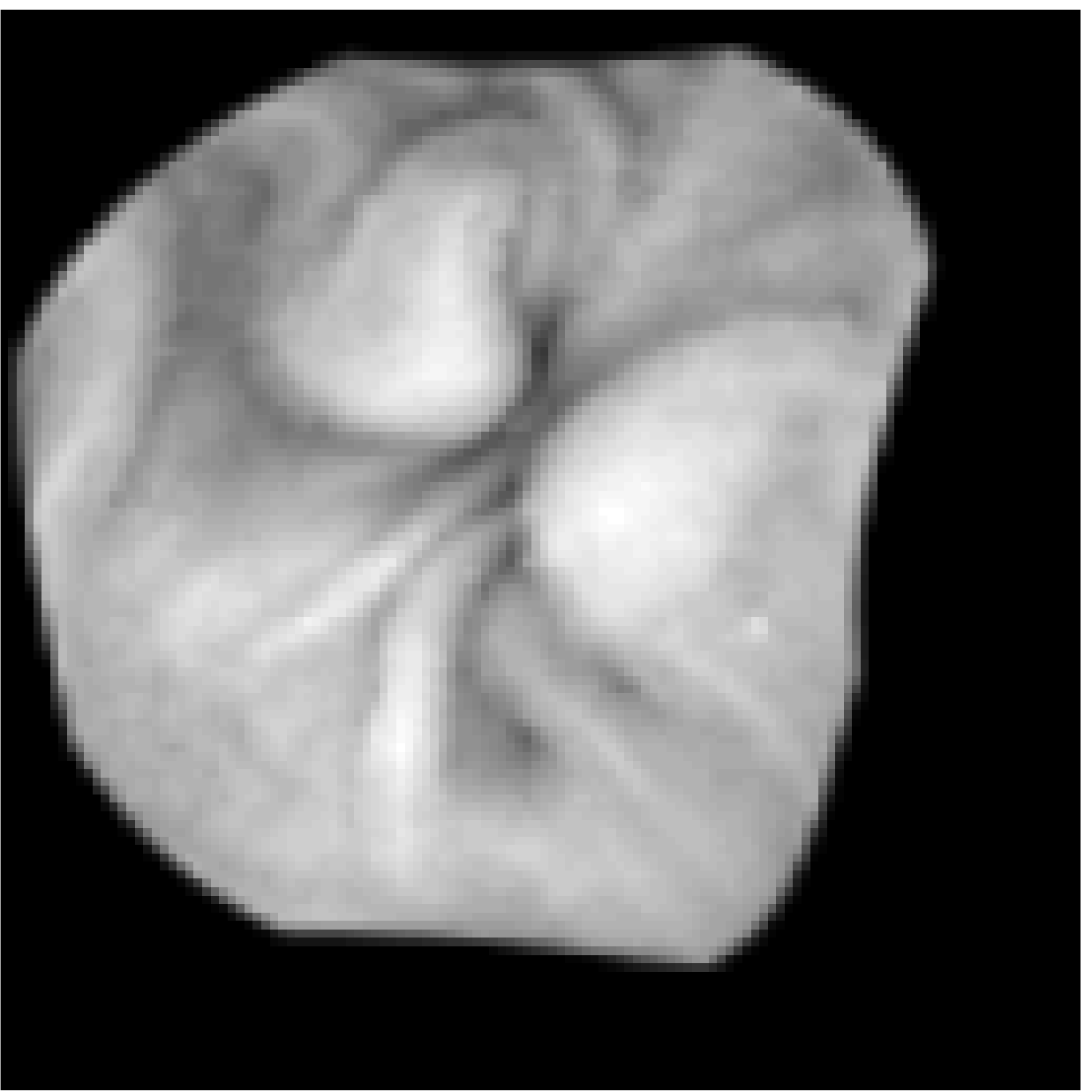}\hspace{2 mm}
\includegraphics[width=4.0cm,height=4.0cm]{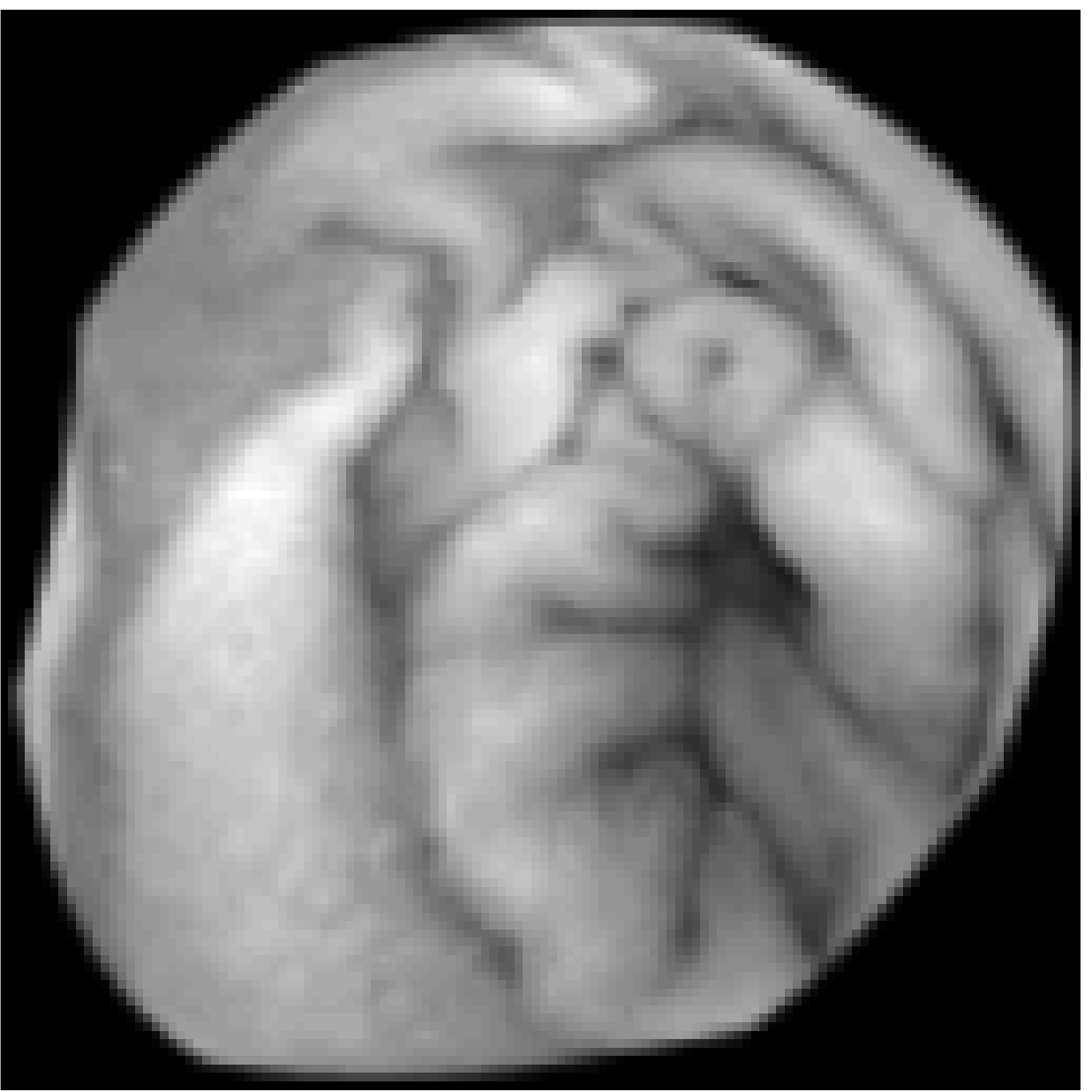}\\
\caption{In each column : original frame (top),  grayscale version (middle) and correspondent synthetic elastic deformed version (bottom).}\label{fig:elas}
\end{figure}

\subsubsection{Tests with elastic deformations}\label{sec:testelas}

We describe now the results provided by the tests performed with synthetic elastic deformations.
 We have generated the elastic deformation for a frame in the following way : a) First we define a 128 by 128 random matrix, whose components are pseudorandom values drawn from the standard uniform distribution on the open interval $(0,1)$ and  smooth this matrix by using a Gaussian filter.  b) Then we create a perturbed grid by adding the previous matrix to the regular grid of the image domain $\Omega=[0,1]\times [0,1]$,  with $128\times 128$ points.  c) Finally,  the elastically  deformed version of the image is obtained by interpolating the image on this perturbed grid. This procedure is repeated for all the 780  images
of the dataset. Therefore, a unique elastic deformation is associated with each image.
 The Figure \ref{fig:elas} depicts several  grayscale original frames and the corresponding elastic deformed versions by the aforementioned procedure.

 The result of the first experiment is shown in Figure \ref{fig:incr_elast}. It displays a comparison, for a single frame, between the $NDM$ curves obtained with MEIR and MPIR as the amount of elastic deformation (induced artificially) increases.
 The graphic corresponds  to the registration results for a single frame (displayed on the top right) whose grayscale version  (displayed on the bottom left)  is always the reference image $R$.  The different templates are deformed versions of the reference image $R$,  generated by increasing the amount of elastic deformation (and also by applying   a  rotation angle of   $10$ and a change of scale with scale factor  $0.8$). The vertical axis represents the $NDM$ values and the horizontal axis the intensity of elastic deformations, by increasing order.
 The results of   $NDM$  for MEIR and MPIR with the deformed images exhibited in the third column as templates, correspond to the left and right, respectively, vertical dashed lines in the middle graph. The amount of elastic deformation applied to generate the top and bottom frames, denoted by $T_t$ and $T_b$ respectively and represented   in the third column, are indicated by the left and right vertical dashed lines,  respectively,  in the middle graph. The intersection of these vertical lines with the curves  are the NDM  the results for MEIR and MPIR. Obviously this  graphic  reinforces the advantage  of the MEIR approach over
the MPIR approach,  when there are elastic deformations involved.  Figure \ref{fig:incr_elast_2} illustrates the MPIR and MEIR results for the reference $R$ and two template images $T_t$ (a weak elastic deformation of $R$) and $T_b$  (a strong elastic deformation of $R$)  shown in Figure \ref{fig:incr_elast}. These results clearly demonstrate the superiority  of MEIR over MPIR, when the amount of elastic deformation increases.

\begin{figure}[t!]
\centering
\begin{minipage}[c][7.5cm][c]{0.23\textwidth}
\centering
  \includegraphics[width=3.5cm,height=3.5cm]{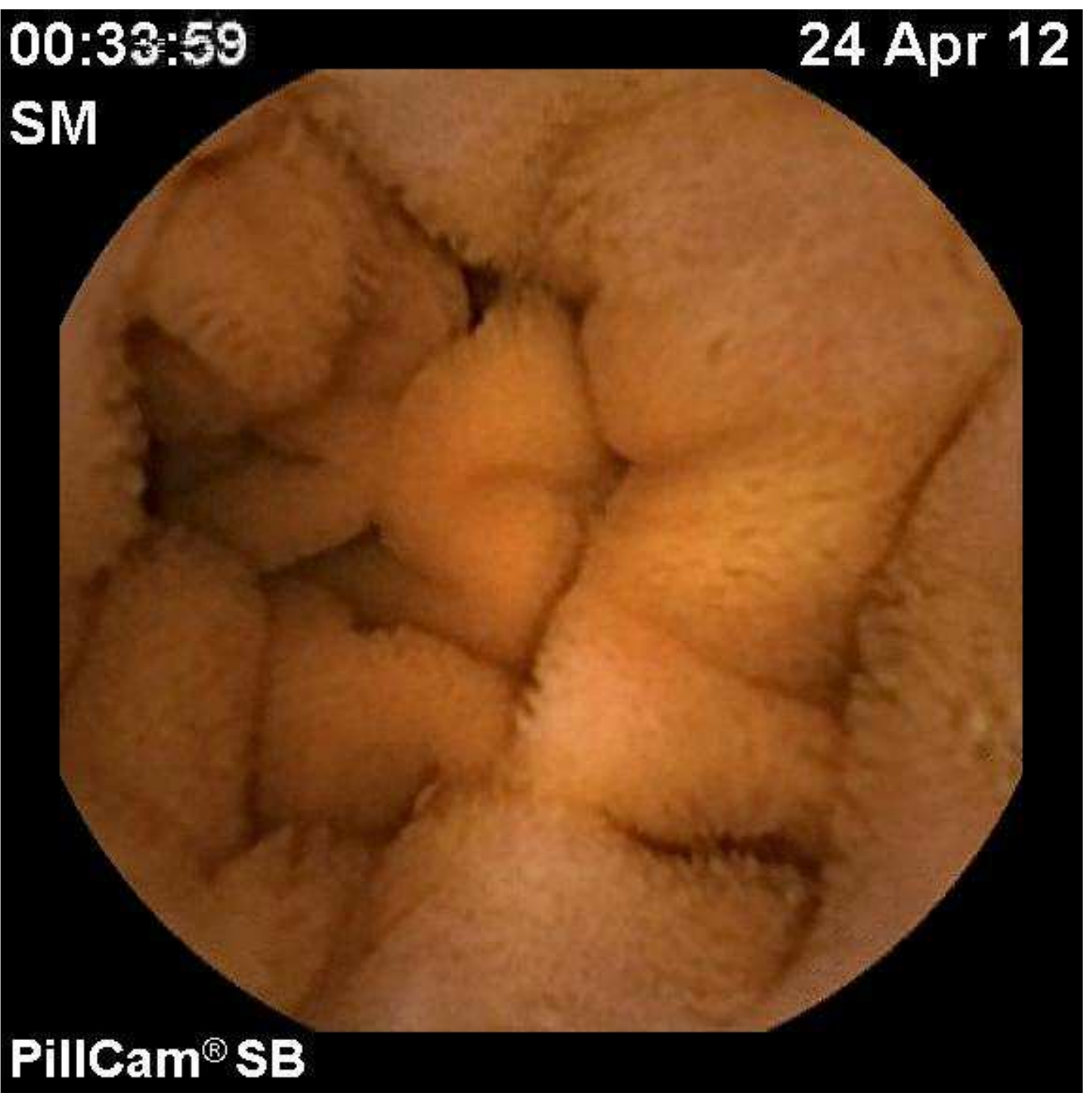}\par \vfill
  \includegraphics[width=3.5cm,height=3.5cm]{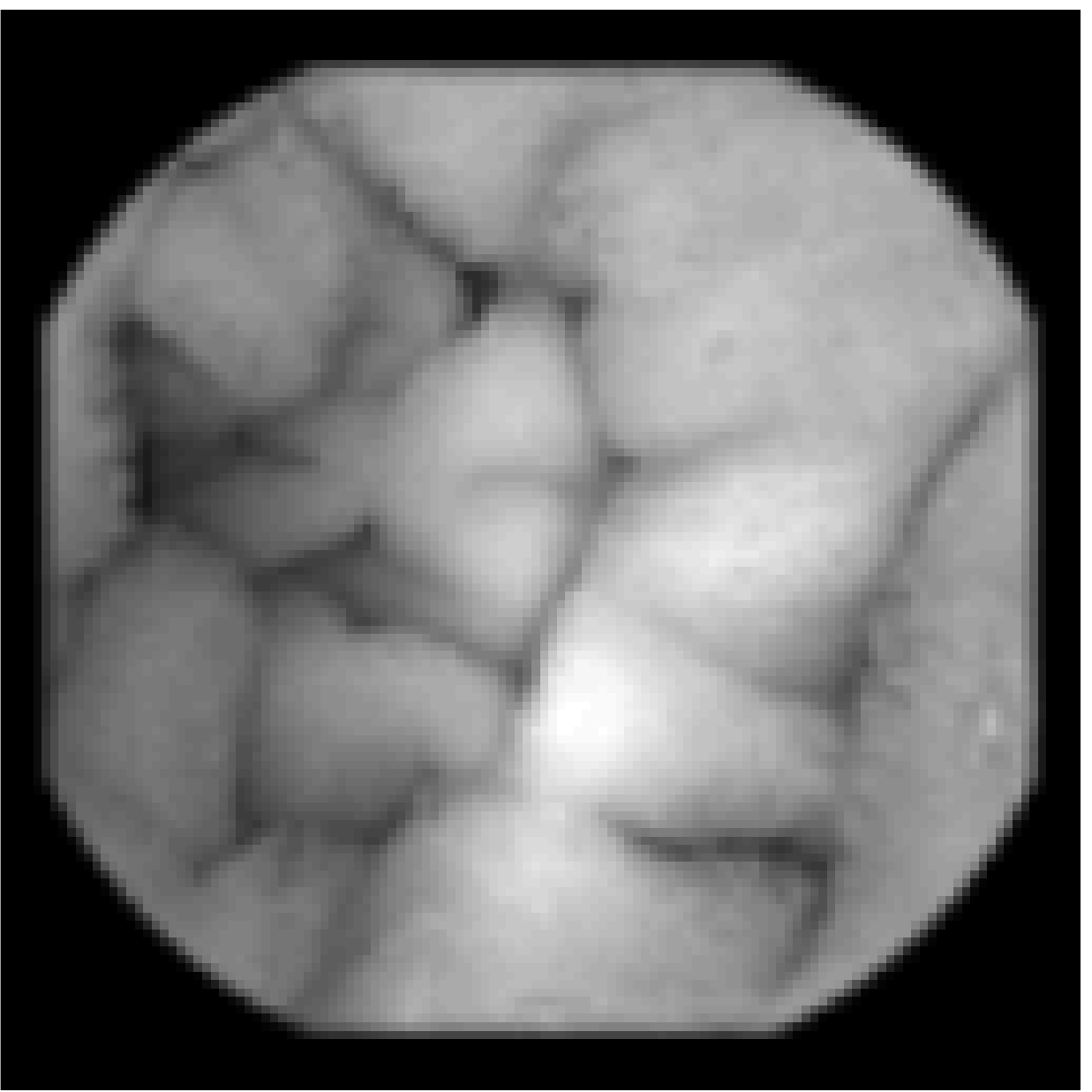}
\end{minipage}
\begin{minipage}[c][7.5cm][c]{0.5\textwidth}
\centering
  \includegraphics[width=7.5cm,height=7.5cm]{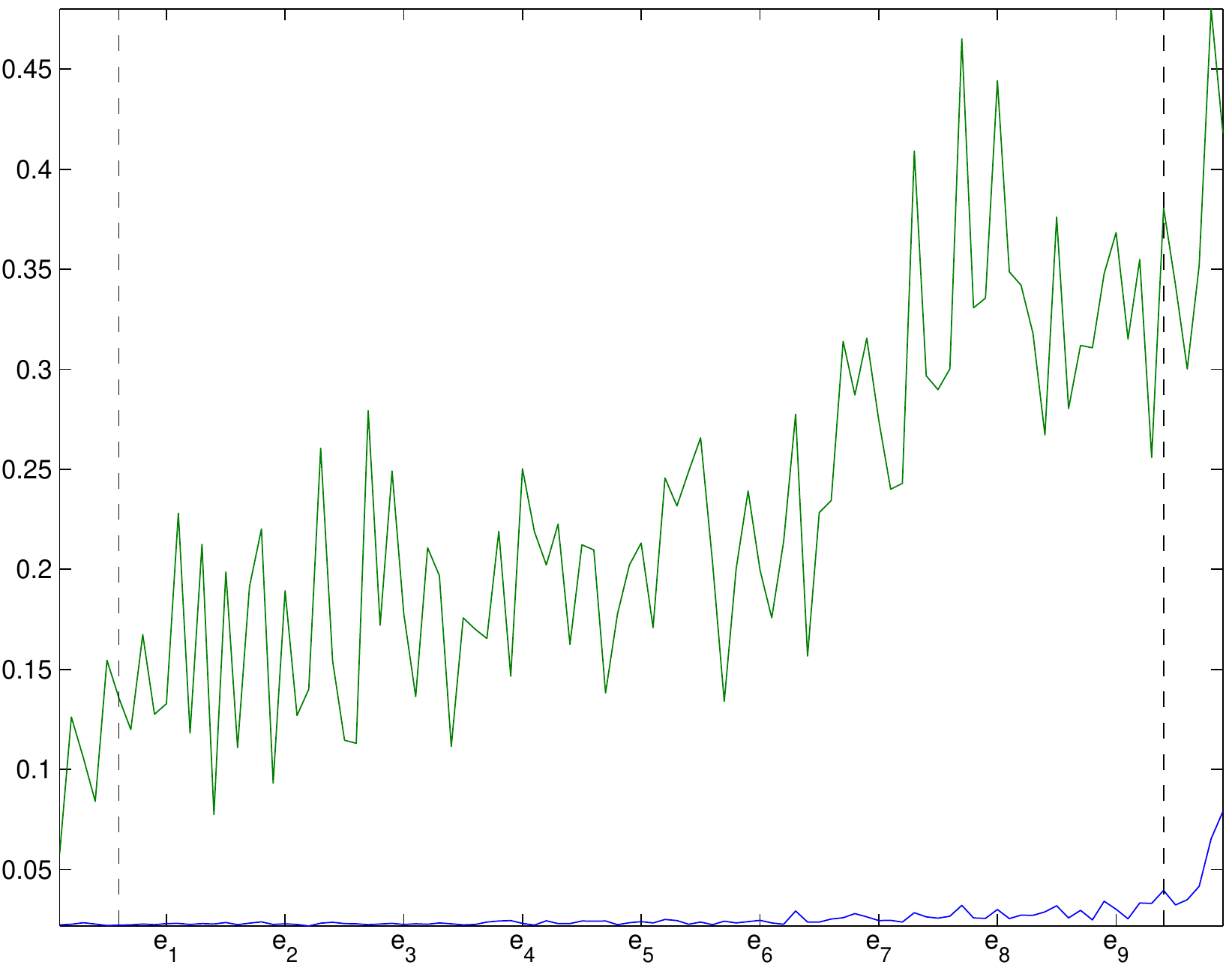}
\end{minipage}
\begin{minipage}[c][7.5cm][c]{0.23\textwidth}
\centering
  \includegraphics[width=3.5cm,height=3.5cm]{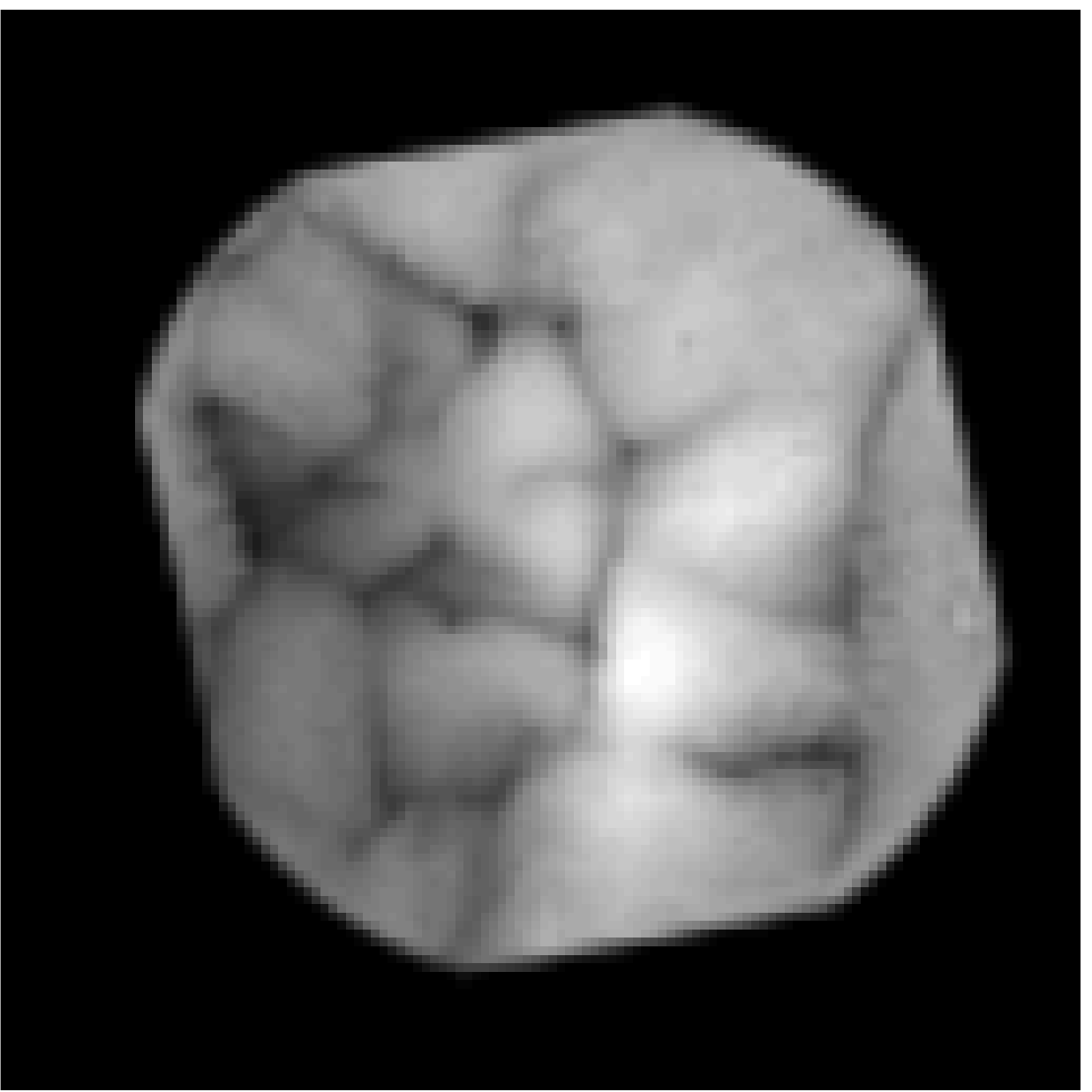}\par \vfill
  \includegraphics[width=3.5cm,height=3.5cm]{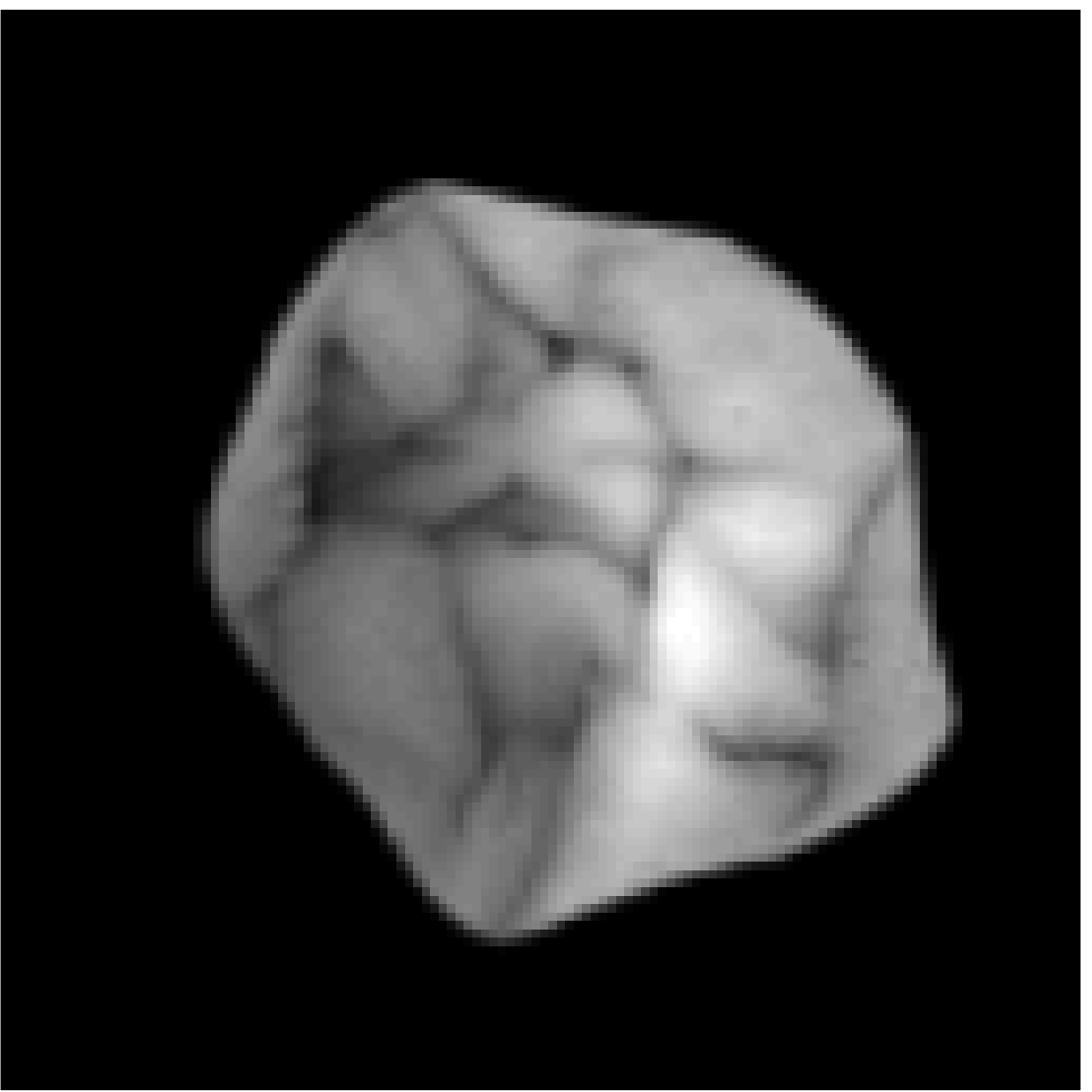}
\end{minipage}
\caption{Middle graphic: Comparison for a single frame (shown in the top right) between the $NDM$ curves obtained with MEIR (blue curve) and MPIR (green curve) as the amount of elastic deformation (induced artificially) increases. The parameter, $e_i$, $i=1,\ldots, 9$, represents the intensity of elastic deformation.
First  column: Original frame and its grayscale version (the reference image $R$).
Third column: examples of two template images that are  synthetically,  scaled, rotated and elastic deformed versions of the reference image $R$. The template on the top, $T_t$,  corresponds to a weak elastic deformation of $R$, while that on the bottom, $T_b$ to a strong elastic deformation of $R$.}
\label{fig:incr_elast}
\end{figure}

\begin{figure}[h!]
\centering
\includegraphics[width=3.5cm,height=3.5cm]{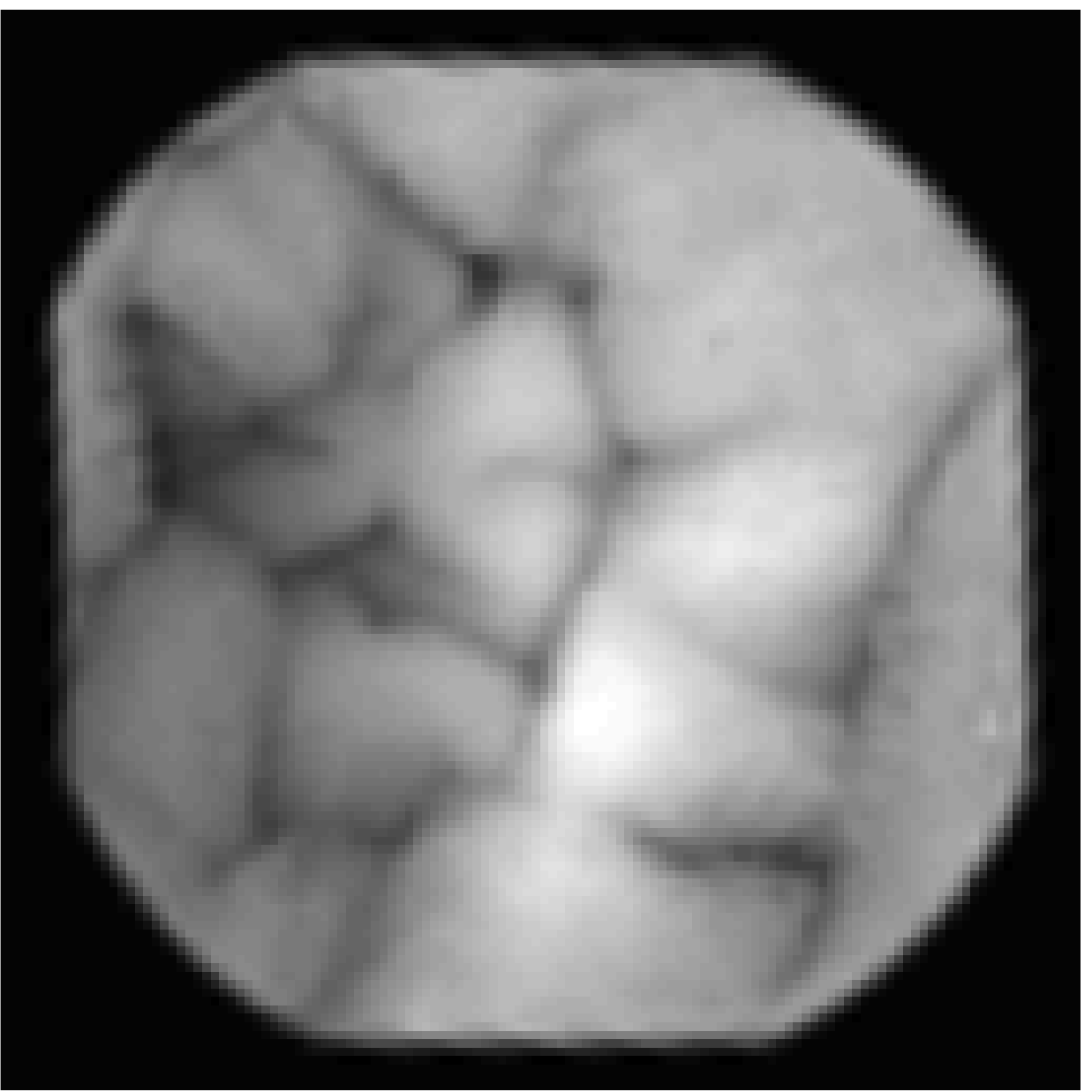}\hspace{2 mm}
\includegraphics[width=3.5cm,height=3.5cm]{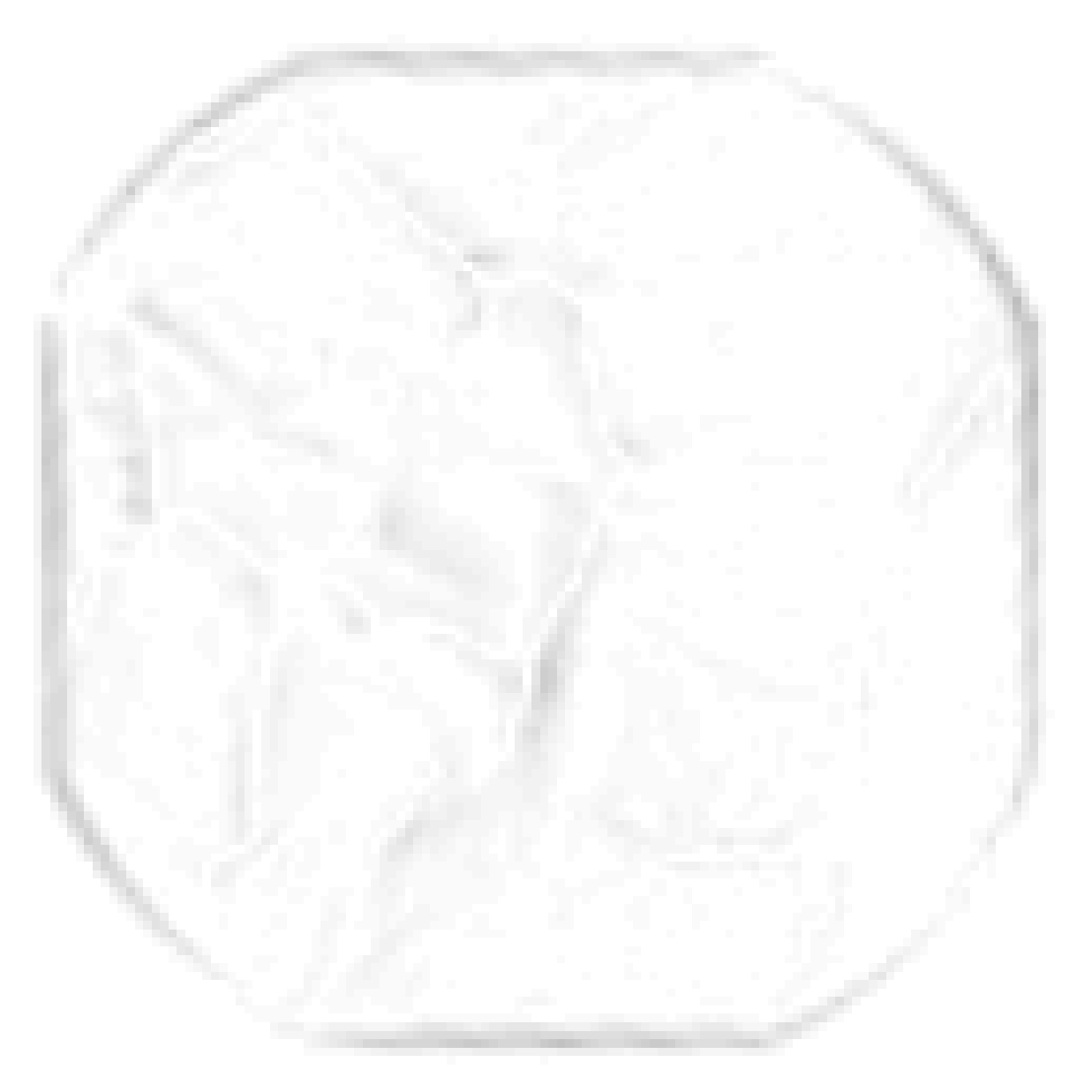}\hspace{2 mm}
\includegraphics[width=3.5cm,height=3.5cm]{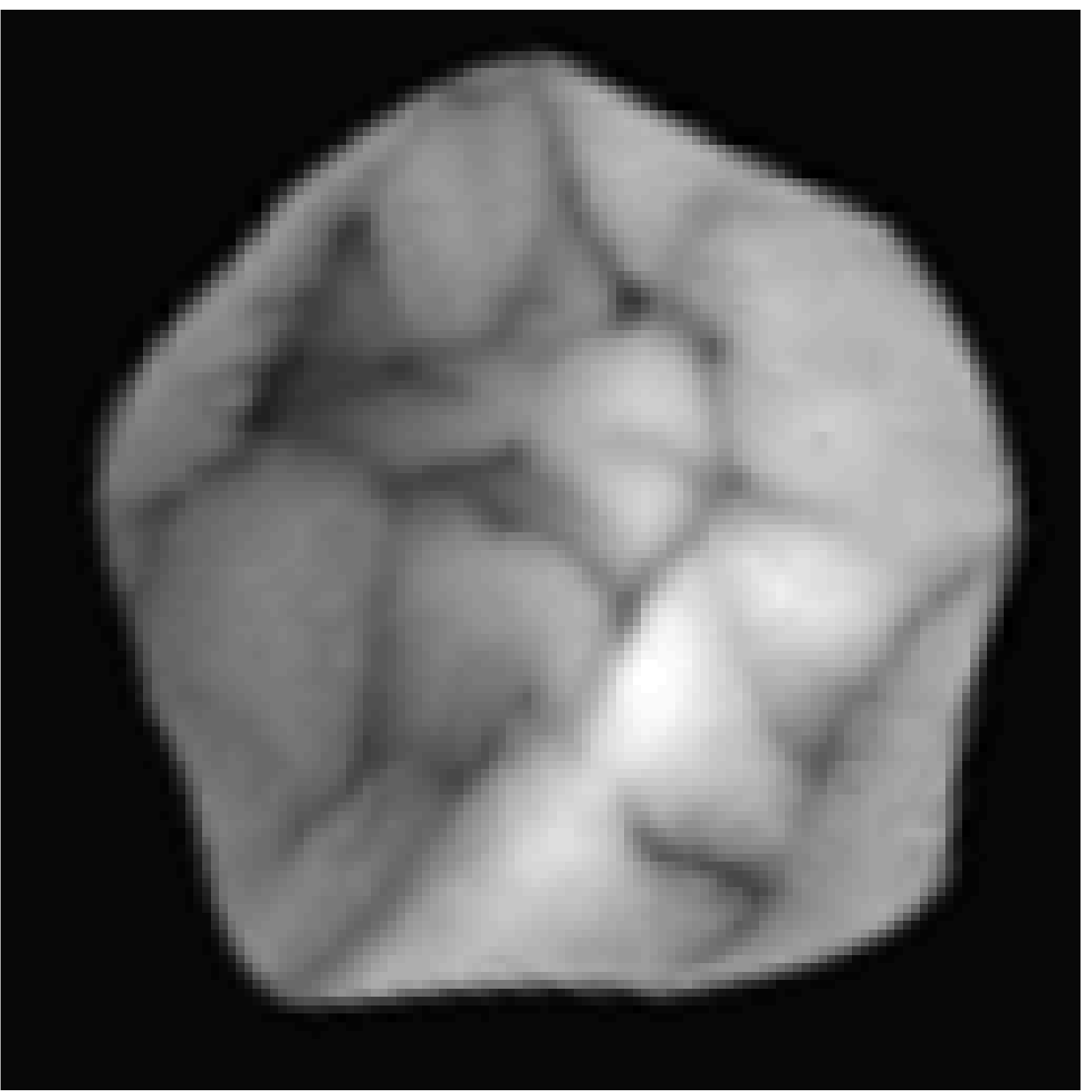}\hspace{2 mm}
\includegraphics[width=3.5cm,height=3.5cm]{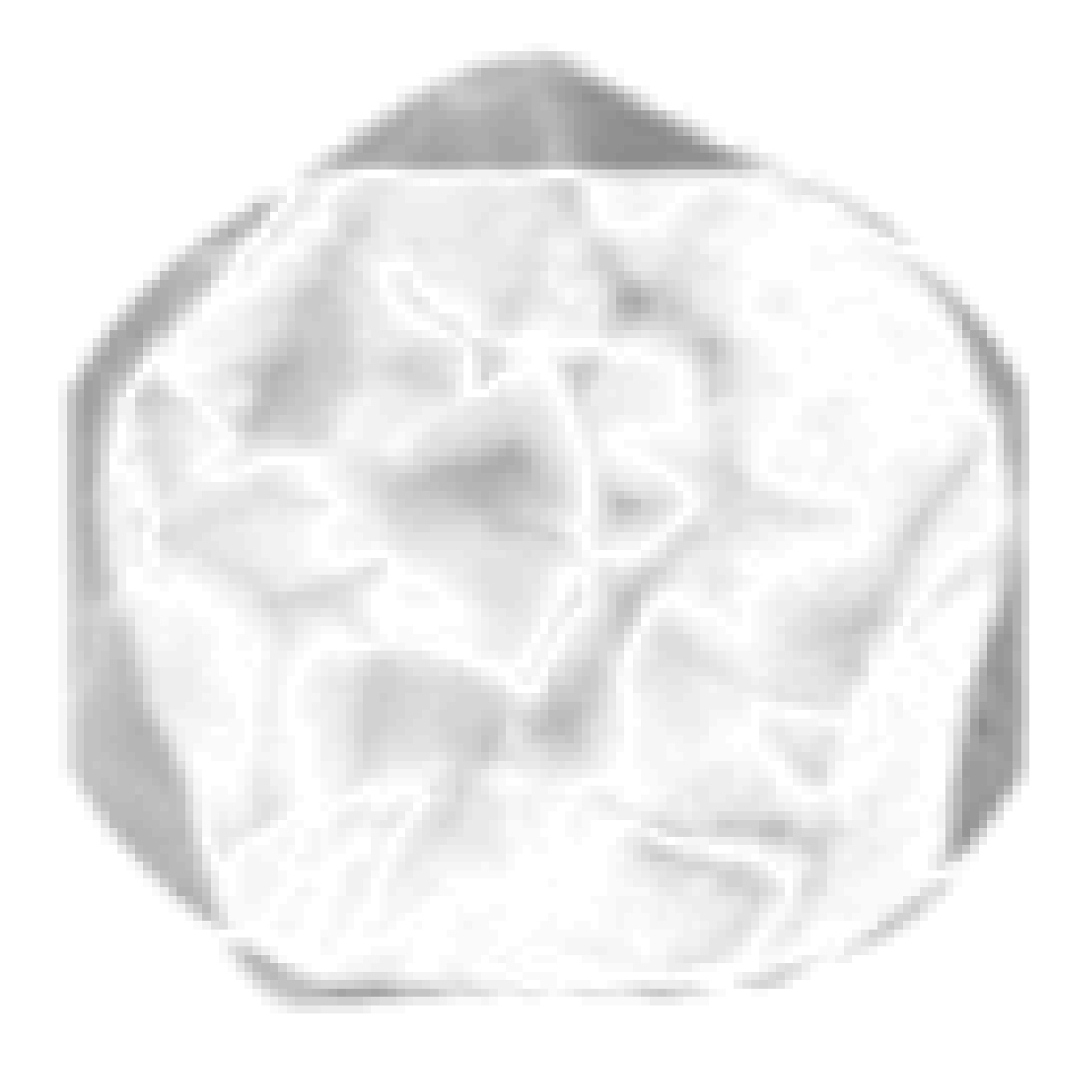}\\
\vspace{2 mm}
\includegraphics[width=3.5cm,height=3.5cm]{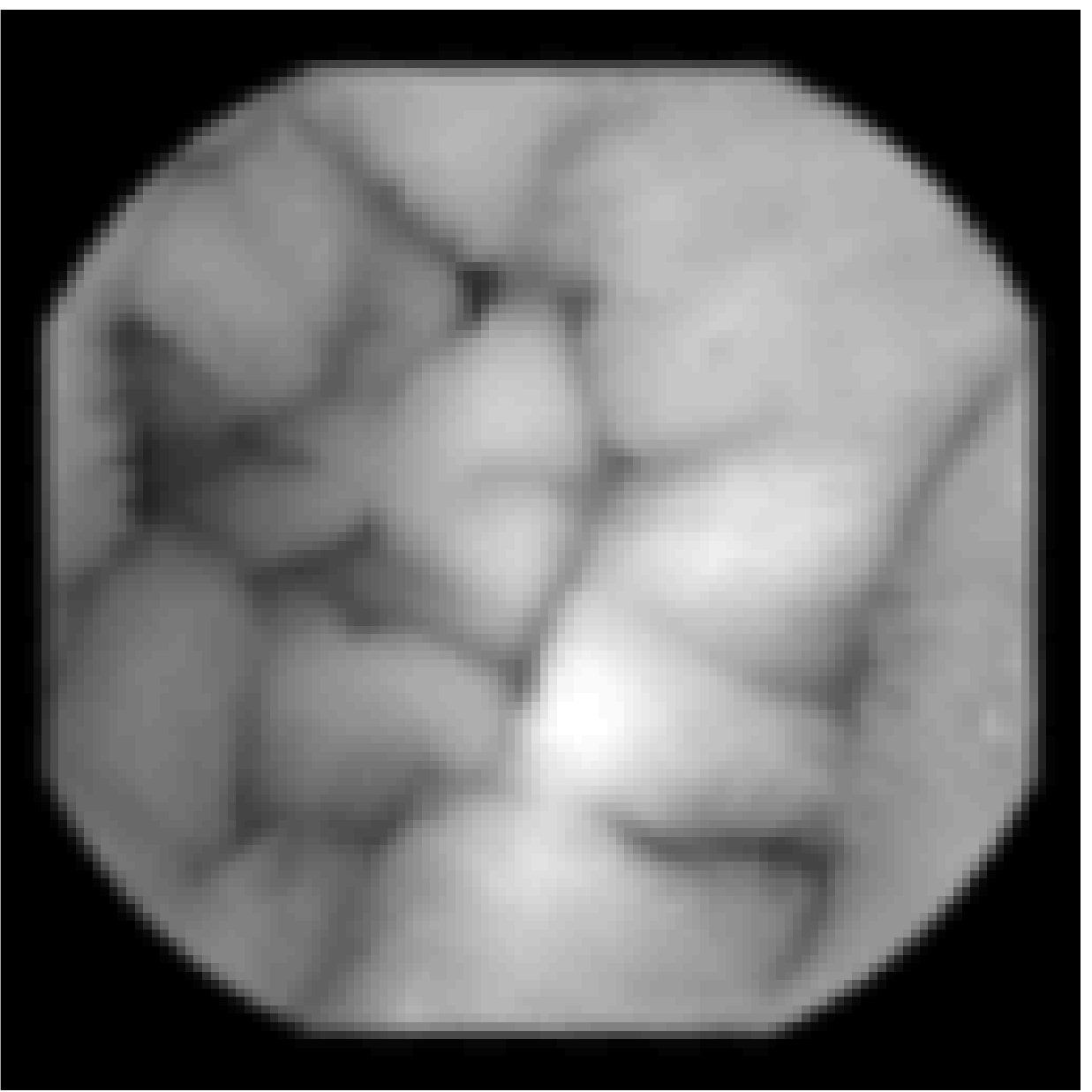}\hspace{2 mm}
\includegraphics[width=3.5cm,height=3.5cm]{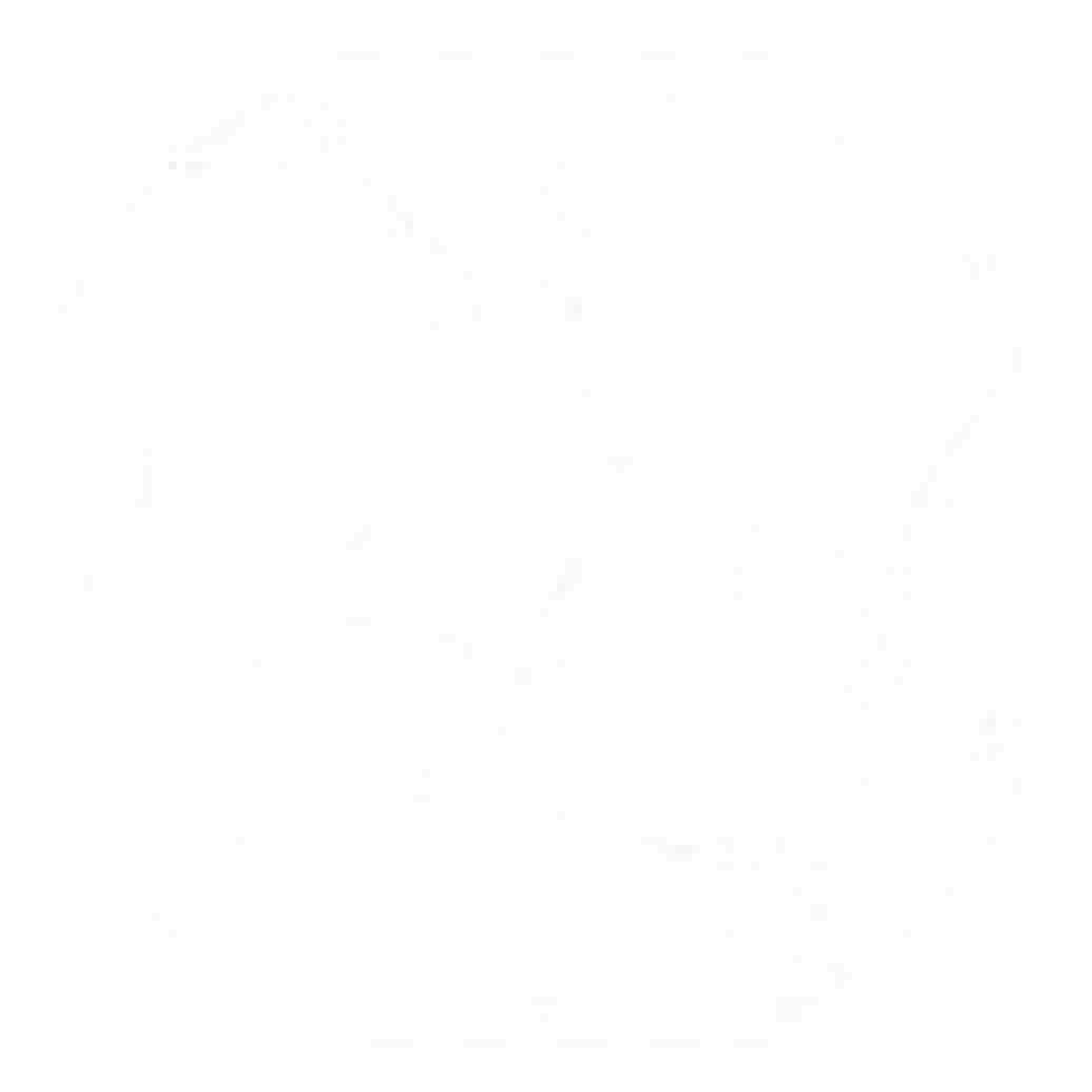}\hspace{2 mm}
\includegraphics[width=3.5cm,height=3.5cm]{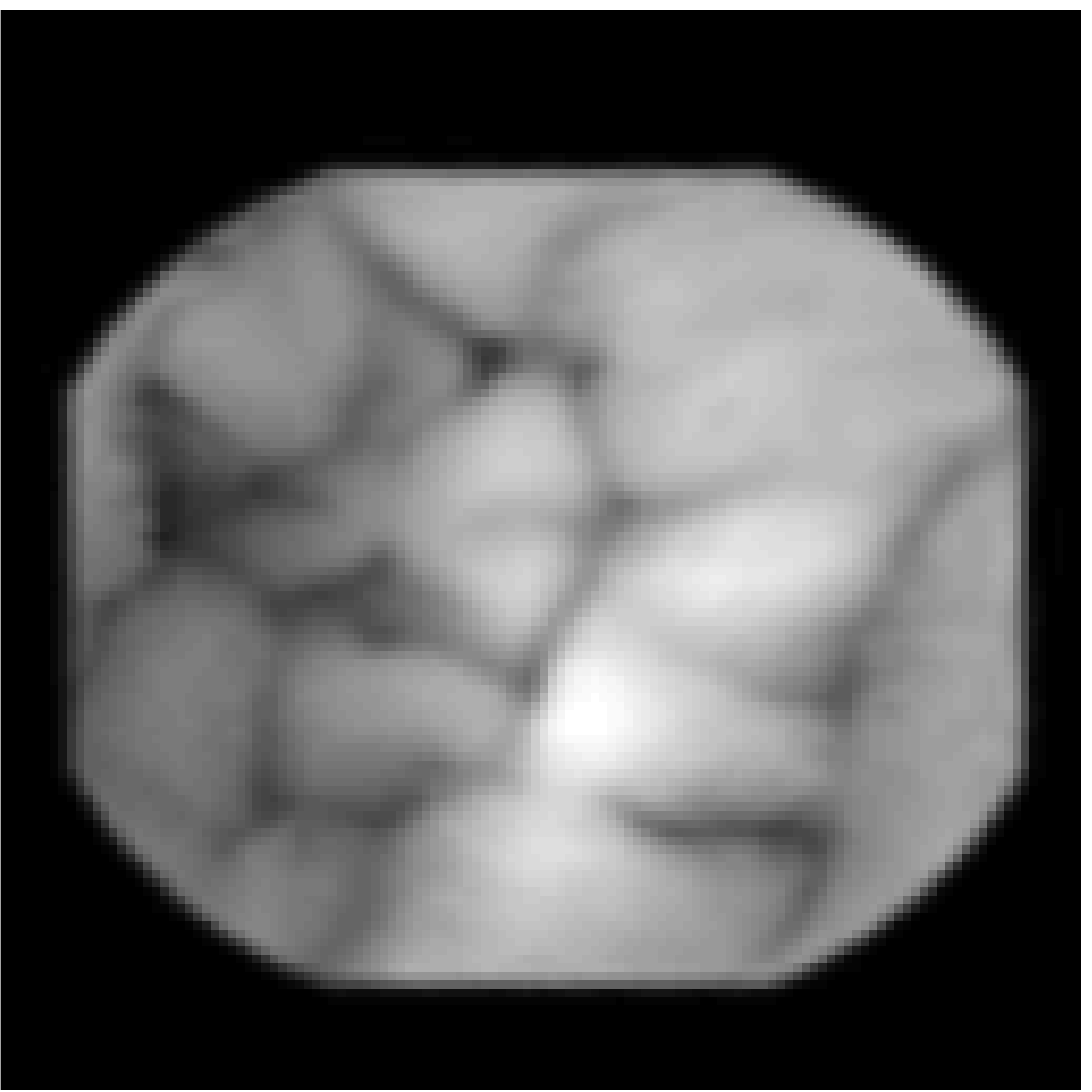}\hspace{2 mm}
\includegraphics[width=3.5cm,height=3.5cm]{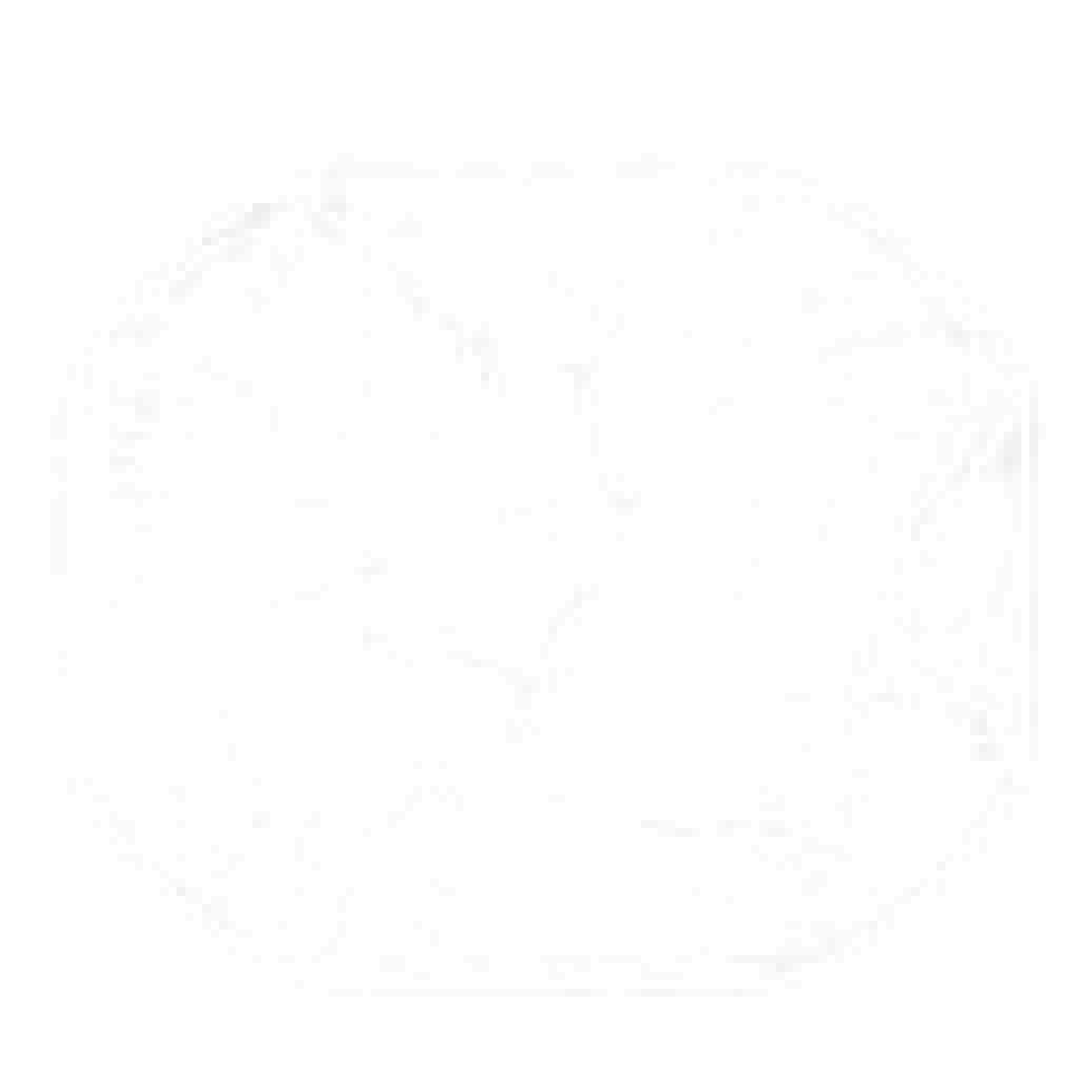}\\
\caption{First row:  MPIR results. Second row: MEIR results.
In each row (from left to right): $T(\varphi)$ (to compare with the reference image) and difference between $R$ and $T(\varphi)$ for template  $T_t$  in Figure \ref{fig:incr_elast} ; $T(\varphi)$  (to compare with the reference image)  and difference between $R$ and $T(\varphi)$ for  template $T_b$   in Figure \ref{fig:incr_elast}. }\label{fig:incr_elast_2}
\end{figure}

After this first experiment, four types of synthetic frames were generated, using for each type the 780 frames : Case i) applying an elastic deformation only, at the original scale and original orientation. Case ii) applying  a rotation and an elastic deformation at the original scale. Case iii) applying  a scale factor and an elastic deformation at the original orientation. Case iv)  applying  a rotation, a scale factor and an elastic deformation.

The results of the tests for the cases i) to iv) are displayed in Tables \ref{tab:elas}, \ref{tab:orie_elas} and \ref{tab:sca_elas}, for i), ii) and iii) respectively,  and for iv) in Table
 \ref{tab:sca_orie20_elas}, where the rotation angle $\omega_1$ is fixed at $20$, and in Table  \ref{tab:sca1.4_orie_elas},  where the scale factor $\omega_0$ is kept fixed at $1.4$ (the errors listed in the tables are always mean absolute value errors).

 \begin{table*}[h!]
\renewcommand{\arraystretch}{1.3}
\caption{Case i) at the original scale and orientation}
\label{tab:elas}
\centering
 \begin{tabular}{|c|c||c|c||c|c|}\hline
   \multicolumn{2}{|c||}{$ NDM$} &  \multicolumn{2}{c||}{Mean Scale  Error}   &  \multicolumn{2}{c|}{Mean Rotation  Error}    \\ \hline
   MEIR & MPIR &    MEIR & MPIR   &    MEIR & MPIR \\  \hline   \hline
0.077865 & 0.305940	& 0.046420	& 0.050328	& 4.111000 & 4.821000 \\ \hline
\end{tabular}
\end{table*}

 As shown in these tables, the normalized dissimilarity measure $NDM$ is always better for MEIR than for MPIR. A similar results is true for  the mean (absolute value) errors,  either for the scale or the rotation angle, that is,  the performance of  MEIR is always superior to MPIR. This conclusion was somewhat expected, since the MEIR approach is obviously more convenient than MPIR, when elastic deformations are involved.

\begin{table*}[h]
\renewcommand{\arraystretch}{1.3}
\caption{Case ii) at the original scale}
\label{tab:orie_elas}
\centering
 \begin{tabular}{|c||c|c||c|c||c|c|}\hline
    Rotation  & \multicolumn{2}{|c||}{$NDM$} &  \multicolumn{2}{c||}{Mean Scale  Error}   &  \multicolumn{2}{c|}{Mean Rotation  Error}    \\ \cline{2-7}
     $\omega_1$                    &  MEIR & MPIR &     MEIR & MPIR   &    MEIR & MPIR \\  \hline   \hline
5  & 0.077690 & 0.299270 & 0.041627 & 0.044040 & 4.285200 & 4.982500 \\\hline
10 & 0.081106 & 0.303860 & 0.044879 & 0.047900 & 4.001400 & 4.710200 \\\hline
15 & 0.080859 & 0.304740 & 0.044368 & 0.048056 & 4.319300 & 5.013800 \\\hline
20 & 0.086114 & 0.304760 & 0.045060 & 0.048703 & 3.853000 & 4.521800 \\\hline
25 & 0.090683 & 0.306830 & 0.044147 & 0.046658 & 4.439000 & 5.129100 \\\hline
30 & 0.095251 & 0.306230 & 0.045137 & 0.048883 & 4.748100 & 5.269300 \\\hline
\end{tabular}
\end{table*}

 \begin{table*}[h!]
\renewcommand{\arraystretch}{1.3}
\caption{Case iii) at the original orientation}
\label{tab:sca_elas}
\centering
 \begin{tabular}{|c||c|c||c|c||c|c|}\hline
   Scale  & \multicolumn{2}{|c||}{$NDM$} &  \multicolumn{2}{c||}{Mean Scale  Error}   &  \multicolumn{2}{c|}{Mean Rotation  Error}    \\ \cline{2-7}
    $\omega_0$                     &  MEIR & MPIR &     MEIR & MPIR   &    MEIR & MPIR \\  \hline   \hline
0.4 & 0.119440 & 0.287370 & 0.018118 & 0.019389 & 4.436500 & 5.153900 \\ \hline
0.6 & 0.109320 & 0.292800 & 0.026897 & 0.028673 & 4.198000 & 4.926600 \\ \hline
0.8 & 0.091956 & 0.298420 & 0.035632 & 0.038818 & 4.360100 & 5.141100 \\ \hline
1.2 & 0.118630 & 0.304190 & 0.055755 & 0.058641 & 4.764600 & 5.172400 \\ \hline
1.4 & 0.172400 & 0.295720 & 0.066795 & 0.069305 & 4.494900 & 4.714600 \\ \hline
\end{tabular}
\end{table*}

We remark that an elastic deformation always embodies a change in scale and generates a rotation, as illustrated in  the examples depicted  in Figure \ref{fig:elas}. There we can see that for two frames there is an evident rotation associated to the elastic deformation, and for one frame a change of scale is also obvious.
This   is the reason why in Table \ref{tab:elas} we have measured the scale and rotation errors, for  MEIR and MPIR,  in spite of the fact that neither scale factor nor rotation angle were applied to generate the  synthetic  frames, except the elastic deformation. This comment also applies to all the other Tables \ref{tab:orie_elas} to \ref{tab:sca1.4_orie_elas}. In fact the changes in scale and orientation are inherent to the elastic deformation procedure ({\it i.e.} are implicit changes) and interestingly the errors shown in Tables \ref{tab:orie_elas} to \ref{tab:sca1.4_orie_elas} confirm this issue, because the magnitude of the scale and orientation errors displayed in these tables is similar to that of Table  \ref{tab:elas}. This means that these errors are essentially related to the  change in scale an orientation produced by the elastic deformation, and the additional, induced,  explicit change in scale or orientation does not increase the errors.

 \begin{table*}[h]
\renewcommand{\arraystretch}{1.3}
\caption{Case iv) at the rotation angle $20$}
\label{tab:sca_orie20_elas}
\centering
 \begin{tabular}{|c||c|c||c|c||c|c|}\hline
   Scale  & \multicolumn{2}{|c||}{$NDM$} &  \multicolumn{2}{c||}{Mean Scale  Error}   &  \multicolumn{2}{c|}{Mean Rotation  Error}    \\ \cline{2-7}
    $\omega_0$                     &  MEIR & MPIR &     MEIR & MPIR   &   MEIR & MPIR \\  \hline   \hline
0.4 & 0.117770 & 0.282990 & 0.016975 & 0.018131 & 4.397200 & 4.978300 \\ \hline
0.6 & 0.112350 & 0.294420 & 0.026690 & 0.028613 & 4.572500 & 5.258800 \\ \hline
0.8 & 0.093531 & 0.299500 & 0.035684 & 0.038745 & 4.291500 & 5.021300 \\ \hline
1.2 & 0.137110 & 0.309090 & 0.055324 & 0.057912 & 4.517900 & 4.931800 \\ \hline
1.4 & 0.202470 & 0.311510 & 0.064034 & 0.066069 & 4.830400 & 5.129400 \\ \hline
1.6 & 0.237270 & 0.298080 & 0.074932 & 0.076590 & 4.782800 & 4.840000 \\ \hline
\end{tabular}
\end{table*}

 \begin{table*}[h!]
\renewcommand{\arraystretch}{1.3}
\caption{Case iv)  at the scale factor $1.4$}
\label{tab:sca1.4_orie_elas}
\centering
 \begin{tabular}{|c||c|c||c|c||c|c|}\hline
  Rotation  & \multicolumn{2}{|c||}{$NDM$} &  \multicolumn{2}{c||}{Mean Scale  Error}   &  \multicolumn{2}{c|}{Mean Rotation  Error}    \\ \cline{2-7}
    $\omega_1$                     &  MEIR & MPIR &     MEIR & MPIR   &    MEIR & MPIR \\  \hline   \hline
5  & 0.222160 & 0.317380 & 0.070345 & 0.071575 & 4.604600 & 4.883100 \\\hline
10 & 0.219680 & 0.312970 & 0.066740 & 0.069023 & 4.364000 & 4.505900 \\\hline
15 & 0.206470 & 0.307480 & 0.066462 & 0.067773 & 4.562800 & 4.792400 \\\hline
20 & 0.199300 & 0.306580 & 0.066079 & 0.068393 & 4.868400 & 5.045800 \\\hline
25 & 0.190730 & 0.301070 & 0.064825 & 0.066755 & 5.100400 & 5.264700 \\\hline
30 & 0.187460 & 0.302130 & 0.065554 & 0.067711 & 5.299400 & 5.510100 \\\hline
\end{tabular}
\end{table*}

\subsubsection{Comments and extra tests}

 The tests described  in Section \ref{sec:testelas},  with artificial frames (elastically deformed),  clearly  show the advantage of  MEIR over MPIR,  to the real objective of WCE localization and orientation, when elastic deformations are involved.
These tests demonstrate that the scale and rotation errors for MEIR are smaller than for MPIR.
This is also connected with the exhibited $NDM$ values. In fact, the  measure  $NDM$ evaluates the quality of the registration approach (more precisely the similarity between reference and template images),  and as  Tables \ref{tab:elas} to \ref{tab:sca1.4_orie_elas} show, NDM is always smaller for MEIR than for MPIR.
So, based on these results and those displayed in  Figure \ref{fig:speed2} (for a video with real successive frames, where $NDM$ is cleary smaller for MEIR than for MPIR),  we expect the scale and rotation errors to be smaller for MEIR, in real consecutive WCE frames, and thus a better accuracy can be achieved  in WCE localization  with the MEIR approach.

We remark that in many existing approaches, dealing with capsule endoscope localization, as for instance   \cite{liu2009capsule,spyrou2014video}, the evaluation of the   methods is done using artificially scaled and rotated video frames, but synthetic elastic deformations are never considered. This is an unrealistic procedure, because the movement of the WCE is caused precisely by the (elastic) deformation of the intestine. Therefore, the  movement between two consecutive video frames with overlapping areas, is always intrinsically associated with a non-rigid movement, which is a much more complex movement than the one originated just by the combination of a rotation and a change of scale.

However, for comparison with the experiments and results, reported in the literature, and obtained by other methods, we have also performed experimental tests with frames that are only artificially rotated and scaled, and whose results we briefly described herein.

Obviously, for these particular tests where the frames are only synthetically rotated and scaled,   MPIR is a better approach than MEIR. In fact,  for these tests the obtained results show that the   scale and orientation errors are lower for MPIR than for MEIR, while the values for the normalized dissimilarity measure $NDM$ are comparable in both approaches (of the order of $10^{-2}$). This is a straightforward, evident and expected result,  due to the definition of MPIR  that searches exactly for an affine transformation, while in MEIR the main goal is to find an elastic deformation, and therefore
we need to consider the affine transformation of the form \eqref{eq:rig}   closest to the solution of the MEIR approach (iterated twice),    to deduce  the   WCE localization and orientation; this procedure clearly induces some approximation errors that causes the slightly worse performance of MEIR compared to MPIR in these particular tests.

However, we  emphasize that when there are elastic deformations involved, the results from the numerous tests  on the artificial frames  (see Tables \ref{tab:elas} to \ref{tab:sca1.4_orie_elas})
 show that  the $NDM$ values for MEIR are significantly lower than the $NDM$ values for MPIR.  Therefore, a possible procedure to adopt, assuming the unrealistic scenario that there might be some  WCE movements that are strictly rigid-like,  and because in that case the $NDM$ values in both approaches, MEIR and MPIR,  are comparable and of the order of $10^{-2}$ (as aforementioned),   is the following:

\begin{itemize}

\item  For a pair of consecutive frames apply MPIR and also MEIR.

\item Compute $NDM$ for MPIR and MEIR, hereafter denoted by $NDM^{MPIR}$ and $NDM^{MEIR}$, respectively.

\item If $NDM^{MPIR}$ and  $NDM^{MEIR}$  are comparable  (of the order of $10^{-2}$), consider the approach MPIR. If  $NDM^{MEIR}$ is significantly lower than  $NDM^{MPIR}$ (this means that elastic deformations are present), adopt the MEIR approach for this pair of frames.

\end{itemize}

Hence in the sequel we restrict ourselves to the description of the results obtained with MPIR for these particular tests (where the frames are only synthetically rotated and scaled) and which haven proven to be better than those reported in the literature with other methods.

In a first test  we have created  rotated versions of the 780 frames, by using nine rotation angles from $5$ to $45$ with a step of 5, at the original scale  and   then we have proceeded with the  image registration  of the original frames and their rotated versions  with MPIR.
The obtained results concerning  the mean (absolute value) orientation errors are of the order $10^{-3}$, except for angle $45$ , where the error is of the order  $10^{-1}$. These  are better results than those reported in \cite{liu2009capsule,spyrou2014video} with other methods, where very large orientation errors occur when the rotation angle increases.

Then in  a second test we have
generated scaled versions of the 780 frames, using nine different scales from a factor of $0.2$ to $2.0$ and
have performed the registration with the originals, using   MPIR. The  mean (absolute value) scale error stay in the some order of magnitude (approximately  between $10^{-2}$ and $10^{-5}$), while in \cite{spyrou2014video} the mean (absolute value) scale error is extremely big for small scales.

In addition we have also registered with MPIR  each original grayscale image and a synthetically version of it, generated by simultaneously applying a rotation and a factor of scale. More specifically,  in a third test  we have fixed the scale $\omega_0$ at a factor of $2.0$ and varied the rotation angles $\omega_1$ from $5$ to $40$ with a step of $5$, and for the fourth test,  we  fixed the rotation angle $\omega_1$ at $30$ and varied the factor of scale $\omega_0$ from $0.4$ to $2.0$ with a factor of $0.2$. Again, for MPIR the mean absolute value errors,  for scale  and orientation, stay in the same order of magnitude. In the third test the mean rotation error increased with the angle,  from $0.24$ (at angle $5$) to $1.74$  (at angle $40$). In the fourth test the oder of the  mean scale error varied between   $10^{-3}$ to $10^{-5}$. We did not obtain large errors at the small scale or at the big rotation
angle as reported in \cite{spyrou2014video}.

\section{Conclusions}\label{sec:conc}

In this paper a multiscale elastic image registration has been proposed as a tool
for tracking the movement of the walls of the small intestine, in WCE video frames, and subsequently for tracking the WCE motion.
The proposed procedure, that involves an affine pre-registration, takes into account the rigid-like and non-rigid movements to which the WCE is subjected within the small intestine, and that are a consequence of peristalsis.

 The qualitative WCE speed information provided by this approach, through the dissimilarity measure $NDM$,  is medically practical,  useful and facilitates the video interpretation.
The  tests also evidence  the relevance of this  $NDM$ measure, relative to MEIR, since from artificial data we conclude that smaller $NDM$ leads to smaller errors in WCE location and orientation.  In addition, the experiments with real frames, described in Section  \ref{sec:testreal}, demonstrate  the accuracy of the WCE velocity estimation as a function of $NDM$.
However peak speed points, that correspond to sudden changes of the image content in consecutive frames, should be further studied.

The proposed approach is also compared with a multiscale parametric image registration, that is similar to other existing approaches, that as this latter one,  essentially rely on affine correspondences between consecutive frames, and consequently  are only capable of capturing rigid-like movements.
The comparison is done in terms of the qualitative WCE speed information, the  dissimilarity measure for evaluating the registration,  and in terms of the WCE location and orientation by following  \cite{spyrou2014video} (for this  the scale and rotation parameters, resulting from the  affine  transformation   closest to the solution of the proprosed  approach,  are computed  and then identified with the capsule displacement and orientation,  using a projective transformation and the pinhole camera model).
The overall results indicate a better performance of the multiscale elastic image registration than the multiscale parametric image registration, when there are elastic deformations involved, which is a realistic situation in the WCE images.

Finally, we note that the multiscale elastic image registration  herein proposed is an image-based motion procedure, that could  be also integrated or used  as a complement,  in other  more complex existing approaches for  WCE localization,  involving extra sensors other than the WCE, for improving their accuracy.

 \section*{Acknowledgment}
This work was partially supported by the  project
PTDC/MATNAN/0593/2012 funded by FCT (Portuguese national funding agency for science, research and technology), and also by CMUC (Center for Mathematics,  University of Coimbra) and FCT, through European program COMPETE/ FEDER and
project PEst-C/MAT/UI0324/2013. Richard Tsai is supportably partially by National Science Foundation Grant DMS-1217203.

 \bibliographystyle{plain}
 \bibliography{WCE_preprint}

\end{document}